\documentclass{article}

\usepackage{hyperref}

\usepackage[accepted]{icml2023}

\usepackage{float}
\usepackage{url}

\usepackage{caption}
\usepackage{subcaption}
\usepackage{graphicx}
\usepackage{placeins}
\usepackage{paralist}

\usepackage{microtype}
\usepackage{graphicx}
\usepackage{booktabs} %

\usepackage{amsmath}
\usepackage{amssymb, pifont}
\usepackage{mathtools}
\usepackage{amsthm}
\usepackage{enumitem}
\usepackage{adjustbox, bbm}
\usepackage{wrapfig}
\usepackage{multirow}
\usepackage{breakcites}
\usepackage{arydshln}

\global\long\def\reals{\mathbb{R}}
\global\long\def\pr{\mathbb{P}}
\global\long\def\E{\mathbb{E}}

\global\long\def\perf{\text{Perf}}
\global\long\def\psrc{\cD^\text{source}}
\global\long\def\ptar{\cD^\text{target}}
\global\long\def\noscriptcand{\texttt{C}}
\global\long\def\cand{\texttt{C}_{\cD}}
\global\long\def\subcand{\widetilde{\texttt{C}}}
\global\long\def\pa{\text{parent}}
\global\long\def\val{\text{Val}}
\global\long\def\attr{\text{Attr}}

\global\long\def\bigO{\mathcal{O}}

\def\to{{\,\rightarrow\,}}

\mathchardef\mhyphen="2D

\newcommand{\vertiii}[1]{{\left\vert\kern-0.25ex\left\vert\kern-0.25ex\left\vert #1
    \right\vert\kern-0.25ex\right\vert\kern-0.25ex\right\vert}}

\def\cD{\mathcal{D}}

\usepackage[capitalize,noabbrev]{cleveref}

\theoremstyle{plain}
\newtheorem{theorem}{Theorem}[section]

\theoremstyle{definition}

\newtheorem{assumption}[theorem]{Assumption}
\theoremstyle{remark}
\newtheorem{remark}[theorem]{Remark}

\newcommand{\cmark}{\ding{51}}%
\newcommand{\xmark}{\ding{55}}%

\begin{document}

\twocolumn[
\icmltitle{``Why did the Model Fail?'': Attributing Model Performance Changes to Distribution Shifts}

\icmlsetsymbol{equal}{*}

\author{Haoran Zhang\thanks{Equal contribution.}\\
MIT\\
\texttt{haoranz@mit.edu} \\
\And
Harvineet Singh\textsuperscript{*} \\
New York University \\
\texttt{hs3673@nyu.edu} \\
\And
Marzyeh Ghassemi \\
MIT \\
\texttt{mghassem@mit.edu}
\And
Shalmali Joshi \\
Harvard University\\
\texttt{shalmali@seas.harvard.edu}
}

\begin{icmlauthorlist}
\icmlauthor{Haoran Zhang}{equal,yyy}
\icmlauthor{Harvineet Singh}{equal,zzz}
\icmlauthor{Marzyeh Ghassemi}{yyy}
\icmlauthor{Shalmali Joshi}{aaa}
\end{icmlauthorlist}

\icmlaffiliation{yyy}{MIT}
\icmlaffiliation{zzz}{New York University}
\icmlaffiliation{aaa}{Columbia University}

\icmlcorrespondingauthor{Haoran Zhang}{haoranz@mit.edu}

\icmlkeywords{Machine Learning, ICML}

\vskip 0.3in
]

\printAffiliationsAndNotice{\icmlEqualContribution} %

\begin{abstract}

Machine learning models frequently experience performance drops under distribution shifts. The underlying cause of such shifts may be multiple simultaneous factors such as changes in data quality, differences in specific covariate distributions, or changes in the relationship between label and features. When a model does fail during deployment, attributing performance change to these factors is critical for the model developer to identify the root cause and take mitigating actions. In this work, we introduce the problem of attributing performance differences between environments to distribution shifts in the underlying data generating mechanisms. We formulate the problem as a cooperative game where the players are distributions. We define the value of a set of distributions to be the change in model performance when only this set of distributions has changed between environments, and derive an importance weighting method for computing the value of an arbitrary set of distributions. The contribution of each distribution to the total performance change is then quantified as its Shapley value. We demonstrate the correctness and utility of our method on synthetic, semi-synthetic, and real-world case studies, showing its effectiveness in attributing performance changes to a wide range of distribution shifts.
\end{abstract}

\section{Introduction} 

Machine learning models are widely deployed in dynamic environments ranging from recommendation systems to personalized clinical care. Such environments are prone to distribution shifts, which may lead to serious degradations in model performance \citep{guo2022evaluation, chirra2018empirical, koh2021wilds, geirhos2020shortcut, nestor2019feature, yang2023change}. Importantly, such shifts are hard to anticipate and reduce the ability of model developers to design reliable systems.

When the performance of a model \textit{does} degrade during deployment, it is crucial for the model developer to know not only which distributions have shifted, but also \textit{how much} a specific distribution shift contributed to model performance degradation. 
Using this information, the model developer can then take mitigating actions such as additional data collection, data augmentation, and model retraining \citep{ashmore2021assuring, zenke2017continual, subbaswamy2019preventing}.

In this work, we present a method to attribute changes in model performance to shifts in a given set of distributions. Distribution shifts can occur in various marginal or conditional distributions that comprise variables involved in the model. Further, multiple distributions can change simultaneously. We handle this in our framework by defining the effect of changing any set of distributions on model performance, and use the concept of Shapley values \citep{shapley1953value} to attribute the change to individual distributions. The Shapley value is a co-operative game theoretic framework with the goal of distributing surplus generated by the players in the co-operative game according to their contribution. In our framework, the players correspond to individual distributions, or more precisely, mechanisms involved in the data generating process.

Most relevant to our contributions is the work of \citet{budhathoki2021distribution}, which attributes a shift between two joint distributions to %
a specific set of individual distributions. The distributions here  correspond to the components of the factorization of the joint distribution when the data-generating process is assumed to follow causal structural assumptions. This line of work defines distribution shifts as interventions on causal mechanisms \citep{pearl2011transportability,subbaswamy2019preventing,subbaswamy2021evaluating,budhathoki2021distribution, thams2022evaluating}. We build on their framework to justify the choice of players in our cooperative game. We significantly differ from the end goal by attributing a change in \textit{model performance between two environments} to individual distributions. 
Note that each shifted distribution may influence model performance differently and may result in significantly different attributions than their contributions to the shift in the joint distribution between environments. 

In this work, we focus on explaining the discrepancy in model performance between two environments as measured by some metric such as prediction accuracy. We emphasize the non-trivial nature of this problem, as many distribution shifts will have no impact on a particular model or metric, and some distribution shifts may even increase model performance. Moreover, the root cause of the performance change may be due to distribution shifts in variables external to the model input. Thus, explaining performance discrepancy requires us to develop specialized methods. Specifically, we want to quantify the contribution to the performance change of a fixed set of distributions that may change across the environments.  Given such a set, we develop a model-free importance sampling approach to quantify this contribution. We then use the Shapley value framework to estimate the attribution for each distribution shift. This framework allows us to expand the settings where our method is applicable.

We make the following contributions\footnote{Code: \url{https://github.com/MLforHealth/expl_perf_drop}}:
\begin{itemize}[noitemsep,topsep=0pt,  leftmargin=*]
    \item We formalize the problem of attributing model performance changes due to distribution shifts.
    \item We propose a principled approach based on Shapley values for attribution, and show that it satisfies several desirable properties.
    \item We validate the correctness and utility of our method on synthetic and real-world datasets.
\end{itemize}

\begin{figure*}[t]
        \centering
        \includegraphics[width=0.99\linewidth]{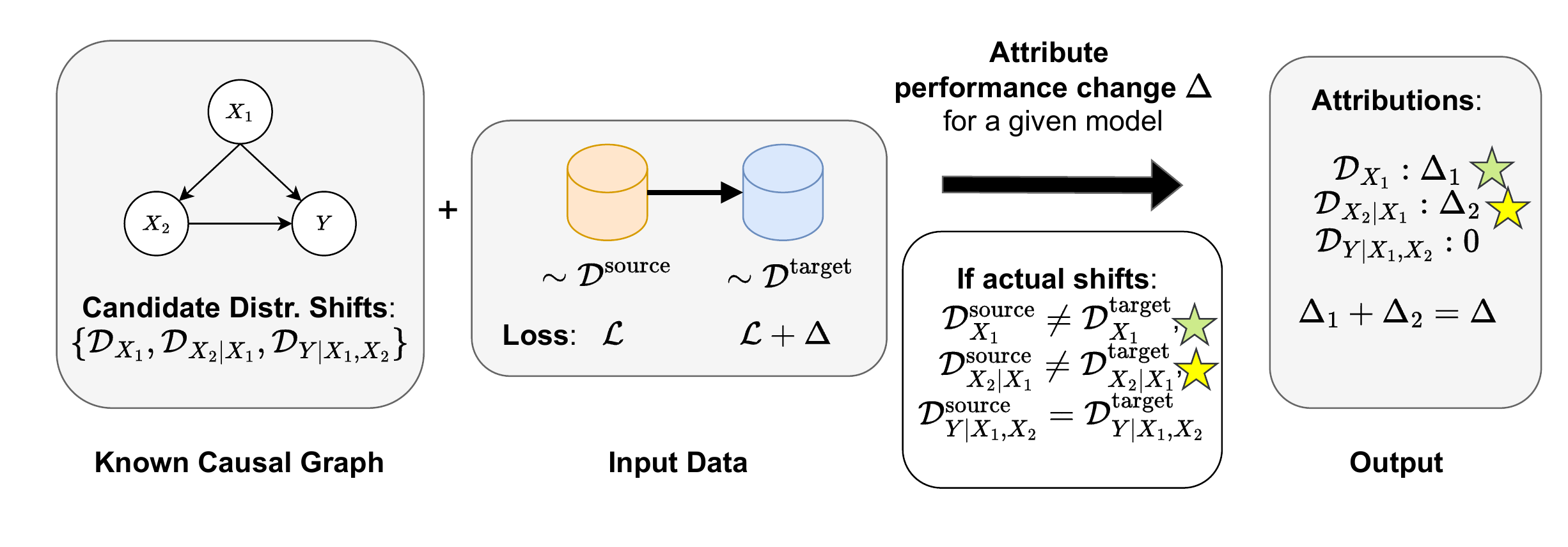}
    \caption{\textbf{Inputs and outputs for attribution.} Input: Causal graph, where all variables are observed providing the candidate distribution shifts we consider. The goal is to attribute the model's performance change $\Delta$ between source and target distributions to these candidate distributions. Here, out of the three candidate distributions, the marginal distribution of $X_1$ and the conditional distribution of $X_2$ given $X_1$ change. Our method attributes changes to each one such that the attributions sum to the total performance change  $\Delta$. Note that nodes in the causal graph may be vector-valued, which allows our method to be used on high-dimensional data such as images. \label{fig:example} }
\end{figure*}

\section{Problem Setup}\label{sec:preliminaries}

\textbf{Notation.} Consider a learning setup where we have some system variables denoted by $V$ consisting of two types of variables $V=(X,Y)$, which comprises of features $X$ and labels $Y$ such that $V \sim \cD$. Realizations of the variables are denoted in lower case. We assume access to  samples from two environments. We use $\psrc$ to denote the source distribution and $\ptar$ for the target distribution. Subscripts on $\cD$ refer to the distribution of specific variables. For example, $\cD_{X_1}$ is the distribution of feature $X_1 \subset X$, and $\cD_{Y| X}$ is the conditional distribution of labels given all features $X$. 

Let $X_{\texttt{M}} \subseteq X$ be the subset of features utilized by a given model $f$. We are given a loss function $\ell((x,y), f) \mapsto  \reals$ which assigns a real value to the model evaluated at a specific setting  $x$ of the variables. For example, in the case of supervised learning, the model $f$ maps $X_{\texttt{M}}$ into the label space, and a loss function such as the squared error $\ell((x,y),f) := (y-f(x_{\texttt{M}}))^2$ can be used to evaluate model performance. We assume that the loss function can be computed separately for each data point. Then, performance of the model in some environment with distribution $\cD$ is summarized by the average of the losses:
\begin{equation*}
 \perf(\cD) := \E_{(x, y)\sim \cD} [\ell((x, y), f)]   
\end{equation*}
This implies that a shift in any variables $V$ in the system may result in performance change across environments, including those that are not directly used by the model, but drive changes to the features $X_{\texttt{M}}$ used by the model for learning.

{\textbf{Setup.}} Suppose we are given a \textit{candidate set} of (marginal and/or conditional) distributions  $\cand$ over $V$ that may account for the model performance change from $\psrc$ to $\ptar$: $\perf(\ptar) - \perf(\psrc)$. %
\textbf{Our goal is to attribute this change to each distribution in the candidate set $\cand$.} %
For our method, we assume access to the model $f$, and samples from $\psrc$ as well as  $\ptar$ (see Figure~\ref{fig:example}).

We assume that dependence between variables $V$ is described by a causal system~\cite{pearl2009causality}. For every variable $X_i \in V$, this dependence is captured by a functional relationship between $X_i$ and the so-called ``causal parents'' of $X_i$ (denoted as $\text{parent}(X_i)$) driving the variation in $X_i$. The causal dependence induces a Markov distribution over the variables in this system. That is, the joint distribution $\cD_V$ can be factorized as, $\cD_V = \prod_{X_i \in V} \cD_{X_i | \text{parent}(X_i)}$. This dependence can be summarized graphically using a Directed Acyclic Graph (DAG) with nodes corresponding to the system variables and directed edges ($\text{parent}(X_i) \rightarrow X_i$) in the direction of the causal mechanisms in the system (see Figure~\ref{fig:example} for an example).

\textbf{Example.} We provide an example that illustrates that the performance attribution problem is ill-specified without knowing how the mechanisms can change to result in the observed performance difference. Suppose we are predicting $Y$ from $X$ with a linear model $f(x):=\phi x$ under the squared loss function. Consider two possible scenarios for data generation -- (1) $X\leftarrow Y$ where $\cD_{Y}$ changes from source to target while $\cD_{X|Y}$ remains the same, (2) $X\rightarrow Y$ where $\cD_{X}$ changes from source to target while $\cD_{Y|X}$ remains the same. The performance difference of $f(x)$ is the same in both the cases. Naturally, we want an attribution method to  assign all of the difference to the mechanism for $Y$ in the first case and to the mechanism of $X$ in the second case. Thus, for the same performance difference between source and target data, we would like a method to output different attributions depending on whether the data generating process is case (1) or (2). Note that, in general, it is impossible to find the appropriate attributions by first finding the direction of the causal mechanisms. This follows from the fact that learning the structure is in general, impossible purely from observational data~\citep{peters2017elements}. Hence knowledge of the data-generating mechanisms is necessary for appropriate attribution.

More concretely, suppose the processes are (1) $Y\sim N(\mu_1,1), X\sim Y+N(0,1)$. The mean of $Y$ shifts to $\mu_2$ in target, and (2) $X\sim N(\mu_1,1), Y\sim X+N(0,1)$ where the mean of $X$ shifts to $\mu_2$ in target. For the model $f(x):=\phi x$, the performance difference $\Delta$ in both cases is $(1-\phi)^2(\mu_2^2-\mu_1^2)$. This example illustrates the need for specifying how the mechanisms can shift from source to target to solve the attribution problem. In this work, we use partial causal knowledge, {\bf{in terms of the causal graph only}}, to specify the data-generating mechanisms.

In general, this partial knowledge further allows us to identify potential shifts to consider. Specifically, the number of marginal and conditional shifts that can be defined over $(X,Y)$ is exponential in the dimension of $X$. The factorization induced by the causal graph or equivalently knowledge of the data-generating mechanism reduces the space of possible shifts to consider for attribution. See Section~\ref{sec:method} for additional advantages of using a causal framework. %

\section{Method}\label{sec:method}

We now formalize our problem setup and motivate a game theoretic method for attributing performance changes to distributions over variable subsets (See Figure \ref{fig:example} for a summary).  We proceed with the following Assumptions.

\begin{assumption}\label{ass:graph_known}
The causal graph corresponding to the data-generating mechanism is known and all variables in the system are observed. Thus, the factorization of the joint distribution $\cD_V$ is known.
\end{assumption}

\begin{assumption}\label{ass:shifts_as_interventions}
Distribution shifts of interest are due to (independent) shifts in one or more factors of $\cD_V$.
\end{assumption}

Given these assumptions, we now describe our game theoretic formulation for attribution.  
\begin{figure}[t]
            \centering
            \includegraphics[width=0.5\textwidth]{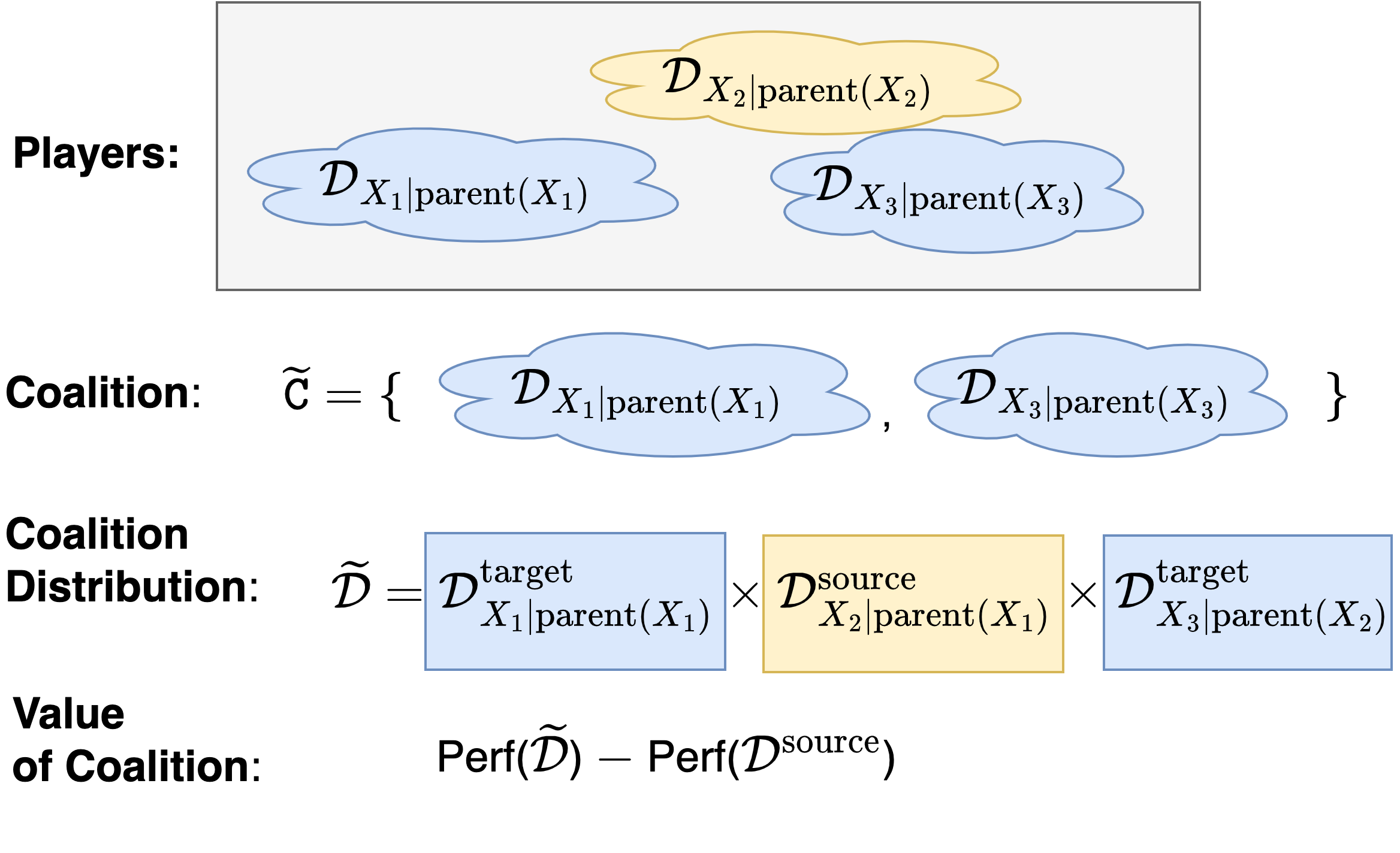}
            \caption{\textbf{Sketch of the game theoretic attribution method.} Each causal mechanism is a player that, if present in the coalition, changes to the target distribution and, if absent, remains fixed at the source distribution. This defines the distribution of the resulting coalition $\widetilde{\cD}$. Performance on $\widetilde{\cD}$ is estimated using importance sampling from training data samples. After computing values for each possible coalition, Shapley value (Eq. \ref{eq:shapley}) gives the attribution to each player. Thus, we estimate the performance change under all possible ways to shift the mechanisms from source to target and use these to distribute the total performance change among the individual distributions.}
            \label{fig:method}
        \end{figure}

\subsection{Game Theoretic Distribution Shift Attribution}
We consider the set of candidate distributions $\cand$ as the   \textit{players} in our attribution game. A \textit{coalition} of any subset of players determines the distributions that are allowed to shift (from their source domain distribution to the target domain distribution), keeping the rest fixed. The \textit{value} for the coalition is the model performance change between the resulting distribution for the coalition and the training distribution. See Figure~\ref{fig:method} for an overview of the method.

\paragraph{Value of a Coalition.} 
Consider a coalition of distributions $\subcand \subseteq \cand$. This coalition implies a joint distribution over system variables $V$, where members in the coalition contribute their target domain distribution, and non-members contribute their source domain distribution:

\begin{equation}\label{eq:src_d}
\begin{split}
    \widetilde{\cD} = & \underbrace{\left(\prod_{i: \cD_{X_i | \pa(X_i)}\in \subcand}  \ptar_{X_i | \pa(X_i)} \right)}_{\text{Coalition}} \cdot \\ &  \quad \underbrace{\left( \prod_{i: \cD_{X_i | \pa(X_i)}\not\in \subcand} \psrc_{X_i | \pa(X_i)} \right)}_{\text{Not in Coalition}}
    \end{split}
\end{equation}

The above factorization follows from Assumptions~\ref{ass:graph_known} and~\ref{ass:shifts_as_interventions}. Note that the coalition only consists of distributions that are allowed to change across environments. All other relevant mechanisms are indeed fixed to the source distribution. We present an example of a coalition of two players in Figure \ref{fig:method}. The value of the coalition $\subcand$ with the coalition distribution $\widetilde{\cD}$ is now given by
\begin{equation}
    \label{eq:value}
     \val(\subcand) := \perf(\widetilde{\cD}) - \perf(\psrc)\\
\end{equation}
Thus, our assumptions  allow us to represent a factorization where only members of the coalition change, while all other mechanisms correspond to the source distribution. If we consider the change in performance for all combinatorial coalitions, we can estimate the total contribution of a specific distribution by aggregating the value for all possible coalitions a candidate distribution is a part of. This is exactly the Shapley value applied to a set of distributions. The Shapley value framework thus allows us to obtain the attribution of each player $d \in \cand$ using Equation \ref{eq:shapley}.

Abstractly, the Shapley values framework \citep{shapley1953value} is a game theoretic framework which assumes that there are $\noscriptcand := \{1, 2, \dots, n\}$ players in a co-operative game, achieving some total value (in our case, model performance change). We denote by $\val : 2^\noscriptcand \mapsto \reals$, the value for any subset of players, which is called a coalition. %
Shapley values correspond to the fair assignment of the value $\val(\noscriptcand)$ to each player $d \in \noscriptcand$. The intuition behind Shapley values is to quantify the change in value when a player (here, a distribution) enters a coalition. Since the change in model performance depends on the order in which players (distributions) may join the coalition, Shapley values aggregate the value changes over all permutations of $\noscriptcand$. Thus the Shapley attribution $\attr(d)$ for a player $d$ is given by:
\begin{equation}
    \label{eq:shapley}
    \attr(d) = \frac{1}{|\noscriptcand|} \sum_{\subcand\subseteq \noscriptcand \setminus \{d\}} \binom{|\noscriptcand| - 1}{|\subcand|}^{-1} \left( \val(\subcand \cup \{d\}) - \val(\subcand)\right)
\end{equation}

where we measure the change in model performance (denoted by Val) after adding $d$ to the coalition averaged over all potential coalitions involving $d$. The computational complexity of estimating Shapley values is  exponential in the number of players. Hence we rely on this exact expression only when the number of candidate distributions is small. That is, the causal graph induces a factorization that results in smaller candidate sets. For larger candidate sets, we use previously proposed approximation methods \citep{castro2009polynomial,lundberg2017unified, janzing2020feature} for reduced computational effort.

\paragraph{Choice of Candidate Distribution Shifts.} We motivate further the choice of candidate distributions that will inform the coalition. %
As mentioned before, without the knowledge of the causal graph, many heuristics for choosing the candidate sets are possible. For example, a  candidate set could be the set of all marginal distributions on each system variable, $\cand=\{\cD_{X_1},\cD_{X_2},\cdots\}$, or distribution of each variable after conditioning on the rest, $\cand=\{\cD_{X_1|V\setminus X_1},\cD_{X_2|V\setminus X_2},\cdots\}$. Since we have combinatorially many shifts that can be defined on subsets of $V=(X,Y)$, choosing candidate sets that would then inform the coalition is challenging.  %
The causal graph, on the other hand, specifies the factorization of the joint distribution into a set of distributions. %
We form the candidate set  constituting each distribution in this factorization. That is,
\begin{equation*}
    \cand = \{\cD_{X_1 | \pa(X_1)}, \cdots,\cD_{X_i | \pa(X_i)}, \cdots\}_{i=1,\cdots,|V|}
\end{equation*}
For a node without parents in the causal graph, the parent set can be empty, which reduces $\cD_{X_i|\text{parent}(X_i)}$ to the  marginal distribution of $X_i$. 
This choice of candidate set has three main advantages. First, it is \textit{interpretable} since the candidate shifts are specified by domain experts who constructed the causal graph. Second, it is \textit{actionable} since identifying the causal mechanisms most responsible for performance change can inform mitigating methods for handling distribution shifts \citep{subbaswamy2019preventing}. Third, it will lead to \textit{succinct} attributions due to the independence property. 

Consider the case where only one conditional distribution $\cD(X_i | \pa(X_i))$ changes across domains. This will result in a change in distributions of all descendants of $X_i$ (due to the above factorization). In this case, a candidate set defined by all marginals is not succinct, as one would attribute performance changes to all marginals of descendants of $X_i$. Instead, the candidate set determined by the causal graph will isolate the correct conditional distribution.

Crucially, to compute our attributions, we need estimates of model performance under $\widetilde{\cD}$. Note that we only have model performance estimates under $\psrc$ and $\ptar$, but not for any arbitrary coalition where only a subset of the distributions have shifted. To estimate the performance of any coalition, we propose to use importance sampling.

\subsection{Importance Sampling to Estimate Performance under a Candidate Distribution Shift}

\begin{assumption}\label{ass:support}
$\texttt{support}(\ptar_{X_i | \pa(X_i)}) \subseteq \texttt{support}(\psrc_{X_i | \pa(X_i)}) $ for all $\ptar_{X_i | \pa(X_i)} {\in} \cand$.
\end{assumption}

Importance sampling allows us to re-weight samples drawn from a given distribution, which can be $\psrc$ or $\ptar$, to simulate expectations for a desired distribution, which is the candidate $\widetilde{\cD}$ in our case. Thus, we re-write the value as
\begin{align}
    \val(\subcand) &= \perf(\widetilde{\cD}) - \perf(\psrc) \label{eq:val_def} \\
    &= \E_{(x, y)\sim \widetilde{\cD}} [\ell((x,y),f)] - \E_{(x,y)\sim \psrc} [\ell((x,y),f)] \nonumber\\
    &= \E_{(x,y)\sim \psrc} \left[\frac{\widetilde{\cD}((x,y))}{\psrc((x,y))}\ell((x,y),f)\right] \nonumber - \\ & \qquad \E_{(x,y)\sim \psrc} [\ell((x,y),f)] \nonumber
\end{align}
The importance weights are themselves a product of ratios of source and target distributions corresponding to the causal mechanisms in $\cand$ as follows:
\begin{equation}
\begin{split}
    w_{\subcand}((x,y)) &:= \frac{\widetilde{\cD}((x,y))}{\psrc((x,y))} = \prod_{d \in \subcand} \frac{\ptar_d((x,y))}{\psrc_d((x,y))} \\ &=: \prod_{d \in \subcand} w_{d} ((x,y)) %
\end{split}
\end{equation}
By Assumption \ref{ass:support}, we ensure that all importance weights are finite.

\paragraph{Computing Importance Weights.} 
There are multiple ways to estimate importance weights $w_d((x, y))$, which are a ratio of densities \citep{sugiyama2012density}. Here, we use a simple approach for density ratio estimation via training probabilistic classifiers as described in \citet[][Section 2.2]{sugiyama2012density}.

Let $D$ be a binary random variable, such that when $D=1, Z\sim \ptar_d(Z)$, and when $D=0, Z\sim \psrc_d(Z)$. Suppose $d = \cD_{X_i | \pa(X_i)}$, then $$w_d =\frac{\pr(D = 0 | \pa(X_i))}{\pr(D = 1| \pa(X_i))} \cdot \frac{\pr(D = 1 | X_i, \pa(X_i))}{\pr(D = 0 | X_i, \pa(X_i))}, $$ where each term is computed using a probabilistic classifier trained to discriminate data points from $\psrc$ and $\ptar$ from the concatenated dataset. We show the derivation of this equation in Appendix \ref{app:iw_deriv}. In total, we need to learn $\bigO(|\cand|)$ models for computing all importance weights.

\subsection{Properties of Our Method}

Under perfect computation of importance weights, the Shapley attributions resulting from the performance-change game have the following desirable properties, which follow from the properties of Shapley values. We provide proofs of these properties in Appendix \ref{app:proof_property}.

\begin{asparaenum}
\item[] {\bf{Property 1. (Efficiency)}} $\displaystyle\sum_{d \in \cand} \attr(d) = \val(\cand) = \perf(\ptar) - \perf(\psrc)$ %

\item[] {\bf{Property 2.1. (Null Player)}} $\psrc_d = \ptar_d \implies \attr(d) = 0$.

\item[] {\bf{Property 2.2. (Relevance)}}  Consider a mechanism $d$. If $\perf(\subcand \cup \{d\}) = \perf(\subcand)$ for all $\subcand \subseteq \cand \setminus d$, then ${\attr(d) = 0}$. 

\item[] {\bf{Property 3. (Attribution Symmetry)}} Let $\attr_{\cD_1, \cD_2}(d)$ denote the attribution to some mechanism $d$ when $\cD_1 = \psrc$ and $\cD_2 = \ptar$. Then, $\attr_{\cD_1, \cD_2}(d) = -\attr_{\cD_2, \cD_1}(d) \ \forall d \in \cand$.
\end{asparaenum}

Thus, the method attributes the overall performance change only to distributions that actually change in a way that affects the specified performance metric. The contribution of each distribution is computed by considering how much they impact the performance if they are made to change in different combinations alongside the other distributions. %

\subsection{Analysis using a Synthetic Setting}
\label{sec:simple_syn}
We derive analytical expressions for  attributions in a simple synthetic case with the following data generating process.
\begin{equation*}
    \begin{aligned}
    \text{Source}:  X &\sim \mathcal{N}(\mu_1, \sigma_X^2) \\
    Y &\sim \theta_1 X + \mathcal{N} (0, \sigma_Y^2) \\
    \text{Target}: X &\sim \mathcal{N}(\mu_2, \sigma_X^2) \\
    Y &\sim \theta_2 X + \mathcal{N} (0, \sigma_Y^2)
    \end{aligned}
\end{equation*}

The model that we are investigating is $f(X) = \phi X$, and $l((x, y), f) = (y - f(x))^2$.

\begin{table*}[t]
    \centering
     \caption{Analytical expressions of the attributions for the synthetic case described in Section \ref{sec:simple_syn}. For the full derivation, see Appendix \ref{app:synthetic}.}
    \begin{adjustbox}{max width=\textwidth}
    \begin{tabular}{lll}
    \toprule
        & \textbf{Attr($\cD_X$)} & \textbf{Attr($\cD_{Y|X}$)} \\
        \midrule
        \textbf{Ours} & $(\frac{1}{2}\mu_2^2 - \frac{1}{2}\mu_1^2) ((\theta_2 - \phi)^2 + (\theta_1 - \phi)^2)$  &  $ (\sigma_X^2 + \frac{1}{2} \mu_1^2 + \frac{1}{2} \mu_2^2) ((\theta_2 - \phi)^2 - (\theta_1 - \phi)^2) $ \\
        \midrule
        \textbf{\citet{budhathoki2021distribution}} & $\frac{(\mu_2 - \mu_1)^2}{2\sigma_X^2}$ & $\frac{(\theta_2 - \theta_1)^2}{2 \sigma_Y^2} (\sigma_X^2 + \mu_2^2)$ \\
        \bottomrule
    \end{tabular}
    \end{adjustbox}
    \label{tab:synthetic_expr}
\end{table*}

We show the attribution of our method, along with the attribution using the joint method from \citet{budhathoki2021distribution}, in Table \ref{tab:synthetic_expr}. The complete derivation, along with experimental verification of the derived expressions, can be found in Appendix \ref{app:synthetic}. We highlight several advantages that our method has over the baseline.

First, our attribution takes the model parameter $\phi$ into account in order to explain model performance changes, whereas~\citet{budhathoki2021distribution} do not, as they only explain shifts in $(X, Y)$, or changes in simple functions such as $\mathbb{E}[X]$ of the variables. Second, we find that our $\attr(\cD_X)$ is a function of $\theta_2$. This is desirable, as covariate shift may compound with concept shift to increase loss non-linearly. This also ensures that both attributions always sum to the total shift. Third, we note that our attributions are \textit{signed}, which is particularly important as some shifts may decrease loss. Finally, we note that our attributions are symmetric when the source and target data distributions are swapped by Property 3. This is not true of the baseline method in general, as the KL divergence is asymmetric. Since we assume knowledge of the true causal graph (which provides the factorization that determines the coalition), we also evaluate the attribution when the graph is misspecified. In this case, the coalition will consist of $\{\cD_Y, \cD_{X|Y}\}$. We include these attribution results in Appendix \ref{app:add_results_syn}, specifically, Figure  \ref{fig:app_synthetic_misspec}. In this case, as expected, both $\cD_Y$ and $\cD_{X|Y}$ are attributed the change in model performance (at varying levels depending on the magnitude of concept shift). While this may still be a meaningful attribution, knowledge of the causal graph provides a more succinct interpretation of system behavior.  

\section{Related Work}

\textbf{Identifying relevant distribution shifts.}
There has been extensive work that tests whether the data distribution has shifted (e.g. ones evaluated in \citet{rabanser2019failing}). Past work has proposed to identify sub-distributions (factors constituting the joint distribution as determined by a generative model for the data) that comprise the shift between two joint distributions and order them by their contribution to the shift \citep{budhathoki2021distribution}. However, as suggested before, the sub-distributions may have different influence on model performance. Even a small change in some (factors) may have a large effect on model performance (and vice-versa). %
Thus, a model developer has to filter distributions to identify ones that actually impact model performance (see Property 2.2 and Appendix \ref{app:synthetic}). Further, ~\citet{budhathoki2021distribution} focuses on changes to the joint distribution as measured by the KL-divergence, which requires assumptions on the class of distributions to leverage closed-form expressions of KL-divergence (such as exponential families), or non-parametric KL estimation which is challenging in high dimensions \citep{wang2005divergence, wang2006nearest}. %

Other approaches which aim to localize shifts to individual variables (conditional on the rest of the variables) do not provide a way to identify the ones relevant to performance \citep{kulinski2020detection}. In contrast to testing for shifts, \citet{podkopaev2022tracking} tests for changes in model performance when distribution changes in deployment. Recent work by \citet{wu2021performance} decomposes performance change to changes in only marginal distributions using Shapley value framework~\citep{lundberg2017unified}. However, the method as described is restricted to categorical variables. \citet{kulinski2022towards} propose a method for distribution shift explanation based on transport mappings, though their focus is still on how the distribution has shifted and not its impact on some downstream model. %
In parallel work, \citet{cai2023diagnosing} propose a method for attributing performance degradation to distribution shifts, but their decomposition is limited to $P(X)$ and $P(Y|X)$ and thus has limited granularity.
Finally, \citet{jung2022measuring} propose a framework for measuring causal contributions to expected change in outcomes using Shapley values. This work focuses on attributions to specific feature \textit{values}, by framing causal attribution as hard/do-interventions in a causal system. Further, they only examine the system in a single domain,  and focus on the expected causal effect on a target variable $Y$. On the other hand, our work can be considered to focus on attributing model performance change to \textit{mechanisms} by considering mechanism changes across domains as ``soft-interventions''. Further, our main goal is to attribute model performance change across two fixed domains where we have access to iid samples from both domains.

\textbf{Shapley values for attribution.}
Shapley value-based attribution has recently become popular for interpreting model predictions \citep{strumbelj2014explaining,lundberg2017unified,wang2021flow}. In most prior work, Shapley values have been leveraged for attributing a specific model prediction to the input features~\citep{sundararajan2020many}. Challenges to appropriately interpreting such attributions and desirable properties thereof have been extensively discussed in~\cite{janzing2020feature, kumar2021shapley}. In this work, we advance the use of Shapley values for interpreting model performance changes to individual distributions at the dataset level. 

\textbf{Detecting data partitions with low model performance.}
Recent work aims to find subsets of the dataset that have significantly worse (or better) performance \citep{deon2021spotlight,eyuboglu2022domino, jain2022distilling, park2023trak}. However, they do not study changes in the underlying data distribution. The work by \citet{ali2022lifecycle} describes a method to identify and localize a change in model performance, and is applicable under distribution shifts. %
The main difference in our work is the data representations used for attribution. Instead of identifying subsets of \textit{data} that are relevant to performance change, we find individual \textit{distributions} represented by causal mechanisms. %

\FloatBarrier
\begingroup
\section{Empirical Evaluation}

\begin{table}[!hbtp]
\caption{Datasets used to empirically evaluate our method.}
\begin{adjustbox}{max width=\linewidth}
\begin{tabular}{@{}llcc@{}}
\toprule
\textbf{Dataset} & \textbf{Modality} & \textbf{Ground Truth Known} & \textbf{Results Section} \\ \midrule
Synthetic        & Tabular           &   \cmark                     &          \ref{subsec:synthetic_exps}                     \\
ColoredMNIST     & Images            &     \cmark                         &       \ref{sec:cmnist}                       \\
CelebA           & Images            &   \cmark                           &              \ref{sec:celebA}                 \\
eICU             & Tabular           &    \xmark                          &        \ref{subsec:eICU_exps}  \\
Camelyon-17      & Images            &      \xmark                        &                    \ref{sec:camelyon}       \\ \bottomrule
\end{tabular}
\end{adjustbox}
\label{tab:datasets}
\end{table}

We empirically validate our method on five datasets, shown in Table \ref{tab:datasets}. First, we validate the \textit{correctness} of our method on three synthetic and semi-synthetic datasets where the ground truth shift(s) are known, and show that the baseline methods described below do not attain the correct attributions. Next, we demonstrate the \textit{utility} of our method on two real-world datasets. Each attribution experiment was run on a cluster using 4 cores from an Intel Xeon Gold 5218R Processor and 16 GB of memory. We present two case studies on real-world data here, and the remaining results can be found in Appendix Section \ref{app:add_results}.

\paragraph{Baselines.} On datasets with known ground truth shifts (see Appendix Sections \ref{subsec:synthetic_exps}, \ref{sec:cmnist}) we evaluate the following baselines: 
\begin{enumerate}[label=\arabic*., leftmargin=*,topsep=0pt]
    \item[(a)] \textbf{Misspecified or unknown causal graph.} When the causal graph is unknown, the user may create a causal graph based on intuition or causal discovery methods \cite{glymour2019review}. Here, we evaluate two simple mis-specified candidate shift sets. First, we evaluate the candidate shifts corresponding to all marginals (i.e. $\cand = \{\cD_{X_1}, \cdots,\cD_{X_i}, \cdots\}_{i=1,\cdots,|V|}$, which is similar to the method in \citet{wu2021performance}. Second, we evaluate the candidate shifts corresponding to a fully connected graph (i.e. $\cand = \{\cD_{X_1 | V \setminus X_1}, \cdots,\cD_{X_i | V \setminus X_i}, \cdots\}_{i=1,\cdots,|V|}$).
    \item[(b)] \textbf{KL-based attribution.} We evaluate the joint method from \citet{budhathoki2021distribution}.
    \item[(c)] \textbf{SHAP-based attribution.} We test a two-stage heuristic devised for this problem setting. First, we use a conditional independence test \cite{zhang2012kernel} to find the distributions that are significantly different between source and target. Then, we run Kernel SHAP \cite{lundberg2017unified} on all samples in the target domain, taking the mean absolute value of the feature importance only for features that have significant shifts. To create attributions, we normalize these values to sum to the performance drop. Note that this method has several major flaws, namely that it can only attribute to shifts in input features to the model (and not system variables unused by the model), and cannot attribute to the \emph{distribution} generating the target variable.
\end{enumerate}

\begin{figure*}[htbp!]
\begin{subfigure}[t]{.51\linewidth}
  \includegraphics[width=.99\linewidth]{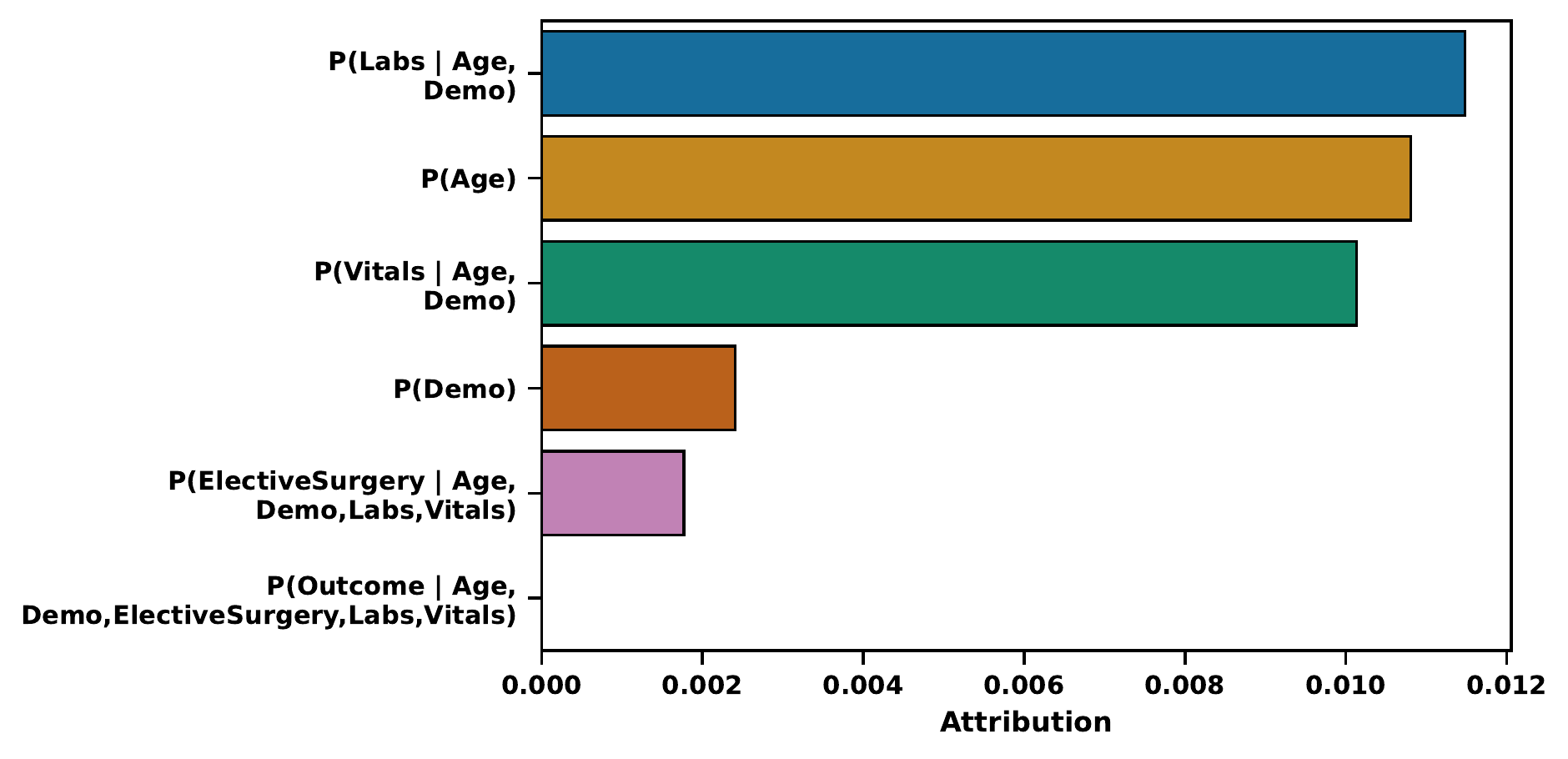}
  \caption{Attribution before data collection}
  \label{fig:xgb_eicu_brier_a}
\end{subfigure}%
\begin{subfigure}[t]{.38\linewidth}
  \centering
  \includegraphics[width=.98\linewidth]{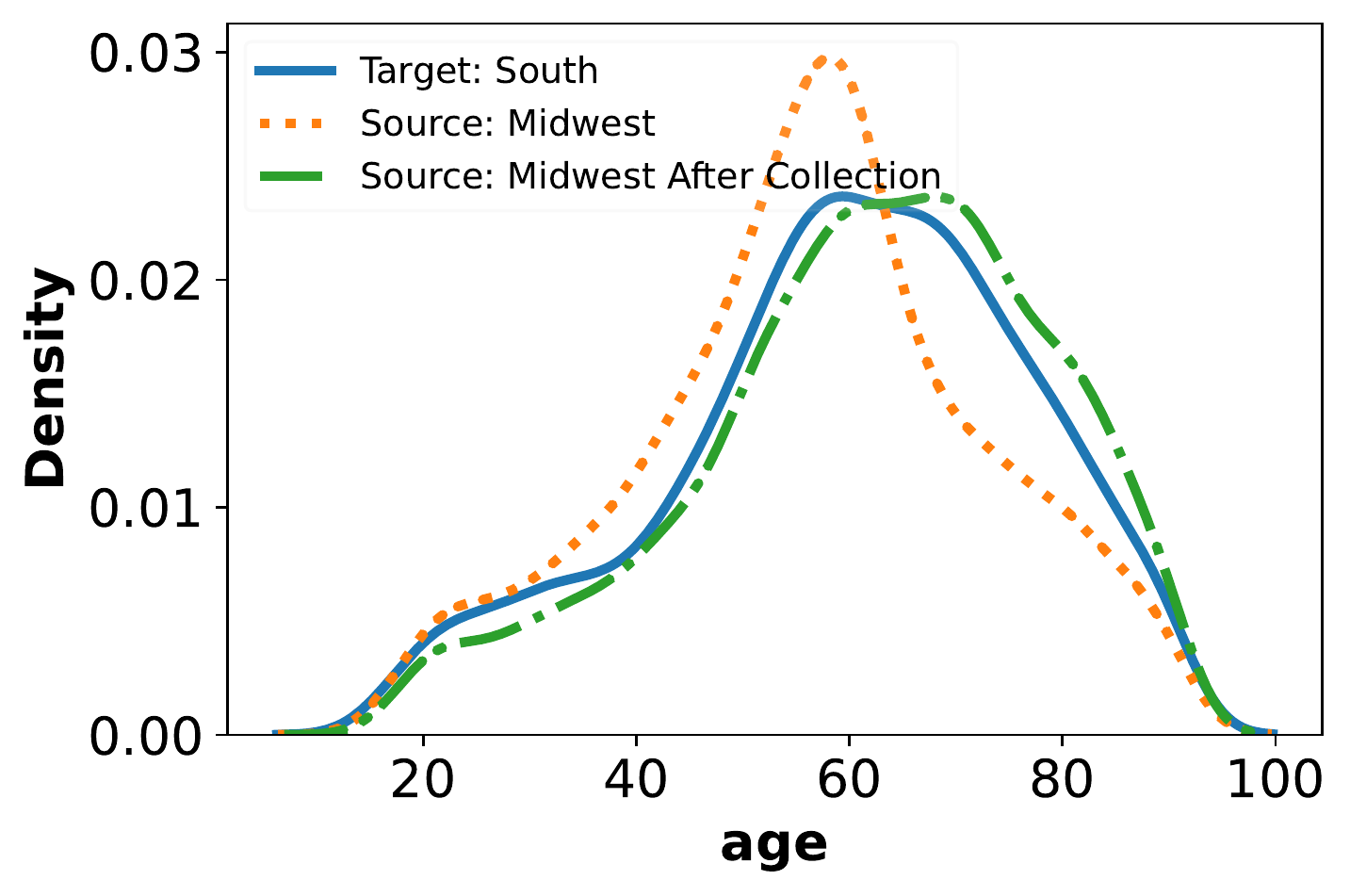}
  \caption{Age distributions}
  \label{fig:xgb_eicu_brier_b}
\end{subfigure}

\begin{subfigure}[b]{.5\linewidth}
  \includegraphics[width=.99\linewidth]{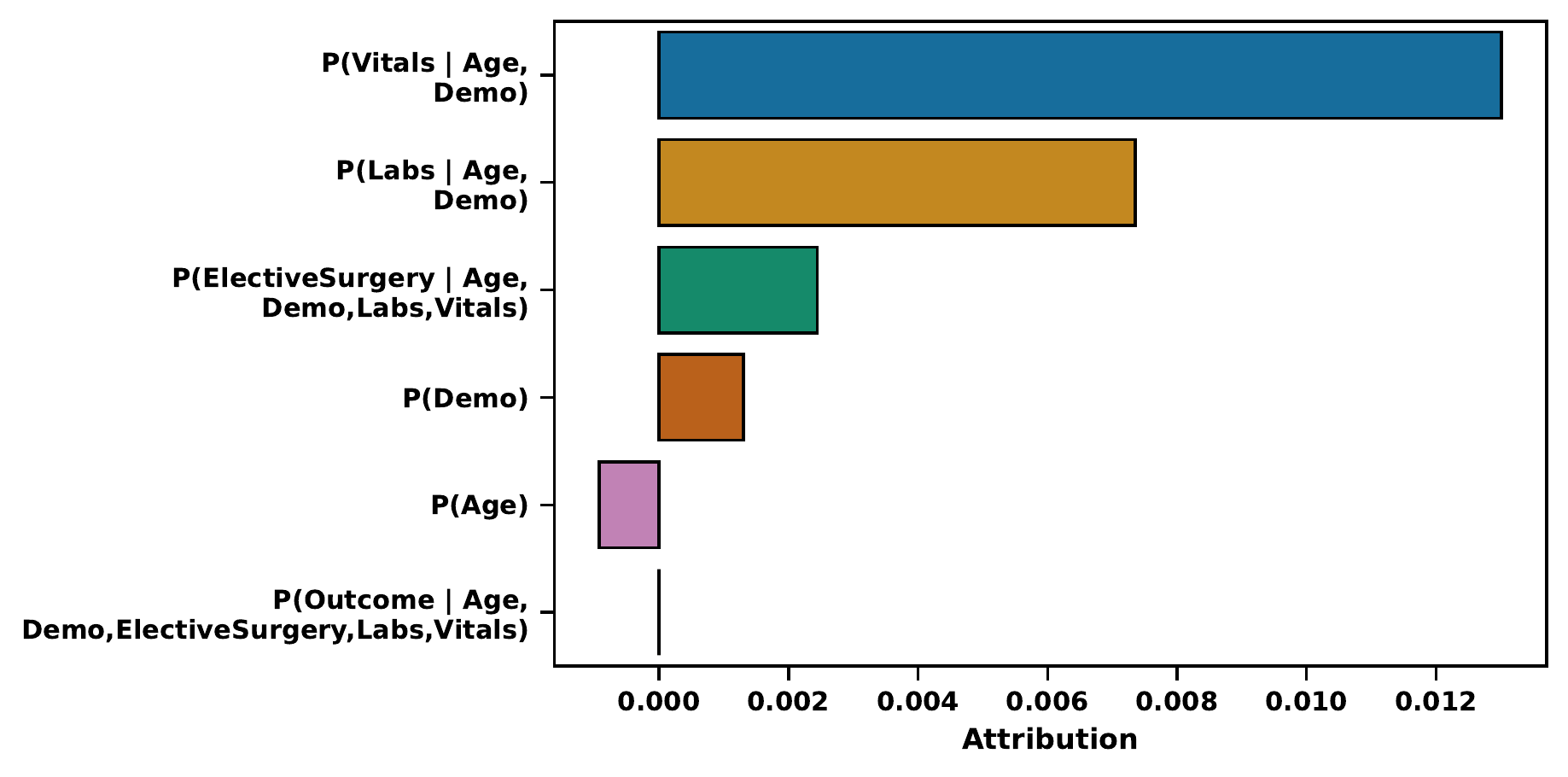}
  \caption{Attribution after data collection}
  \label{fig:xgb_eicu_brier_c}
\end{subfigure}
\begin{subfigure}[b]{.5\linewidth}
\resizebox{\textwidth}{!}{
  \begin{tabular}{@{}lrrr@{}}
\toprule
                                  & Perf($\mathcal{D}^{source})$ & Perf($\mathcal{D}^{target})$          & Perf Diff            \\ \midrule
Before data collection            & 0.0424 & 0.0790          & 0.0366          \\
After data collection from source & 0.0473 & \textbf{0.0705} & \textbf{0.0232} \\ \hdashline
Retrain on target only              & 0.0552 & 0.0970          & 0.0417          \\
Retrain on source + target          & 0.0424 & 0.0776          & 0.0351          \\ \bottomrule
\end{tabular}
}
\vspace{3em}
\caption{Brier scores before and after additional data collection, along with two naive retraining baselines \label{fig:xgb_eicu_brier_d}}
\end{subfigure}
\caption{Attributing Brier score differences to candidate distributions on the eICU dataset for an \texttt{XGB} model trained on either (a) original or (c) age-balanced Midwest, and tested on the South domain.}
\label{fig:xgb_eicu_brier}
\end{figure*}

\subsection{Case Study: Mortality Prediction in the ICU} %

\label{subsec:eICU_exps}
\paragraph{Setup.} Clinical machine learning models are being increasingly deployed in the real-world in hospitals, laboratories, and Intensive Care Units (ICUs) \citep{sendak2020path}. However, prior work has shown that such machine learning models are not robust to distribution shifts, and frequently degrade in performance on distributions different than what is seen during training \citep{singh2022generalizability}. Here, we explore a simulated case study where a model which predicts mortality in the ICU is deployed in a different geographical region from where it is trained. We use data from the eICU Collaborative Research Database V2.0 \citep{pollard2018eicu}, which contains 200,859 de-identified ICU records for 208 hospitals across the United States. We simulate the deployment of a model trained on data from the Midwestern US (source) to the Southern US (target). We subset to 4 hospitals in each geography with the most number of samples. To mimic a realistic deployment scenario with limited sample size, we only observe 250 samples randomly selected from the target domain.

We learn an $\texttt{XGB}$ \cite{chen2016xgboost} model to predict mortality given vitals, labs, and demographics data in the source domain. We assume the causal graph in Figure \ref{fig:eicu_graph}, informed by prior work utilizing causal discovery on this dataset \citep{singh2022generalizability}. As prior work has shown limited performance drops for models in this setting \citep{zhang2021empirical}, we subsample older population in the source environment to create an additional semi-synthetic distribution shift.
We use our method to attribute the increase in Brier score from Midwest to South datasets.

\paragraph{Our method provides actionable attributions.}  First, we observe from our attributions (Figure \ref{fig:xgb_eicu_brier_a}) that shift in the age distribution is responsible for 29.5\% of the total shift (0.0108 of 0.0366). This confirms the validity of the attributions on a known semi-synthetic shift. Suppose that the practitioner decides to focus on mitigating the shift in age in order to improve target domain performance. To do so, they first plot the age distribution in the source and target environments (Figure \ref{fig:xgb_eicu_brier_b}), finding that the target domain has dramatically more older patients. Then, they choose to collect additional data from the older population in the source. Training a new model on this augmented dataset, they find that the target domain performance improves by 10.8\%, and the drop in performance is reduced by 36.6\% (Figure \ref{fig:xgb_eicu_brier_d}). In addition, this targeted mitigating action outperforms naively retraining the model on the merged datasets, or only on the target domain, due to the few target samples observed. Now that $\cD_{\text{Age}}$ is no longer a significant factor in the performance drop across domains (Figure \ref{fig:xgb_eicu_brier_c}), the practitioner may next turn their attention to mitigating shifts in more impactful conditional mechanisms such as $\cD_{\text{Vitals} | \text{Age, Demo}}$, using methods such as GAN data augmentation \citep{mariani2018bagan} or targeted importance weighting \cite{zhang2013domain}, but we leave such explorations to future work. 

\subsection{Case Study: Tumor Prediction from Camelyon17}
\label{sec:camelyon}

We evaluate our method on the Camelyon17 dataset \cite{bandi2018detection, koh2021wilds}, which consists of histopathology images from five hospitals, and the goal is to classify whether the central region contains any tumor tissue. We assume the causal setting (i.e. an $X \rightarrow Y$ causal graph)~\cite{bandi2018detection}, as labels are generated by pathologists from the image. Here, $X$ is a vector-valued node for the image, which we represent using static features extracted from an ImageNet-pretrained ResNet-18. We train linear models on these representations to predict $Y$ separately for each site. We use our method to attribute drops in accuracy of each model to each of the other four sites, to $P(X)$ and $P(Y|X)$.

\paragraph{Results.} In Figure \ref{fig:camelyon_heatmaps}, we show the attributions from our method for each distribution, as well as the total accuracy drop. We find that our method attributes most of the performance drop to covariate shift $P(X)$ as opposed to concept shift $P(Y|X$). This aligns with prior work showing that unsupervised domain adaptation methods improve domain robustness in this dataset~\cite{wiles2021fine, ginsberg2022learning}. Using this result, a  practitioner can apply targeted mitigating methods such as targeted data augmentation~\cite{gao2022outofdistribution} or domain-adversarial training~\citep{ganin2016domain}.

\begin{figure*}[!htbp]
\begin{subfigure}{0.33\textwidth}
  \centering 
  \includegraphics[width=0.95\linewidth]{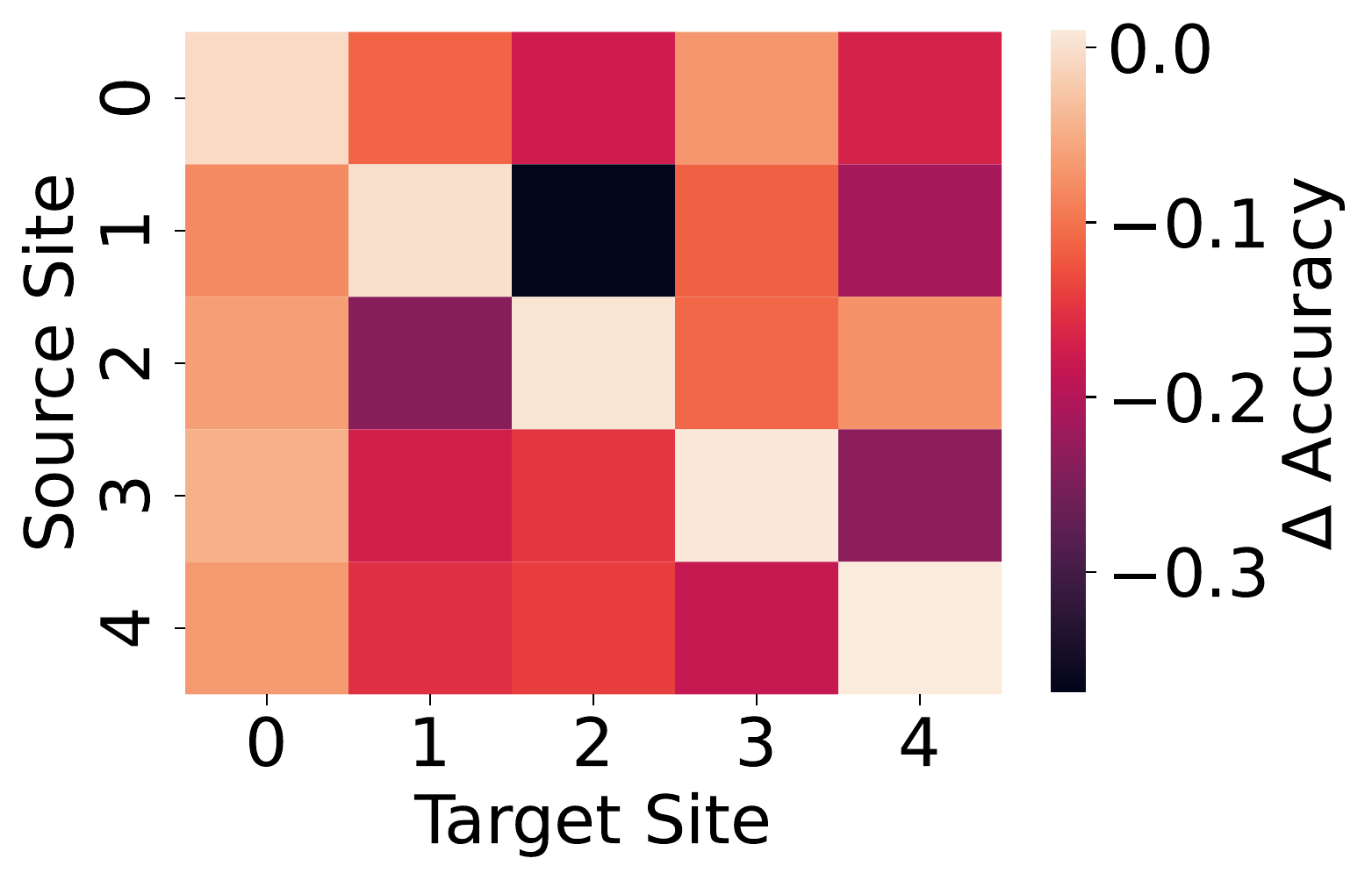}
    \caption{$\perf(\cD^{target}) - \perf(\cD^{source})$}
\end{subfigure}
\begin{subfigure}{0.33\textwidth}
  \centering 
  \includegraphics[width=0.95\linewidth]{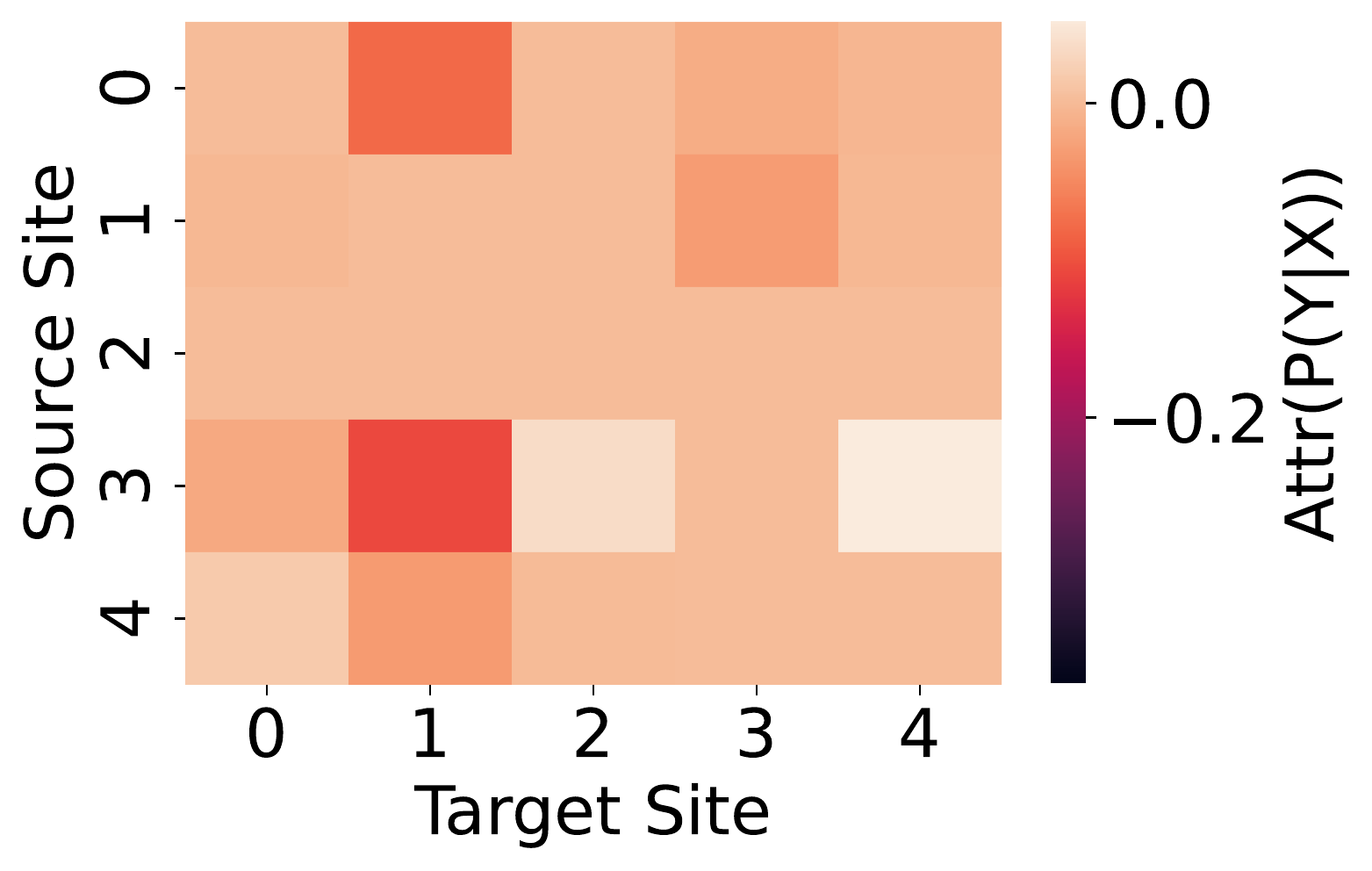}
    \caption{Attr($\cD_{Y|X}$)}
\end{subfigure}
\begin{subfigure}{0.33\textwidth}
  \centering 
  \includegraphics[width=0.95\linewidth]{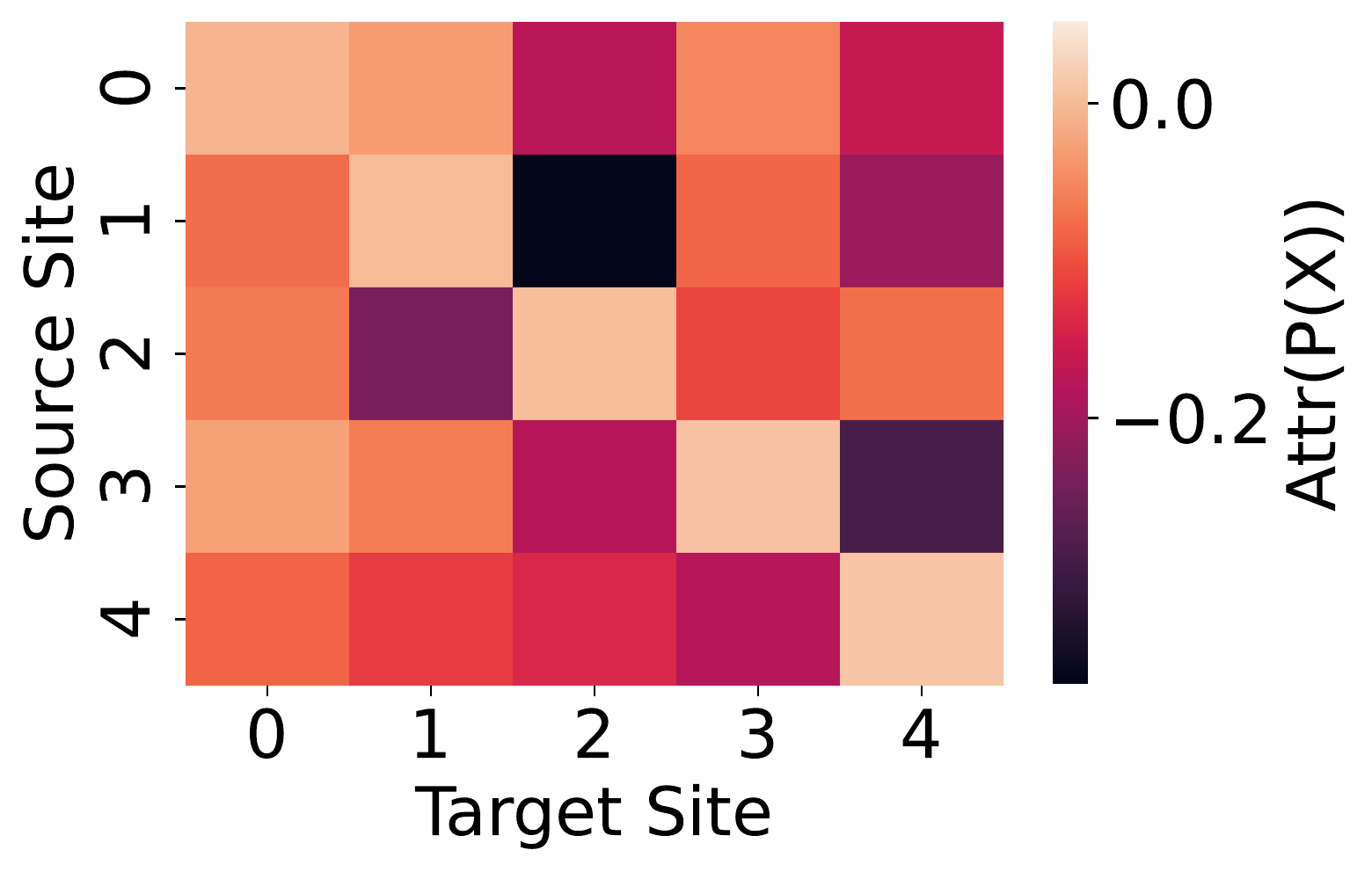}
    \caption{Attr($\cD_{X}$)}
\end{subfigure}
    \caption{Attributions made by our method to domain shifts in Camelyon-17, using accuracy as the metric. We show (a) the total change in performance, (b) our attribution to $P(Y|X)$, (c) our attribution to $P(X)$. }
    \label{fig:camelyon_heatmaps}
\end{figure*}

\section{Discussion}

We develop a method to attribute changes in performance of a model deployed on a different distribution from the training distribution. We assume that distribution shifts are induced due to interventions in the causal mechanisms which result in model performance changes. We use the knowledge of the causal graph to formulate a game theoretic attribution framework using Shapley values. The coalition members are mechanisms contributing to the change in model performance. We demonstrate the correctness and utility of our method on synthetic, semi-synthetic, and real-world data. 

\paragraph{Limitations and Future Work.} Our work assumes knowledge of the causal graph to obtain interpretable and succinct attributions. When the causal graph is unknown, methods in causal discovery \cite{glymour2019review} can produce a Markov equivalence class of causal graphs for tabular datasets, though these methods often have strict assumptions. While we may still be able to obtain reasonable attributions from a misspecified graph, we argue that such attributions may not be minimal. In addition, we observe some variance in the importance weighting estimates, which may potentially be remedied by using more advanced density estimation techniques (e.g. \citep{liu2021density}). 
We note that our experiments on the CelebA dataset are for demonstration purposes only, and do not advocate for deployment of such models. Similarly, while we demonstrate case studies on publicly available health data, our work is only a proof of concept, and we recommend further evaluation before practical deployment. Future work includes relaxing the assumption that all variables are observed, comparing strategies for mitigating conditional shifts, and extending the experiments to additional settings such as unsupervised learning and reinforcement learning.

\endgroup

\FloatBarrier

\section*{Acknowledgements}
This work was supported in part by a grant from Quanta Computing. We would like to thank Taylor Killian and three anonymous reviewers for their valuable feedback. HS acknowledges support from the National Science Foundation under NSF Award 1922658. HS would like to thank Rumi Chunara and Vishwali Mhasawade for helpful discussions leading up to this work.

\bibliography{references.bib}
\bibliographystyle{icml2023}

\clearpage
\appendix
\onecolumn

\section{Derivation of Importance Weights}\label{app:iw_deriv}
Let $D$ be a binary random variable, such that when $D=1, X\sim \ptar(X)$, and when $D=0, X\sim \psrc(X)$. Suppose $d = \cD_{X_i | \pa(X_i)}$, then, for a particular value $(x, y)$: 
\begin{align*}
\ptar_d((x,y)) &:=\pr(X_i = x  | \pa(X_i) = \pa(x_i), D = 1) \\
&= \frac{\pr(D = 1 , \pa(X_i) = x_i | X_i = x_i) \cdot \pr(X_i = x_i)} {\pr(D = 1, \pa(X_i) = x_i)} \\
&= \frac{\pr(D = 1 | \pa(X_i) = x_i,  X_i = x_i) \cdot \pr(X_i = x_i, \pa(X_i) = X_i)} {\pr(D = 1 | \pa(X_i) = x_i) \cdot \pr(\pa(X_i) = x_i) } 
\end{align*}

Then, 
\begin{align*}
    w_d &= \frac{\ptar_d((x,y))}{\psrc_d((x,y))} \\
    &=\frac{\pr(D = 0 | \pa(X_i) = \pa(x_i))}{\pr(D = 1| \pa(X_i) = \pa(x_i))} \cdot \frac{\pr(D = 1 | X_i = x_i, \pa(X_i) = \pa(x_i))}{\pr(D = 0 | X_i = x_i, \pa(X_i) = \pa(x_i))} \\
    &= \frac{1 - \pr(D = 1 | \pa(X_i) = \pa(x_i))}{\pr(D = 1| \pa(X_i) = \pa(x_i))} \cdot \frac{\pr(D = 1 | X_i = x_i, \pa(X_i) = \pa(x_i))}{1 - \pr(D = 1 | X_i = x_i, \pa(X_i) = \pa(x_i))}
\end{align*}

Thus, we learn a model to predict $D$ from $X_i$, and a model to predict $D$ from $[X_i; \pa(X_i)]$, on the concatenated dataset. In practice, we learn these models on a 75\% split of both the source and target data, and use the remaining 25\% for Shapley value computation, which only requires inference on the trained models. Therefore, an upper limit on the number of weight models required is $2|\cand|$, though in practice, this number is often smaller as several nodes may have the same parents.

In the case where $X_i$ is a root node, the expression becomes:
\begin{align*}
    w_d &= \frac{1 - \pr(D = 1)}{\pr(D = 1)} \cdot \frac{\pr(D = 1 | X_i = x_i)}{1 - \pr(D = 1 | X_i = x_i)}
\end{align*}
Where we simply compute $P(D=1)$ as the relative size of the provided source and target datasets.

\clearpage

\section{Proof of Properties}\label{app:proof_property}

\paragraph{Property 1. (Efficiency)} $\displaystyle\sum_{d \in \cand} \attr(d) = \val(\cand) = \perf(\ptar) - \perf(\psrc)$ %

By the efficiency property of Shapley values \citep{shapley1953value}, we know that the sum of Shapley values equal the value of the all-player coalition. Thus, we distribute the total performance change due to the shift from source to target distribution to the shifts in causal mechanisms in the candidate set.

\paragraph{Property 2.1. (Null Player)} $\psrc_d = \ptar_d \implies \attr(d) = 0$.

\paragraph{Property 2.2. (Relevance)}  Consider a mechanism $d$. If $\perf(\subcand \cup \{d\}) = \perf(\subcand)$ for all $\subcand \subseteq \cand \setminus d$, then ${\attr(d) = 0}$.

We can verify that our method gives zero attribution to distributions that do not shift between the source and target, and distribution shifts which do not impact model performance. First, we observe that in both cases, $\val(\widetilde{\cD}) = \val(\widetilde{\cD} \cup \{d\})$. For Property 2.1, this is because $\widetilde{\cD} = \widetilde{\cD} \cup \{d\}$ for any $\widetilde{\cD} \subseteq \cand$ since the factor corresponding to $d$ remains the same between source and target even when it is allowed to change as part of the coalition. For Property 2.2, this is clear from Eq. \ref{eq:val_def}. By definition of Shapley value in Eq. \ref{eq:shapley}, $\attr(d) = 0$.

\paragraph{Property 3. (Attribution Symmetry)} Let $\attr_{\cD_1, \cD_2}(d)$ denote the attribution to some mechanism $d$ when $\cD_1 = \psrc$ and $\cD_2 = \ptar$. Then, $\attr_{\cD_1, \cD_2}(d) = -\attr_{\cD_2, \cD_1}(d) \ \forall d \in \cand$.

We overload $\perf_{src\rightarrow tar}(\subcand)$ for some coalition $\subcand$ to denote $\perf(\widetilde{\cD})$ where $\widetilde{\cD}$ is given by Equation \ref{eq:src_d}. Analogously, we denote $\perf_{tar\rightarrow src}(\subcand)$ to be $\perf(\widetilde{\cD}')$ when $\widetilde{\cD}'$ is given by

$$\widetilde{\cD}' = \left(\prod_{i: \cD_{X_i | \pa(X_i)}\in \subcand} \psrc_{X_i | \pa(X_i)} \right) \left( \prod_{i: \cD_{X_i | \pa(X_i)}\not\in \subcand} \ptar_{X_i | \pa(X_i)} \right)$$

Note that $\perf_{src\rightarrow tar}(\subcand)$ = $\perf_{tar\rightarrow src}(\cand \setminus \subcand)$ for all $\subcand \subseteq \cand$.

We can use Equation \ref{eq:value} to rewrite Equation \ref{eq:shapley} as:

\begin{align*}
    \attr_{\cD_1, \cD_2}(d) &= \frac{1}{|\cand|} \sum_{\subcand\subseteq \cand \setminus \{d\}} \binom{|\cand| - 1}{|\subcand|}^{-1} \left( \perf_{src\rightarrow tar}(\subcand \cup \{d\}) - \perf_{src\rightarrow tar}(\subcand)\right) \\
    &=\frac{-1}{|\cand|} \sum_{\subcand\subseteq \cand \setminus \{d\}} \binom{|\cand| - 1}{|\subcand|}^{-1} \left(\perf_{tar\rightarrow src}(\cand \setminus \subcand) - \perf_{tar\rightarrow src}(\cand \setminus (\subcand \cup \{d\}))  \right) \\
    &=\frac{-1}{|\cand|} \sum_{\subcand'\subseteq \cand \setminus \{d\}} \binom{|\cand| - 1}{|\subcand'|}^{-1} \left(\perf_{tar\rightarrow src}(\subcand' \cup \{d\}) - \perf_{tar\rightarrow src}(\subcand')  \right) \\[0.5em]
    &= -  \attr_{\cD_2, \cD_1}(d)
\end{align*}

\clearpage

\section{Shapley Values for A Synthetic Setting}\label{app:synthetic}

\renewcommand\thefigure{\thesection.\arabic{figure}}
\renewcommand\thetable{\thesection.\arabic{table}} 
\setcounter{figure}{0}    
\setcounter{table}{0}  

\subsection{Derivation}

Suppose that we have the following data generating process for the source environment:
\begin{align*}
    X &\sim \mathcal{N}(\mu_1, \sigma_X^2) \\
    Y &\sim \theta_1 X + \mathcal{N} (0, \sigma_Y^2)
\end{align*}
And for the target environment:
\begin{align*}
    X &\sim \mathcal{N}(\mu_2, \sigma_X^2) \\
    Y &\sim \theta_2 X + \mathcal{N} (0, \sigma_Y^2)
\end{align*}
The model that we are investigating is $\hat{Y} = f(X) = \phi X$, and $l((x, y), f) = (y - f(x))^2$. Then,
\begin{align*}
    \perf(\psrc) &= \E_{(x,y) \sim \psrc}[l((x, y), f)] \\
    &= \E_{(x,y) \sim \psrc}[\left(\theta_1 X + \mathcal{N}(0, \sigma_Y^2) - \phi X\right)^2] \\
    &= \E_{(x,y) \sim \psrc}[\left(\mathcal{N}((\theta_1 -\phi)\mu_1, (\theta_1 - \phi)^2 \sigma_X^2) + \mathcal{N}(0, \sigma_Y^2)\right)^2] \\
    &= \E_{(x,y) \sim \psrc}[\left(\mathcal{N}((\theta_1 - \phi) \mu_1,  (\theta_1 - \phi)^2\sigma_X^2 + \sigma_Y^2)\right)^2] \\
    &= (\theta_1 - \phi)^2\sigma_X^2 + \sigma_Y^2 + (\theta_1 - \phi)^2 \mu_1^2 \\[1em]
    \perf(\ptar) &= \E_{(x,y) \sim \ptar}[l((x, y), f)] \\
    &= (\theta_2 - \phi)^2\sigma_X^2 + \sigma_Y^2 + (\theta_2 - \phi)^2 \mu_2^2 \\[1em]
    \Delta &= \perf(\ptar) -  \perf(\psrc)\\
    &=\sigma_X^2 ((\theta_2 - \phi)^2 - (\theta_1 - \phi)^2)+ (\theta_2 - \phi)^2 \mu_2^2 - (\theta_1 - \phi)^2 \mu_1^2 \\
    &= \val(\cand) \\[1em]
    \val(\{\cD_X\}) &= (\theta_1 - \phi)^2 (\mu_2^2 - \mu_1^2)  \quad &(\theta_2:=\theta_1) \\
    \val(\{\cD_{Y|X}\}) &= (\sigma_X^2 + \mu_1^2) ((\theta_2 - \phi)^2 - (\theta_1 - \phi)^2)  \quad &(\mu_2:=\mu_1)  \\[1em]
    \attr(\cD_X) &= \frac{1}{2}\left( \val(\cand) - \val(\{\cD_{Y|X}\}) + \val(\{\cD_X\}) - \val(\{\})  \right)\\
    &=  \frac{1}{2}\left( (\theta_2 - \phi)^2 (\mu_2^2 - \mu_1^2) + (\theta_1 - \phi)^2 (\mu_2^2 - \mu_1^2) \right) \\
    &= (\frac{1}{2}\mu_2^2 - \frac{1}{2}\mu_1^2) ((\theta_2 - \phi)^2 + (\theta_1 - \phi)^2)\\[1em]
    \attr(\cD_{Y|X}) &= \frac{1}{2}\left( \val(\cand) - \val(\{\cD_X\}) + \val(\{\cD_{Y|X}\}) - \val(\{\}) \right)\\ 
    &= \frac{1}{2}\left( (\sigma_X^2 + \mu_2^2) ((\theta_2 - \phi)^2 - (\theta_1 - \phi)^2) + (\sigma_X^2 + \mu_1^2) ((\theta_2 - \phi)^2 - (\theta_1 - \phi)^2)\right) \\
    &= (\sigma_X^2 + \frac{1}{2} \mu_1^2 + \frac{1}{2} \mu_2^2) ((\theta_2 - \phi)^2 - (\theta_1 - \phi)^2) 
\end{align*}

Note that $\attr(\cD_X) + \attr(\cD_{Y|X})   = \Delta$.

Using the method proposed by \citet{budhathoki2021distribution}, we get that:
\begin{align*}
D(\tilde{P}_X || P_X) &= \frac{(\mu_2 - \mu_1)^2}{2\sigma_X^2} \\
D(\tilde{P}_{Y|X} || P_{Y|X}) &= \E_{X\sim \tilde{P}_X}[D(\tilde{P}_{Y|X=x} || P_{Y|X=x}) ] \\
&= \E_{X\sim \tilde{P}_X} \left[ \frac{((\theta_2 - \theta_1) X )^2}{2\sigma_Y^2} \right] = \frac{(\theta_2 - \theta_1)^2}{2 \sigma_Y^2} (\sigma_X^2 + \mu_2^2)
\end{align*}

\subsection{Experiments}
Now, we verify the correctness of our method by conducting a simulation of this setting, using $\mu_1 = 0$, $\theta_1 = 1$, $\sigma_X^2 = 0.5$, $\sigma_Y^2 = 0.25$, $\phi = 0.9$, and varying $\mu_2$ (the level of covariate shift), and $\theta_2$ (the level of concept drift). We generate $10,000$ samples from the source environment, and, for each setting of $\mu_2$ and $\theta_2$, we generate $10,000$ samples from the corresponding target environment. We then apply our method to attribute shifts to $\{\cD_X, \cD_{Y|X}\}$, using \texttt{XGB} to estimate importance weights. We also apply the joint method in \citet{budhathoki2021distribution}.

\begin{figure*}[h]
\centering
\begin{subfigure}{.49\linewidth}
  \centering
  \includegraphics[width=.9\linewidth]{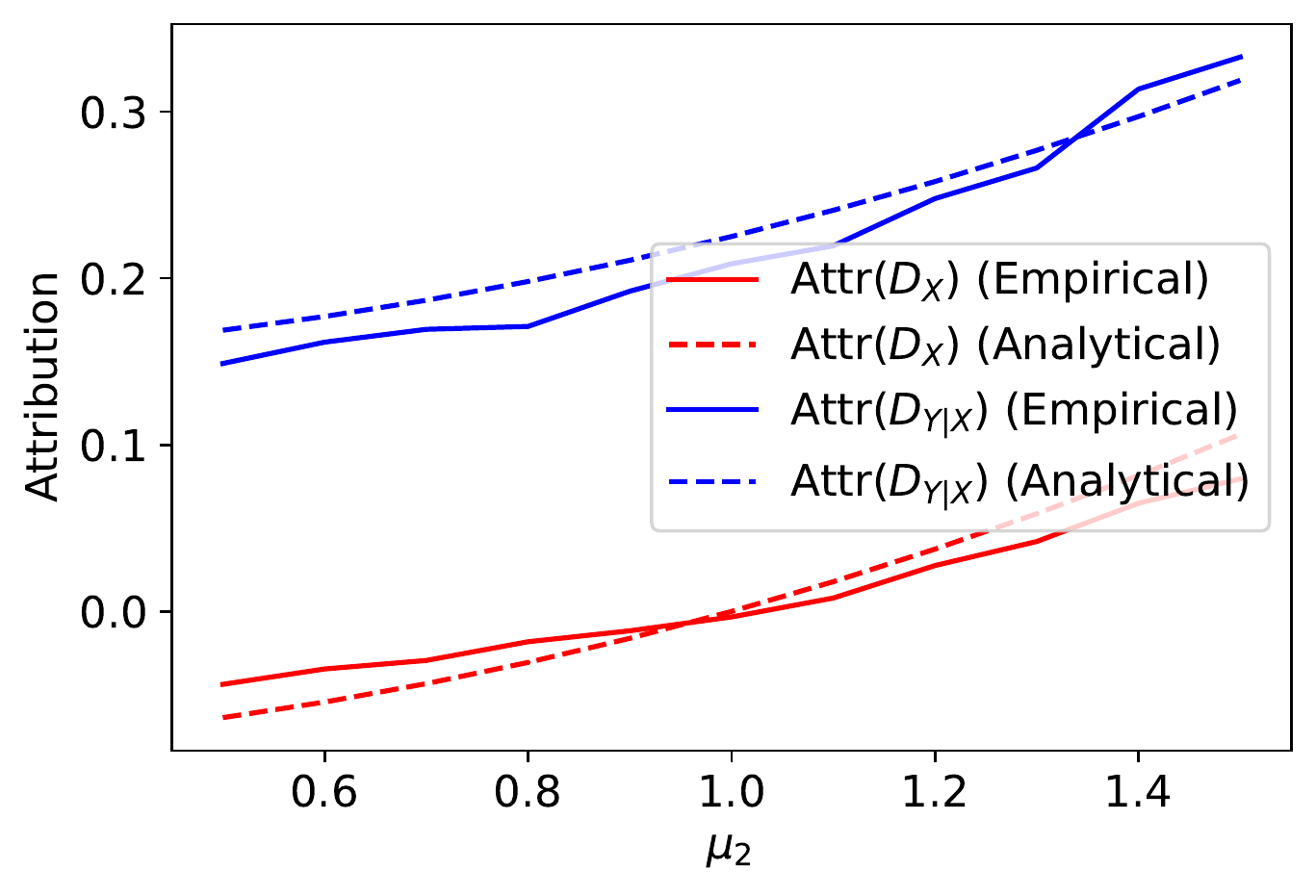}
  \caption{Our method; Fix $\theta_2=1.3$ and vary $\mu_2$.}
\end{subfigure}%
\hfill
\begin{subfigure}{.49\linewidth}
  \centering
  \includegraphics[width=.9\linewidth]{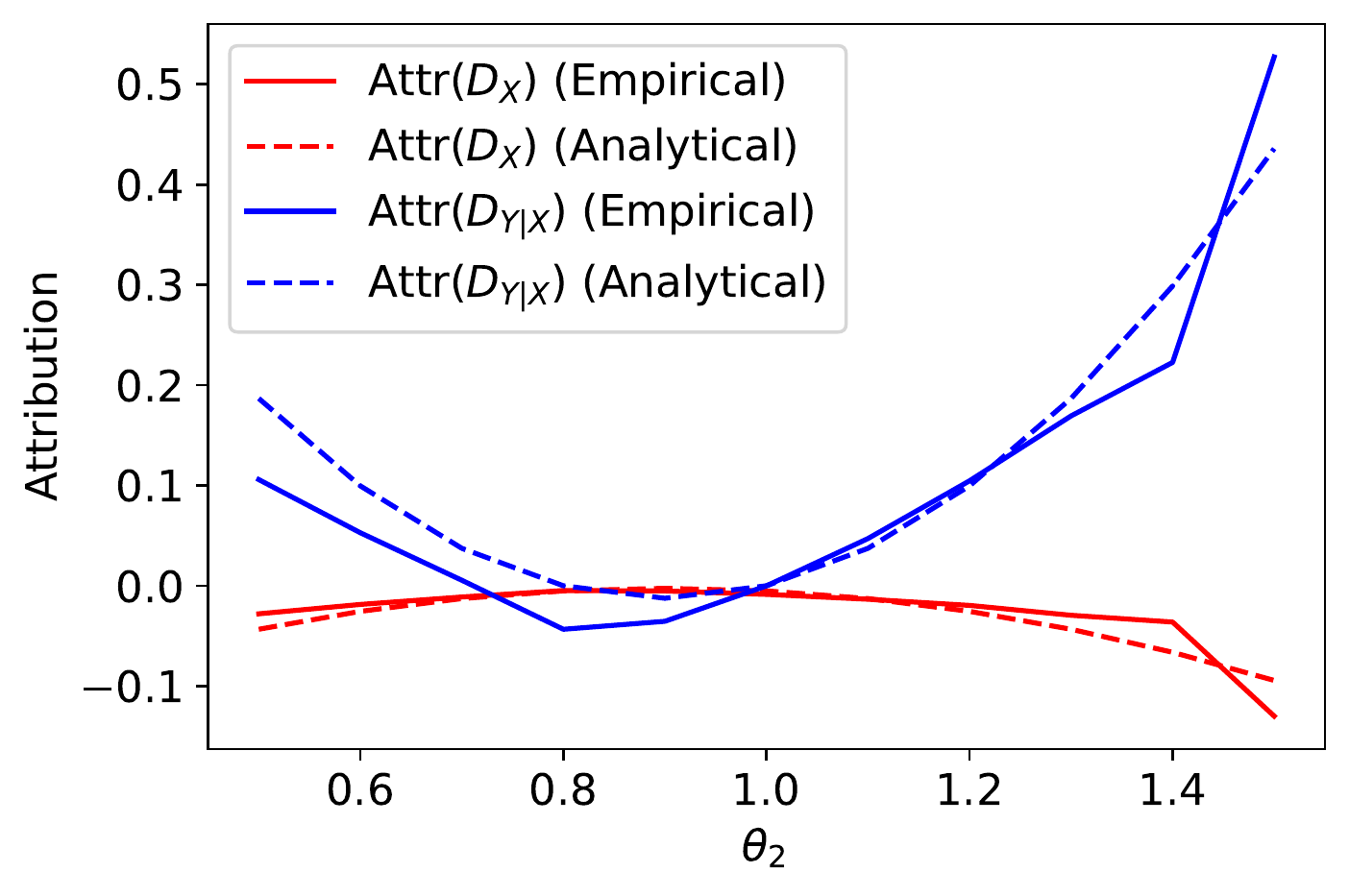}
  \caption{Our method; Fix $\mu_2=0.7$ and vary $\theta_2$.}
\end{subfigure}
\begin{subfigure}{.49\linewidth}
  \centering
  \includegraphics[width=.9\linewidth]{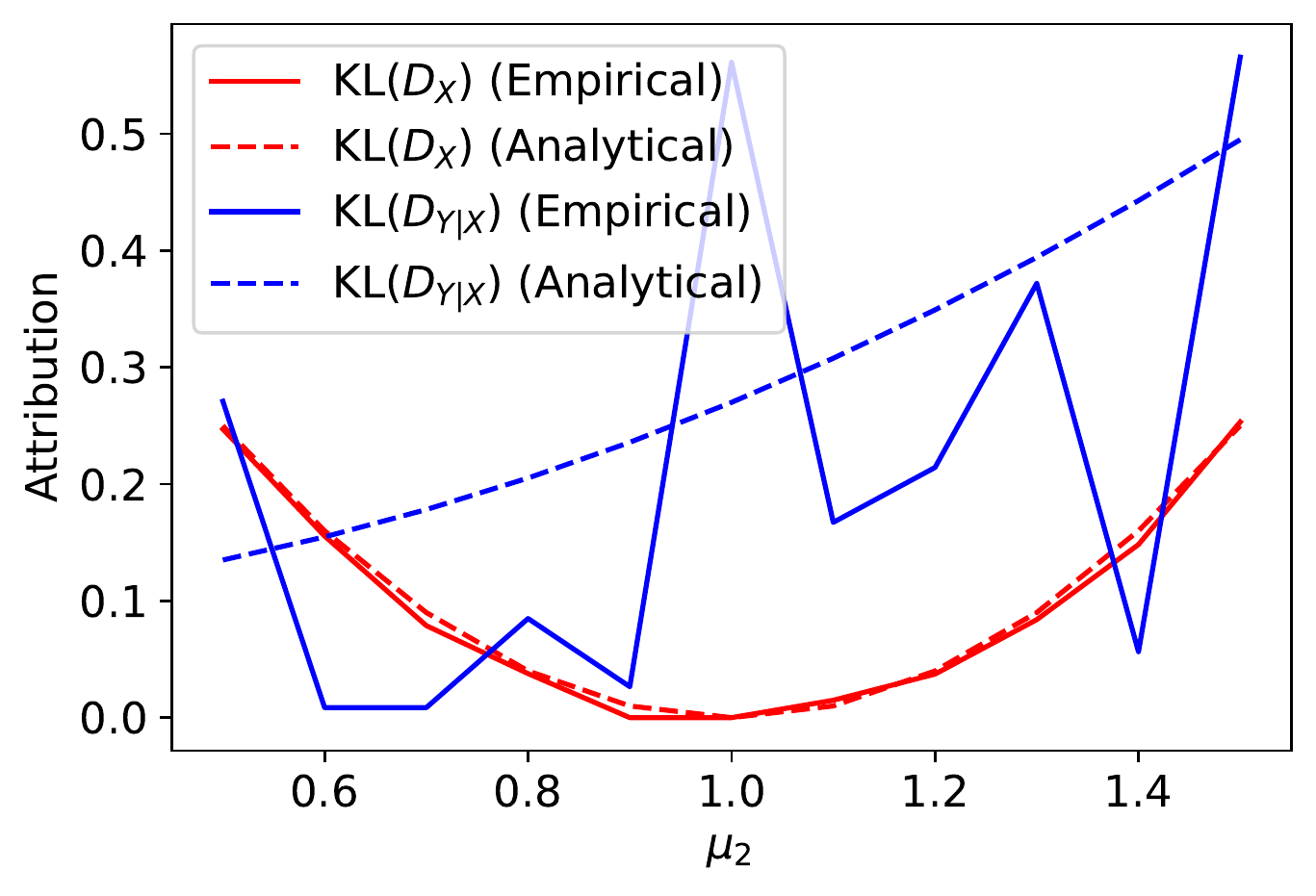}
  \caption{Joint method from \citet{budhathoki2021distribution}; Fix $\theta_2=1.3$ and vary $\mu_2$.}
\end{subfigure}%
\hfill
\begin{subfigure}{.49\linewidth}
  \centering
  \includegraphics[width=.9\linewidth]{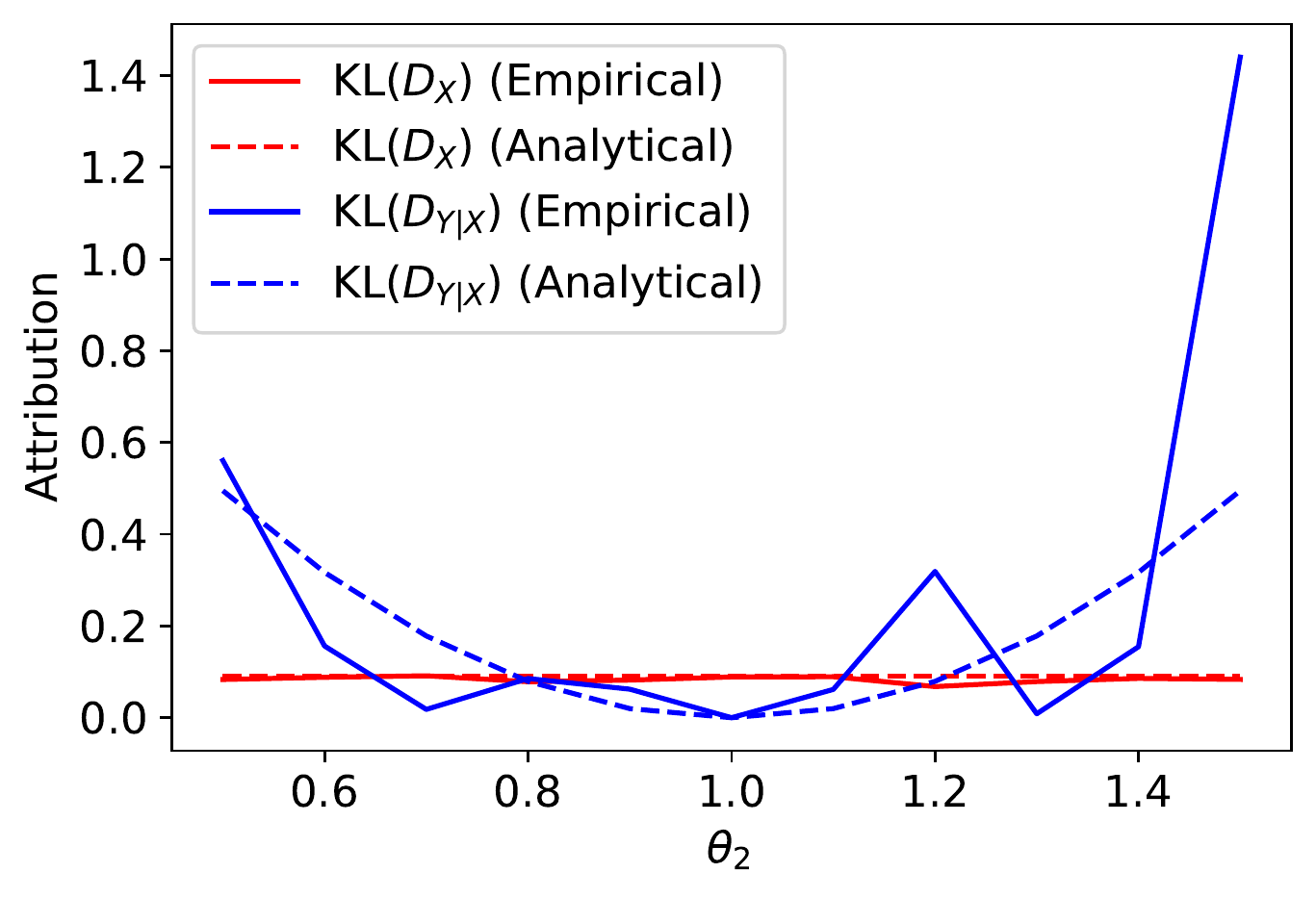}
  \caption{Joint method from \citet{budhathoki2021distribution}; Fix $\mu_2=0.7$ and vary $\theta_2$.}
\end{subfigure}%
\caption{Mean squared error differences attributed by our model and \citet{budhathoki2021distribution} in the synthetic setting described in Appendix \ref{app:synthetic}}
\label{fig:app_synthetic_compr}
\end{figure*}

In Figure \ref{fig:app_synthetic_compr}, we compare our attributions with the baseline, when both covariate and concept drift are present. We find that for our method, the empirical results match with the previously derived analytical expressions, where any deviations can be attributed to variance in the importance weight computations. For \citet{budhathoki2021distribution}, we find that there appears to be very high variance in the attribution the attribution to $\cD_{Y|X}$, which is likely a product of the nearest-neighbors KL estimator \cite{wang2009divergence} used in their work.

\begin{figure*}[h]
\centering
\begin{subfigure}{.49\linewidth}
  \centering
  \includegraphics[width=.9\linewidth]{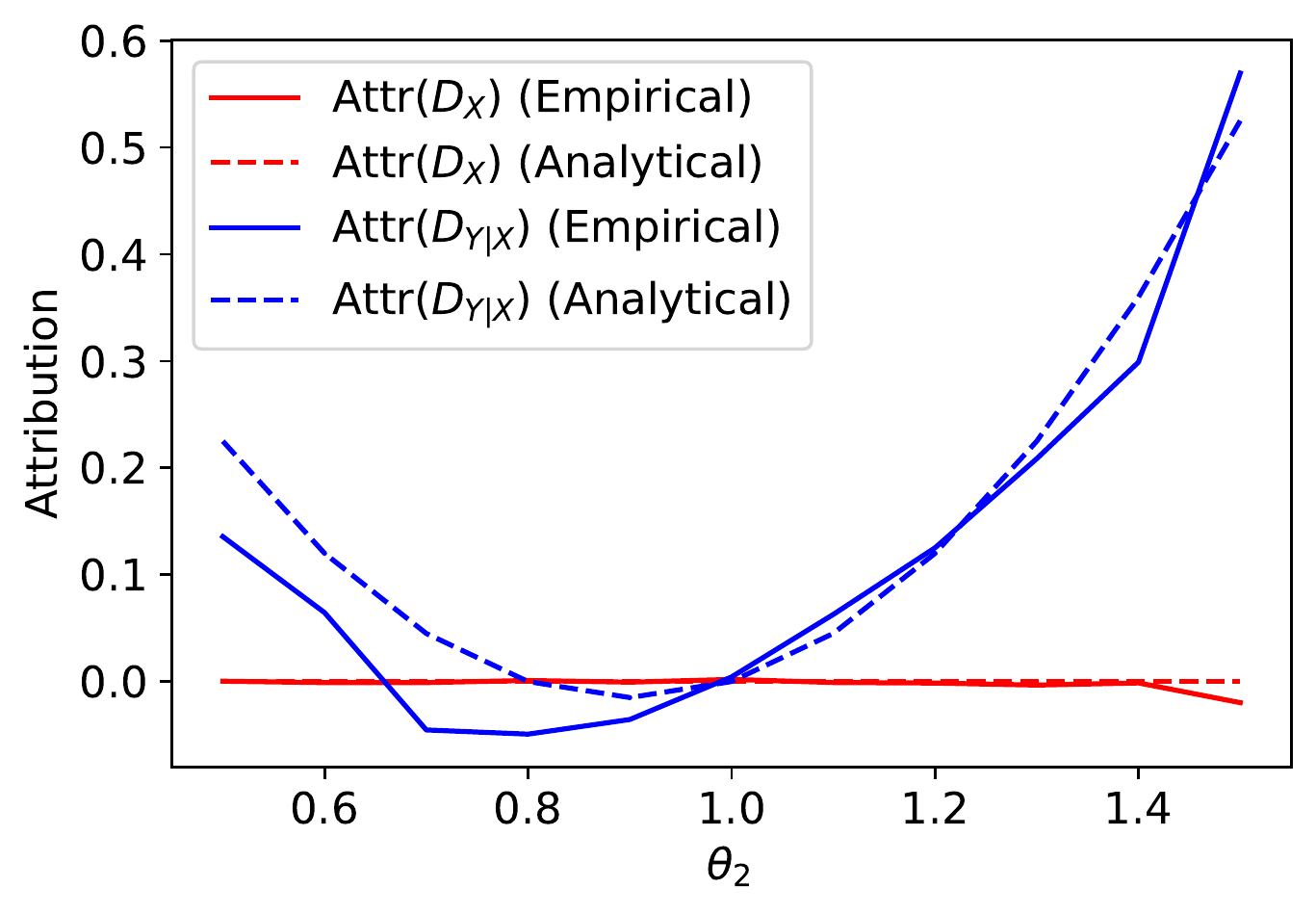}
  \caption{Our method; Fix $\mu_2 = 1$ and vary $\theta_2$, with $\cand = \{\cD_{X}, \cD_{Y|X}\}$, the actual causal graph}
\end{subfigure}%
\hfill
\begin{subfigure}{.49\linewidth}
  \centering
  \includegraphics[width=.9\linewidth]{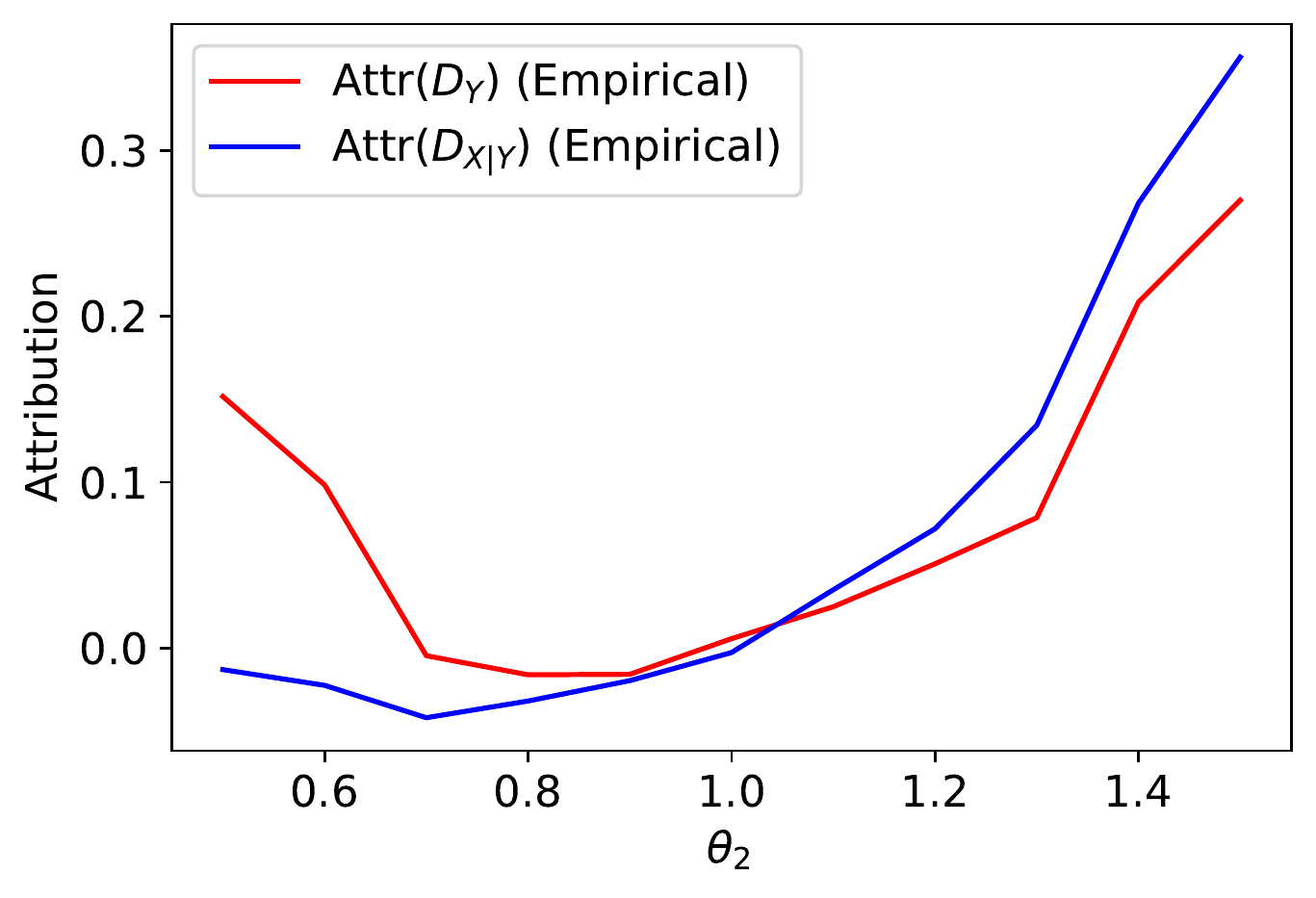}
  \caption{Our method; Fix $\mu_2 = 1$ and vary $\theta_2$, with $\cand = \{\cD_{Y}, \cD_{X|Y}\}$, a mis-specified causal graph}
\end{subfigure}
\caption{Mean squared error differences attributed by our model when there is only concept drift, for the actual causal graph (a), and a mis-specified causal graph (b).}
\label{fig:app_synthetic_misspec}
\end{figure*}

In Figure \ref{fig:app_synthetic_misspec}, we explore the case where we have a misspecified causal graph. Specifically, we examine the case where only concept drift is present, for the actual graphical model ($\cand = \{\cD_{X}, \cD_{Y|X}\}$), and for a misspecified graphical model ($\cand = \{\cD_{Y}, \cD_{X|Y}\}$). We find that using the mechanisms from the true data generating process results in a \textit{minimal} attribution (i.e. $\attr(\cD_X) = 0$), whereas the the misspecified causal graph gives non-zero attribution to both distributions.

\FloatBarrier
\section{Additional Experimental Results}
\renewcommand\thefigure{\thesection.\arabic{figure}}
\renewcommand\thetable{\thesection.\arabic{table}} 
\setcounter{figure}{0}    
\setcounter{table}{0}  
\label{app:add_results}

\subsection{Synthetic Data}
\label{app:add_results_syn}
\label{subsec:synthetic_exps}
\begin{figure}[htbp!]
  \centering
  \includegraphics[width=.3\linewidth]{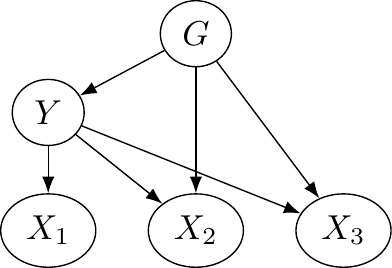}
  \caption{Causal Graph for Synthetic data}
  \label{fig:g1}
\end{figure}

\paragraph{Setup.}
We generate a synthetic binary classification dataset with five variables according to the following data generating process, corresponding to the causal graph shown in Figure \ref{fig:g1}. Here, $\xi_p: \{0, 1\} \rightarrow \{0, 1\}$ is a function that randomly flips the input with probability $p$.

\begin{align*}
    G &\sim Ber(0.5), \qquad
    X_2 = \mathcal{N}(\xi_{0.25} (Y) + G, 1)\\  Y &= \xi_{q}(G),   \qquad
    X_1 = \mathcal{N}(\omega \xi_{0.25} (Y), 1)  \\
    X_3 &= \mathcal{N}(\xi_{0.25} (Y) + \mu G, 1) 
\end{align*}

Where $q, \omega$ and $\mu$ are parameters of the data generating process. Here, $G$ represents a spurious correlation \citep{aubin2021linear, arjovsky2019invariant} that is highly correlated with $Y$, and is easily inferred from $(X_2, X_3$). 
By selecting a large value for $q$ (the spurious correlation strength) on the source environment, we can create a dataset where models rely more heavily on using $X_2$ and $X_3$ to infer $G$ and then $Y$, instead of inferring $\xi_{0.25} (Y)$ across the three features to estimate $Y$ directly.

 In the source environment, we set $q = 0.9, \omega = 1$ and $\mu = 3$. We generate 20,000 samples using these parameters, and train logistic regression (\texttt{LR}) models on ($X_1, X_2, X_3$) to predict $Y$, using 3-fold cross-validation to select the best model. We attribute performance changes for this model using the proposed method. We explore four data settings for the target environment:

\begin{enumerate}[label=\arabic*., leftmargin=*,noitemsep,topsep=0pt]
    \item[(a)] Label Shift: Vary $q \in [0, 1]$. Keep $\omega$ and $\mu$ at their source values. Only $P(Y|G)$ changes. This represents a label shift for the model across domains (which does not have access to $G$).
    \item[(b)] Covariate Shift: Vary $\mu \in [0, 5]$. Keep $q$ and $\omega$ at their source values. Only $P(X_3|G, Y)$ changes across domains.
    \item[(c)] Combined Shift 1: Set $\omega = 0$ in the target environment and vary $q \in [0, 1]$. Keep $\mu$ at its source value. Both $P(X_1|Y)$ and $P(Y|G)$ change across domains, but the shift should be largely attributed to $P(Y|G)$ as the model relies on this correlation much more than $X_1$. 
    \item[(d)] Combined Shift 2: Set $\mu = -1$ in the target environment. Further, vary $q \in [0, 1]$. Keep $\omega$ at its source value. Both $P(X_3|Y)$ and $P(Y|G)$ change across domains, but their specific contribution to model performance degradation is not known exactly.
\end{enumerate}

We use our method to explain performance changes in accuracy and Brier score for each model on target environments generated within each setting (with $n = 20,000$), computing density ratios using \texttt{XGB} \cite{chen2016xgboost} models. Note that the causal graph shown in Figure \ref{fig:g1} implies five potential distribution in the candidate set: $\cand=\{\cD_G, \cD_{Y|G}, \cD_{X_1|Y}, \cD_{X_2|G, Y}, \cD_{X_3| G, Y}\}$.

\begin{figure*}[htbp!]
\begin{subfigure}{.42\linewidth}
  \centering
  \includegraphics[width=.94\linewidth]{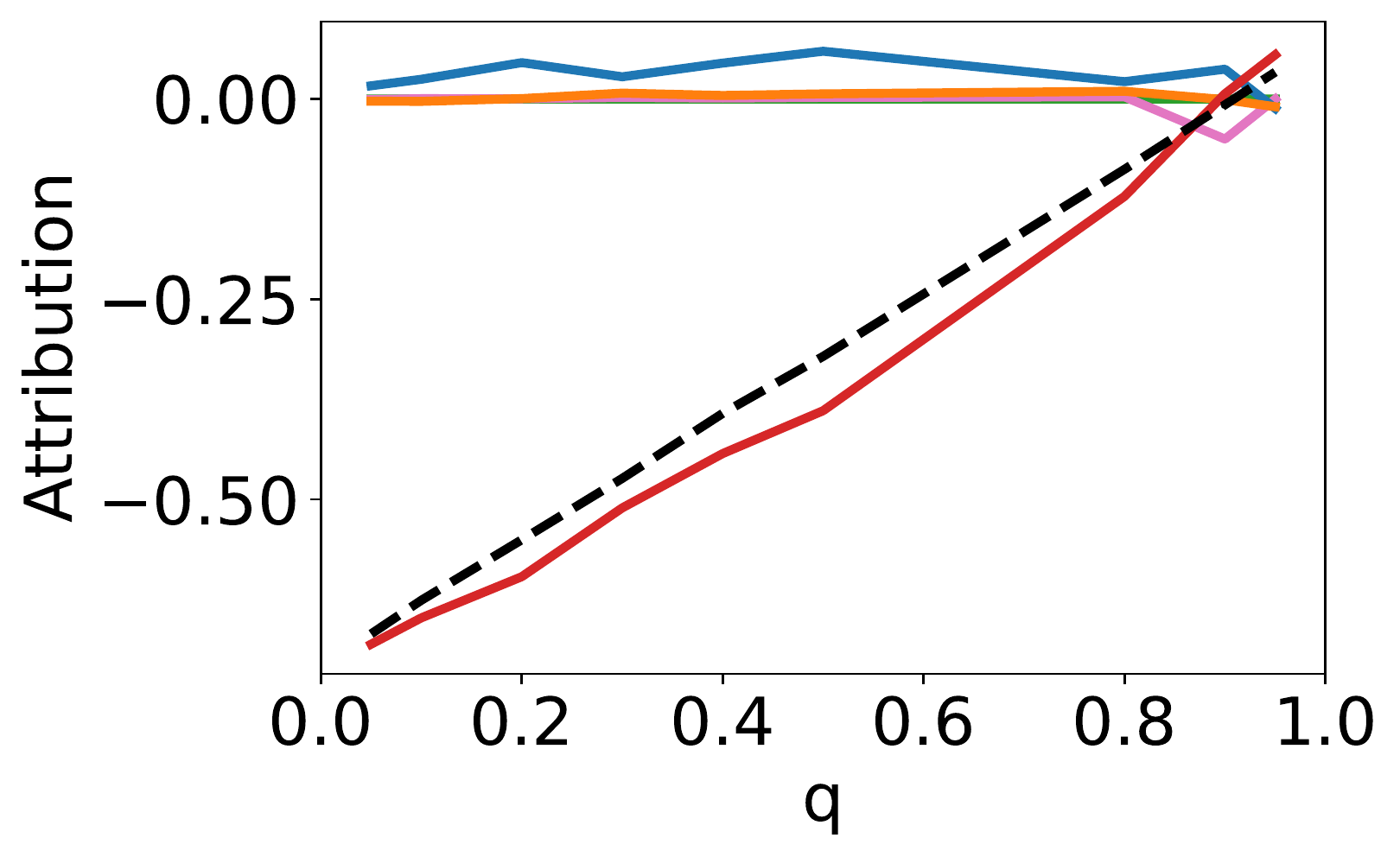}
  \caption{Label Shift}
\end{subfigure}%
\begin{subfigure}{.42\linewidth}
  \centering
  \includegraphics[width=.94\linewidth]{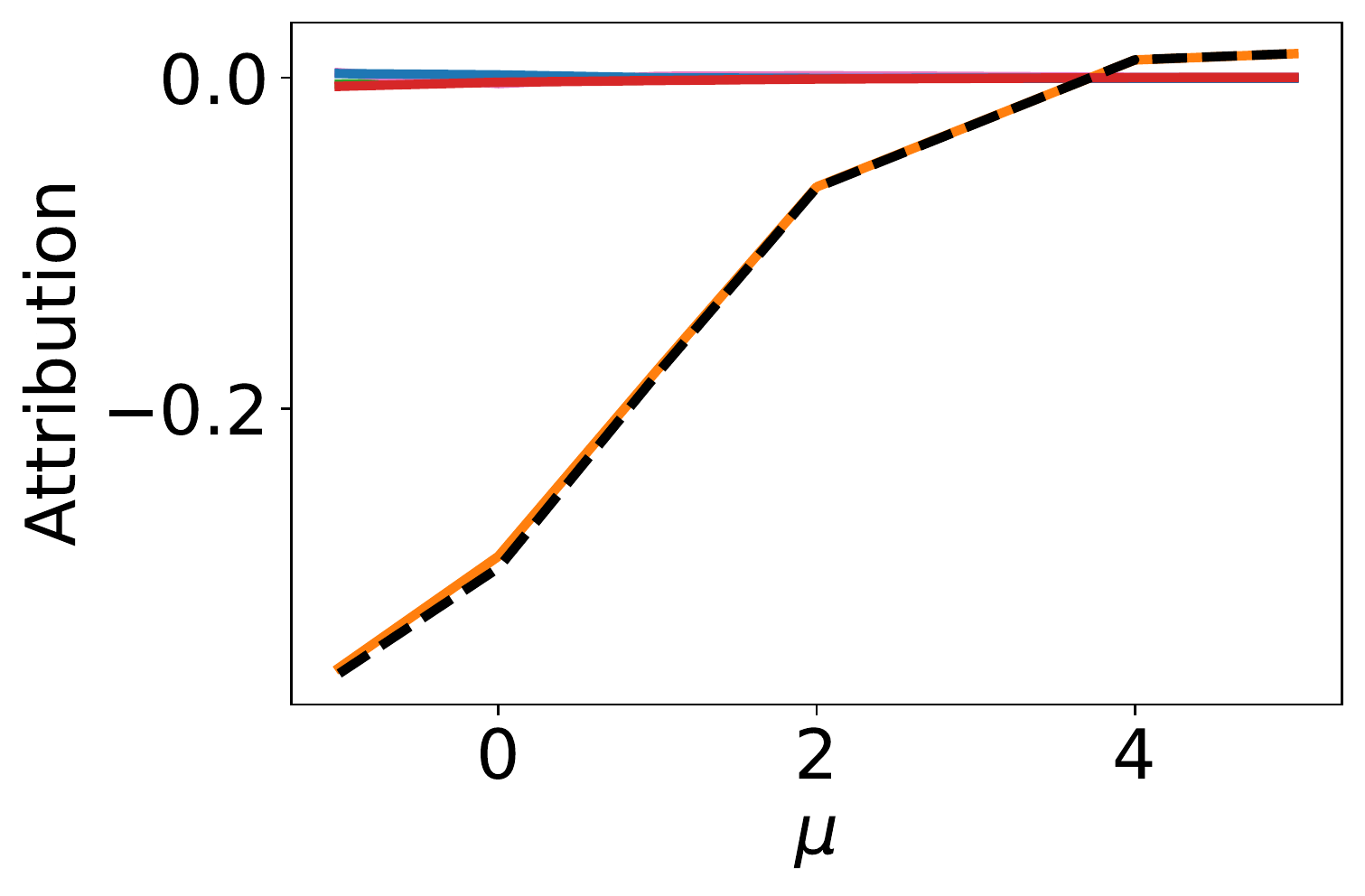}
  \caption{Conditional Covariate Shift}
\end{subfigure}
\begin{subfigure}{.42\linewidth}
  \centering
  \includegraphics[width=.94\linewidth]{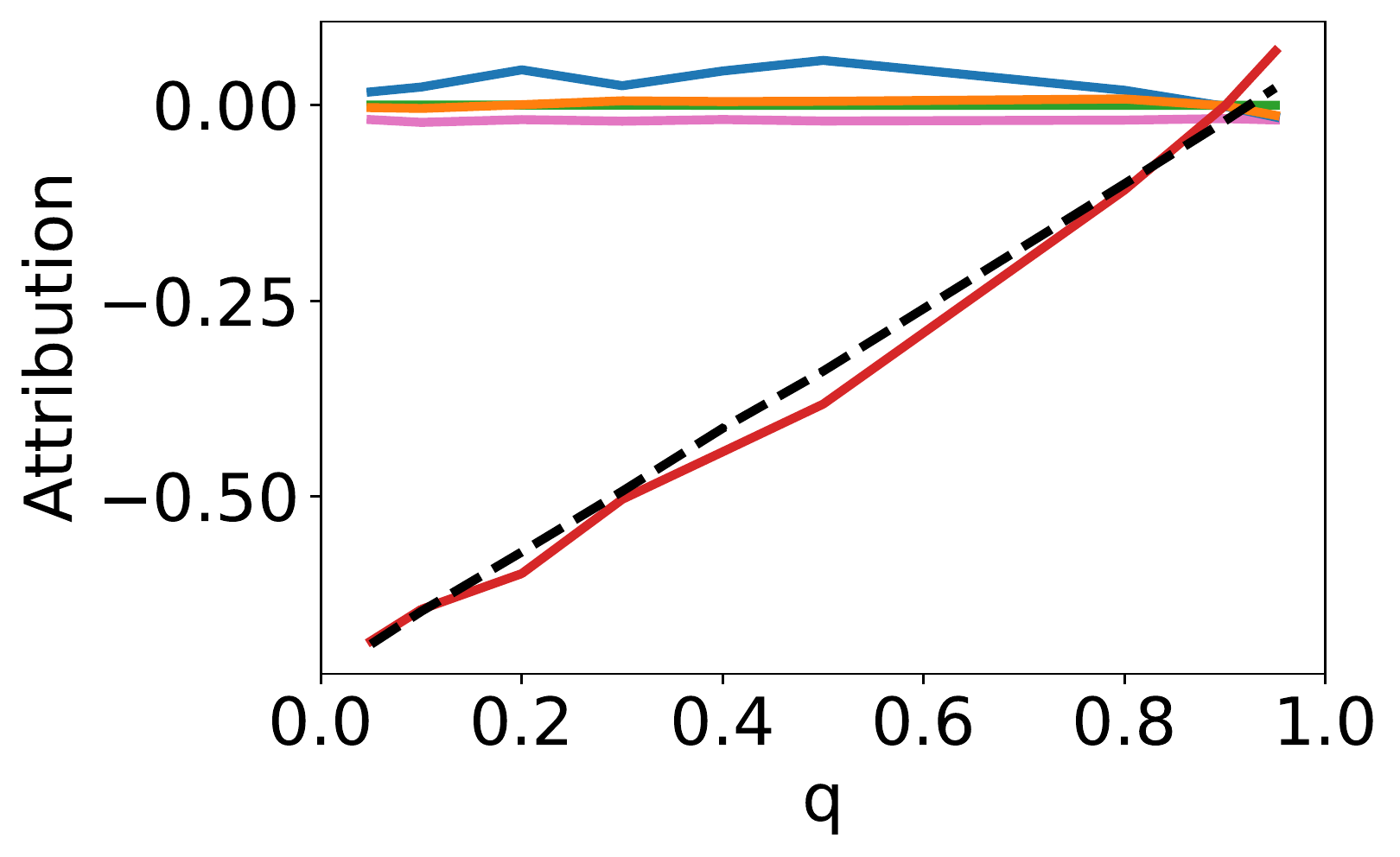}
  \caption{Combined Shift 1}
\end{subfigure}
\begin{subfigure}{.58\linewidth}
  \centering
  \includegraphics[width=.99\linewidth]{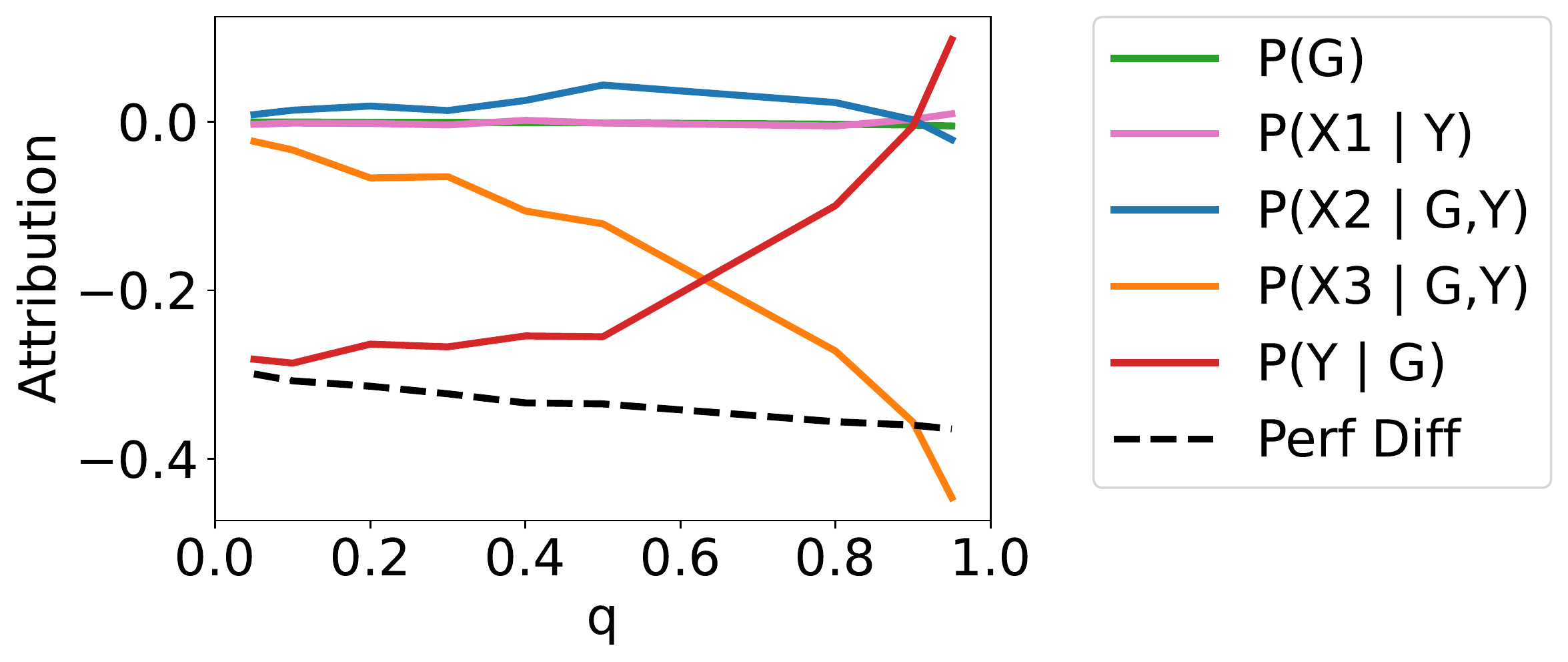}
  \caption{Combined Shift 2}
\end{subfigure}
\caption{Attributions by our method using the correct causal graph for the change in accuracy to five potential distributional shifts on the synthetic dataset for the \texttt{LR} model. Further from $0$ implies higher (signed) attribution. We observe that the overall change (Perf Diff) is attributed to the true shift(s) in all cases.}
\label{fig:lr_synthetic_acc}
\end{figure*}

\paragraph{Our method correctly identifies distribution shifts.}
First, we focus on the output of our method with \texttt{LR} as the model of interest and accuracy as the metric, shown in Figure \ref{fig:lr_synthetic_acc}. We find that our method attributes all of the performance changes to the correct ground truth shifts, both when there is a single shift (Settings (a) and (b)) and when there are multiple shifts (Settings (c) and (d)). In the case of Setting (c), we find that our method attributes all of the performance drop to a shift in $P(Y|G)$. This is because  the model relies largely on the spurious information ($G$ inferred from $X_2$ and $X_3$) in the source environment. We verify this by examining the overall feature importance for both models (see Table \ref{tab:app_synthetic_feat_imp}). Further, in the presence of multiple shifts which simultaneously impact model performance (Setting (d)), we find that our method is able to attribute a meaningful fraction of the performance shift to each distribution. 

\begin{figure*}[htbp!]
\begin{subfigure}{.42\linewidth}
  \centering
  \includegraphics[width=.94\linewidth]{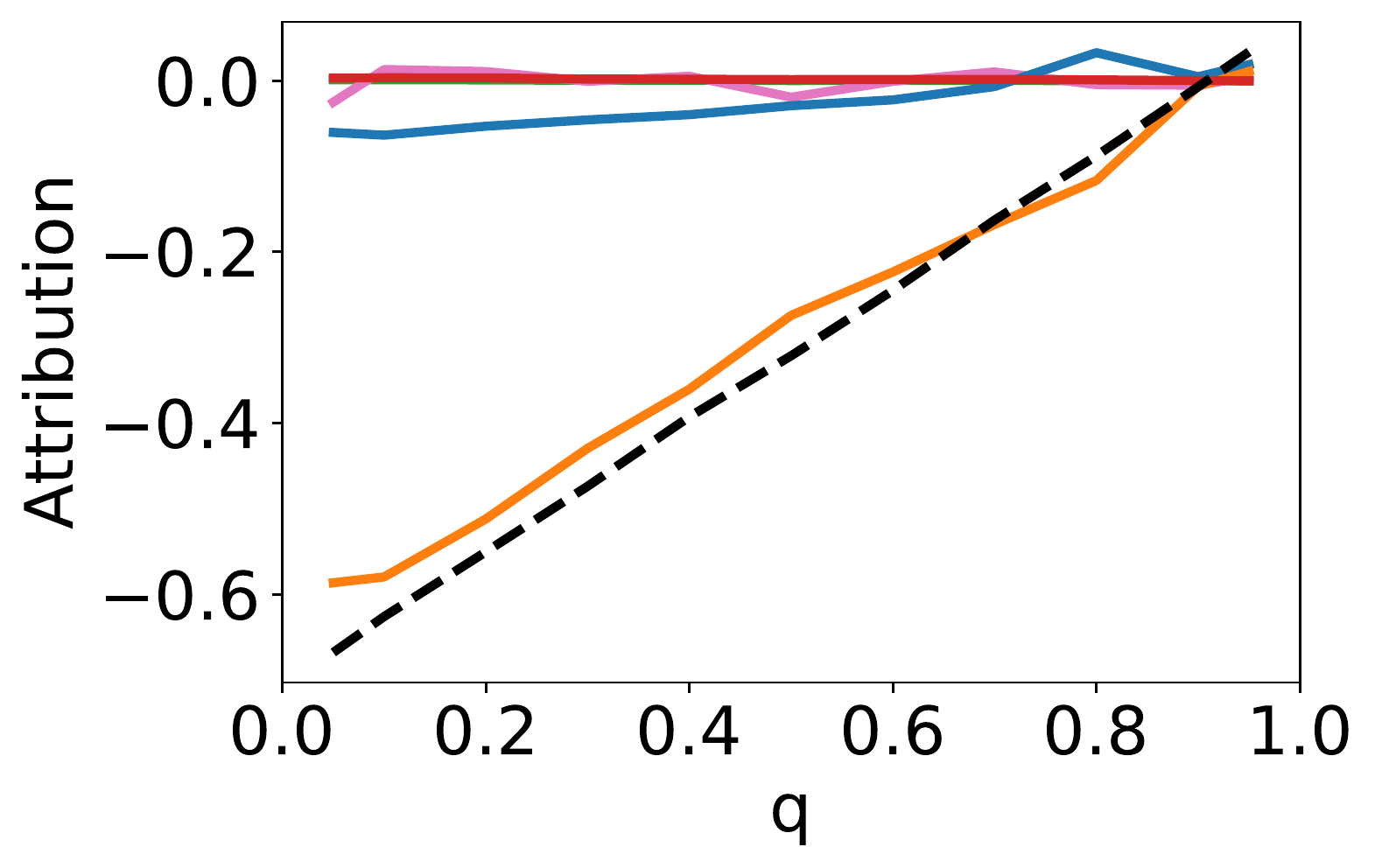}
  \caption{Label Shift}
\end{subfigure}%
\begin{subfigure}{.42\linewidth}
  \centering
  \includegraphics[width=.94\linewidth]{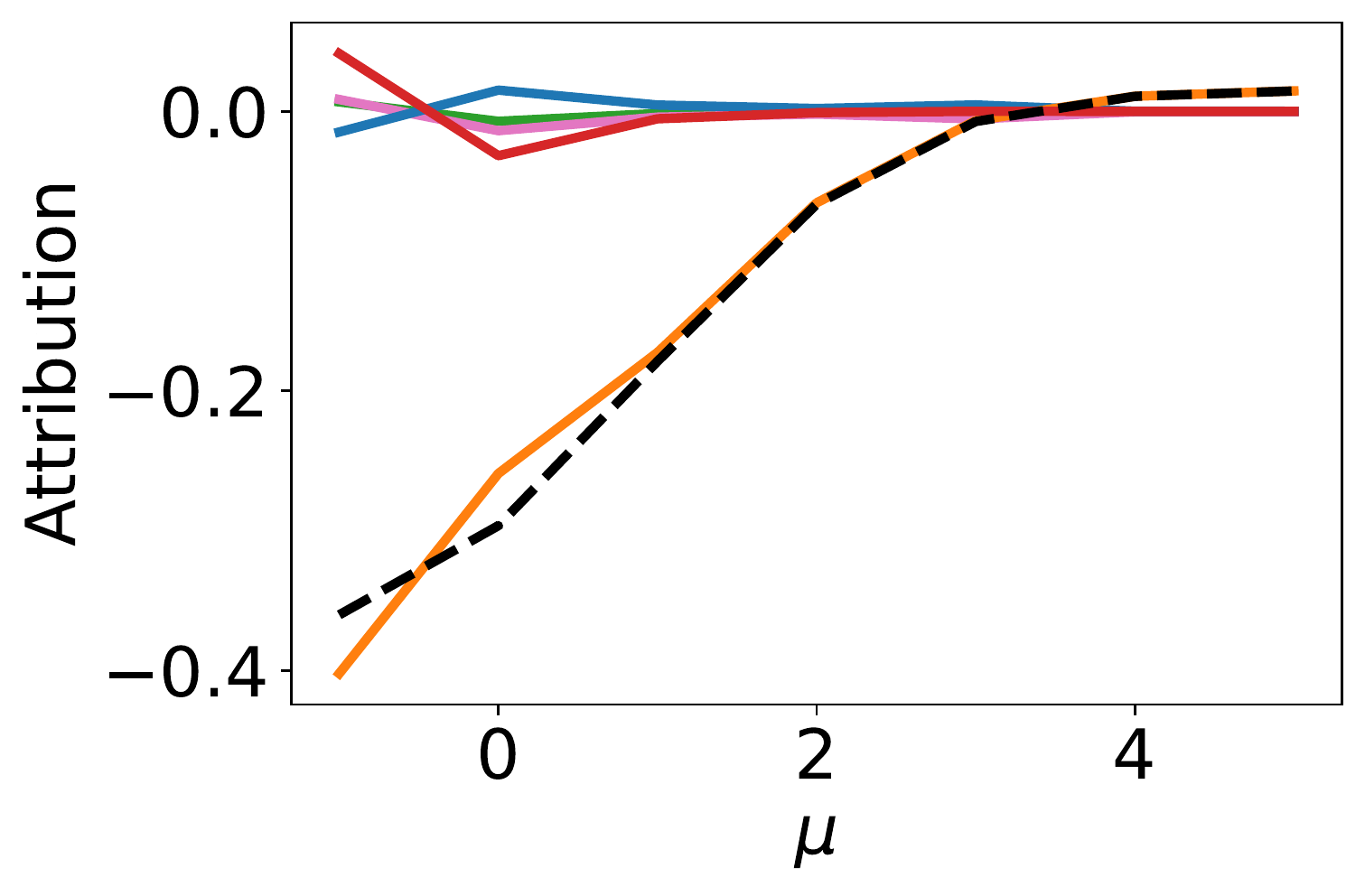}
  \caption{Conditional Covariate Shift}
\end{subfigure}
\begin{subfigure}{.42\linewidth}
  \centering
  \includegraphics[width=.94\linewidth]{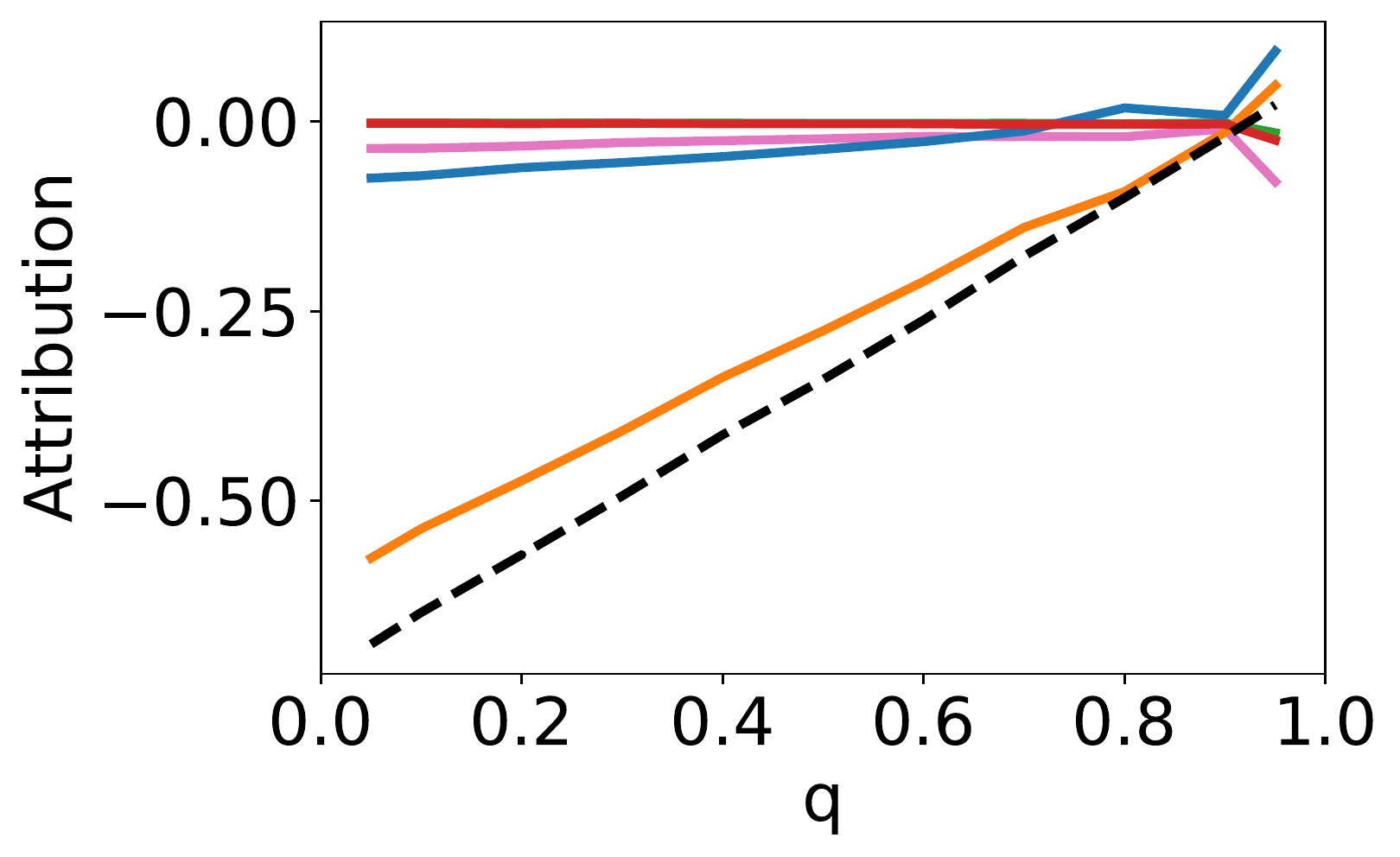}
  \caption{Combined Shift 1}
\end{subfigure}
\begin{subfigure}{.58\linewidth}
  \centering
  \includegraphics[width=.99\linewidth]{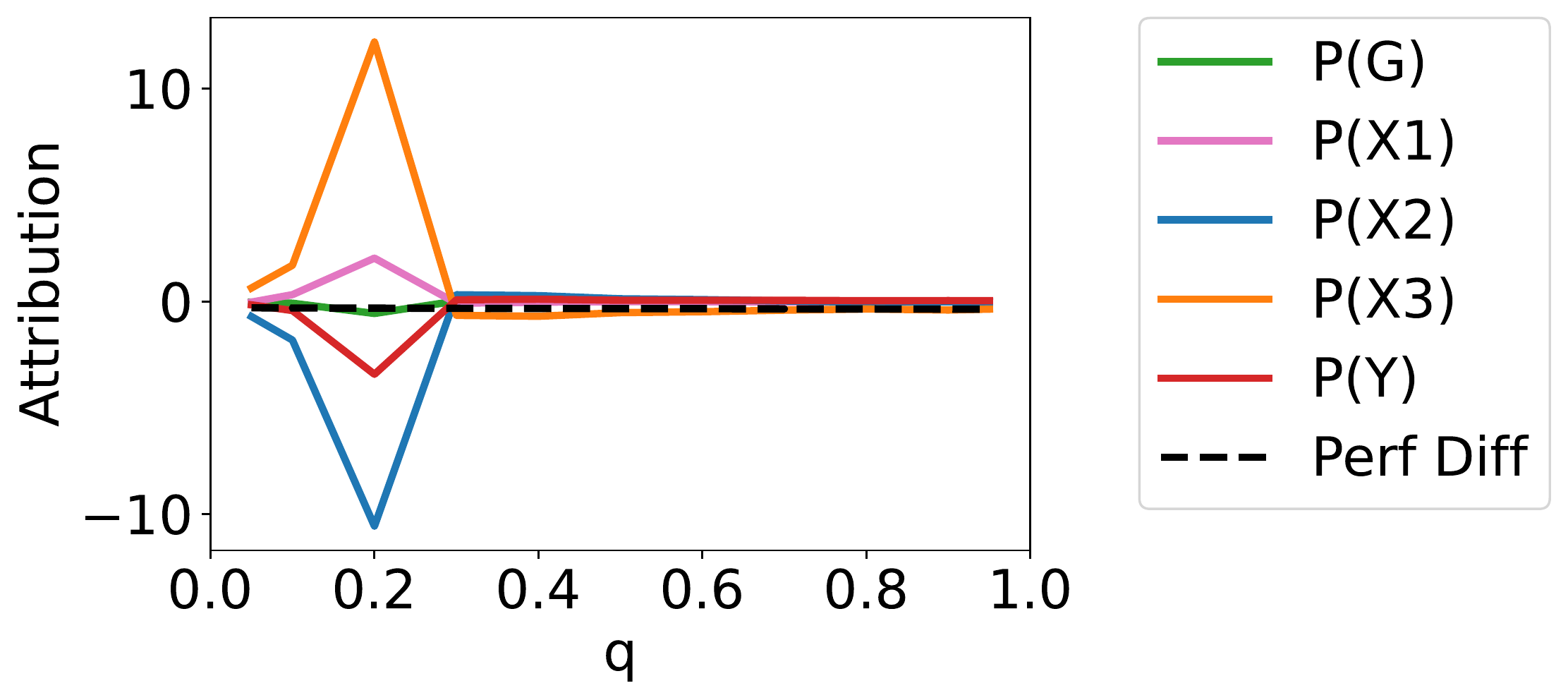}
  \caption{Combined Shift 2}
\end{subfigure}
\caption{Accuracy differences attributed by our method using the marginal graph to five potential distributional shifts on the synthetic dataset for the \texttt{LR} model. }
\label{fig:lr_synthetic_marginals}
\end{figure*}

\begin{figure*}[htbp!]
\begin{subfigure}{.42\linewidth}
  \centering
  \includegraphics[width=.94\linewidth]{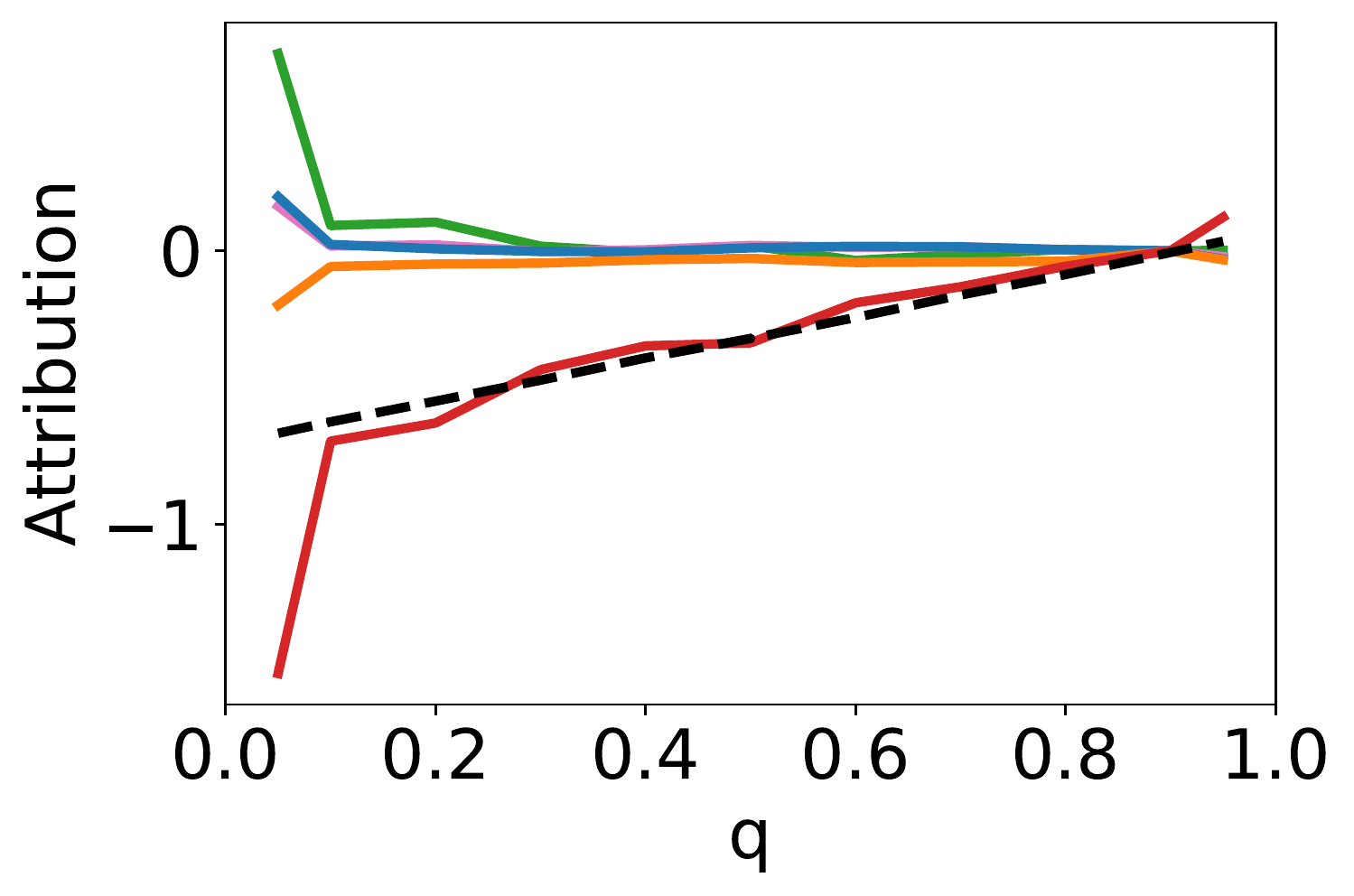}
  \caption{Label Shift}
\end{subfigure}%
\begin{subfigure}{.42\linewidth}
  \centering
  \includegraphics[width=.94\linewidth]{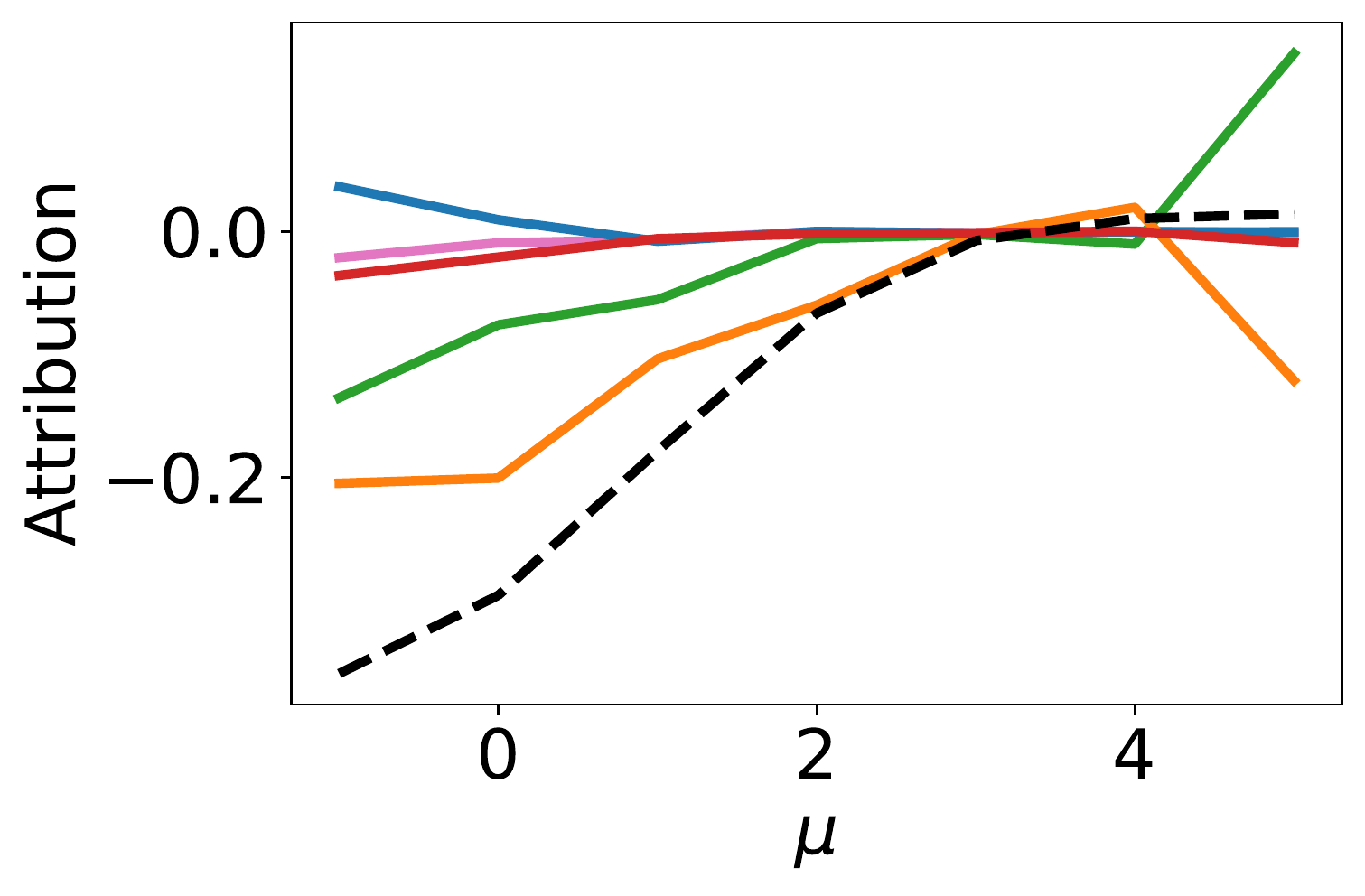}
  \caption{Conditional Covariate Shift}
\end{subfigure}
\begin{subfigure}{.42\linewidth}
  \centering
  \includegraphics[width=.94\linewidth]{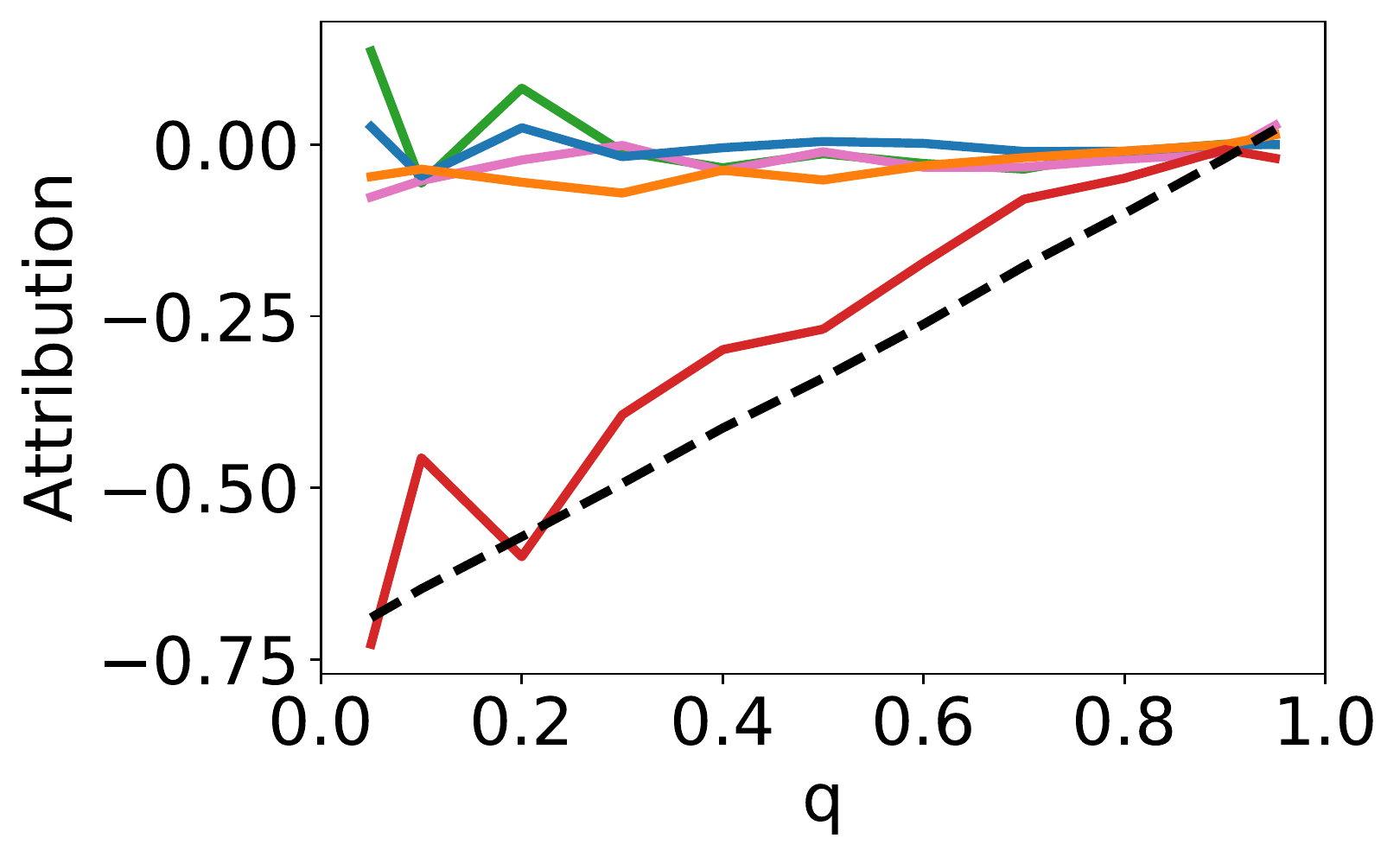}
  \caption{Combined Shift 1}
\end{subfigure}
\begin{subfigure}{.58\linewidth}
  \centering
  \includegraphics[width=.99\linewidth]{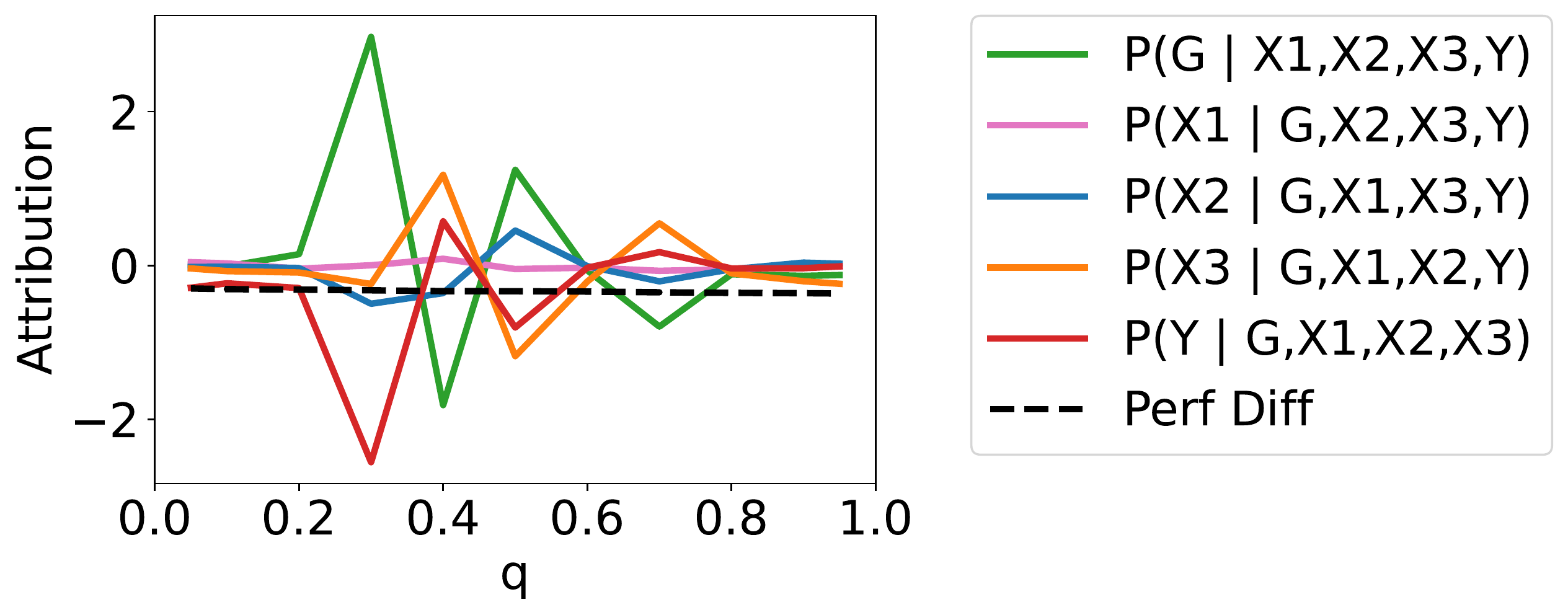}
  \caption{Combined Shift 2}
\end{subfigure}
\caption{Accuracy differences attributed by our method using the fully connected graph to five potential distributional shifts on the synthetic dataset for the \texttt{LR} model. }
\label{fig:lr_synthetic_conds}
\end{figure*}

\begin{figure*}[htbp!]
\begin{subfigure}{.42\linewidth}
  \centering
  \includegraphics[width=.94\linewidth]{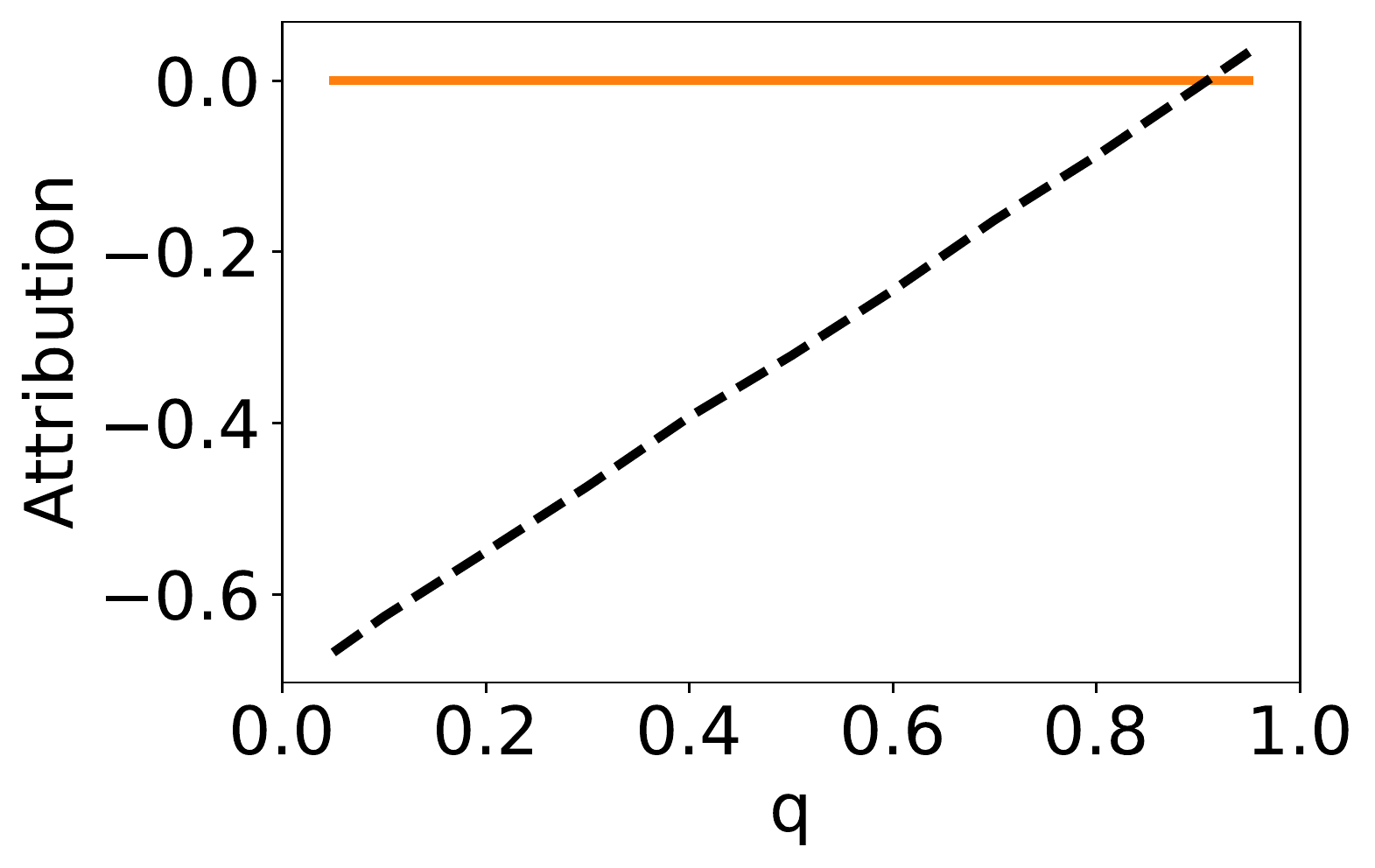}
  \caption{Label Shift}
\end{subfigure}%
\begin{subfigure}{.42\linewidth}
  \centering
  \includegraphics[width=.94\linewidth]{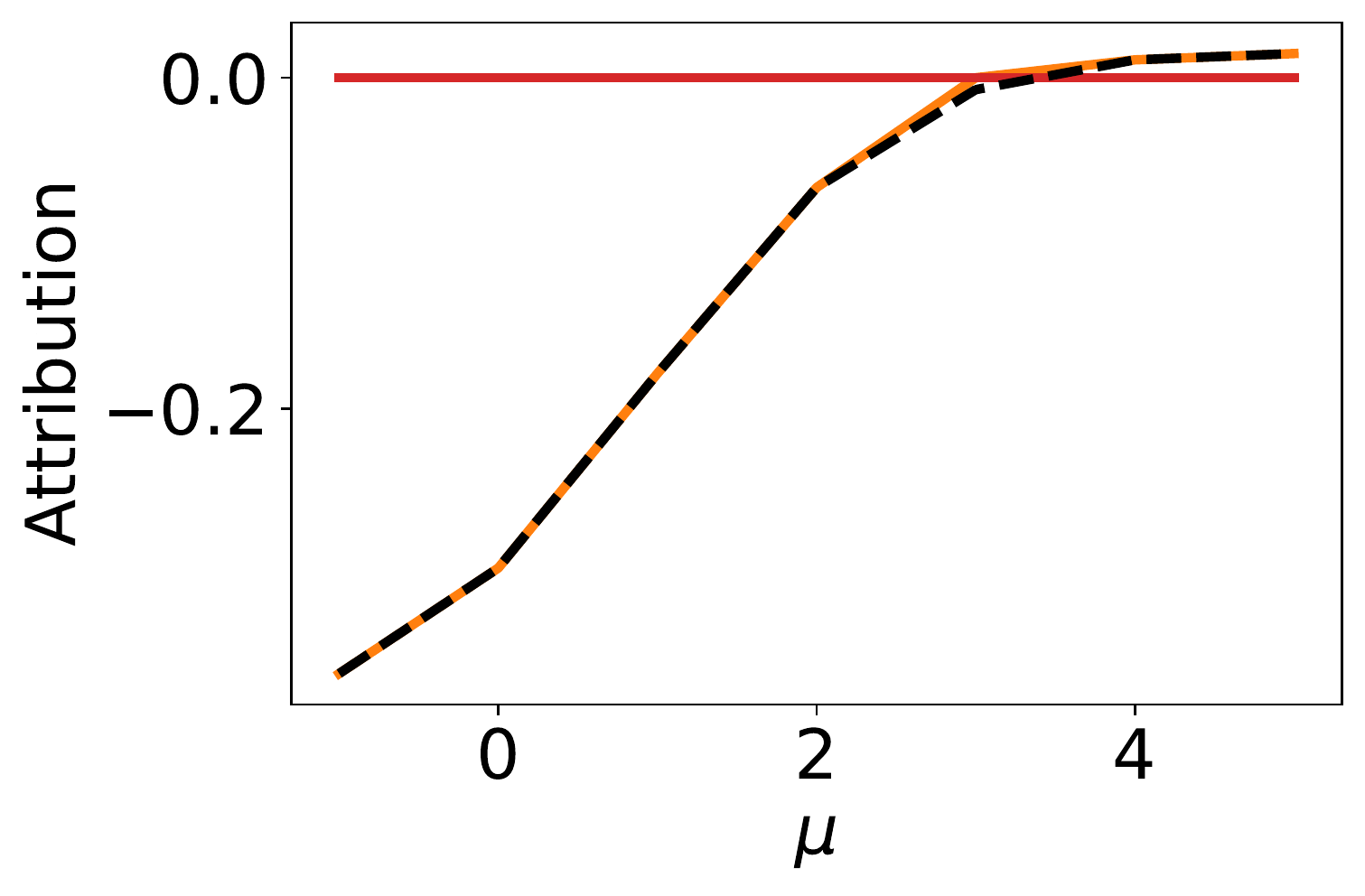}
  \caption{Conditional Covariate Shift}
\end{subfigure}
\begin{subfigure}{.42\linewidth}
  \centering
  \includegraphics[width=.94\linewidth]{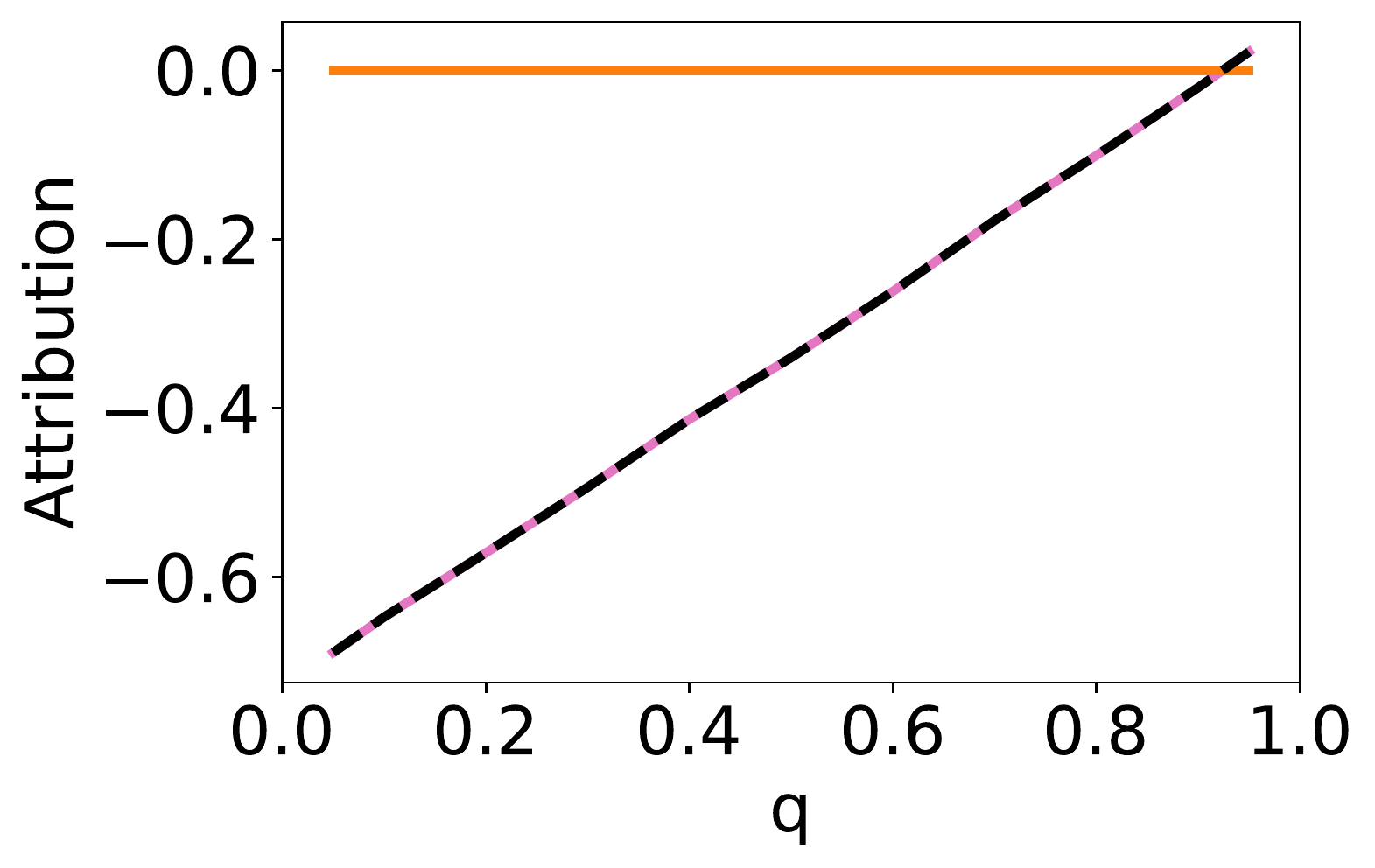}
  \caption{Combined Shift 1}
\end{subfigure}
\begin{subfigure}{.58\linewidth}
  \centering
  \includegraphics[width=.99\linewidth]{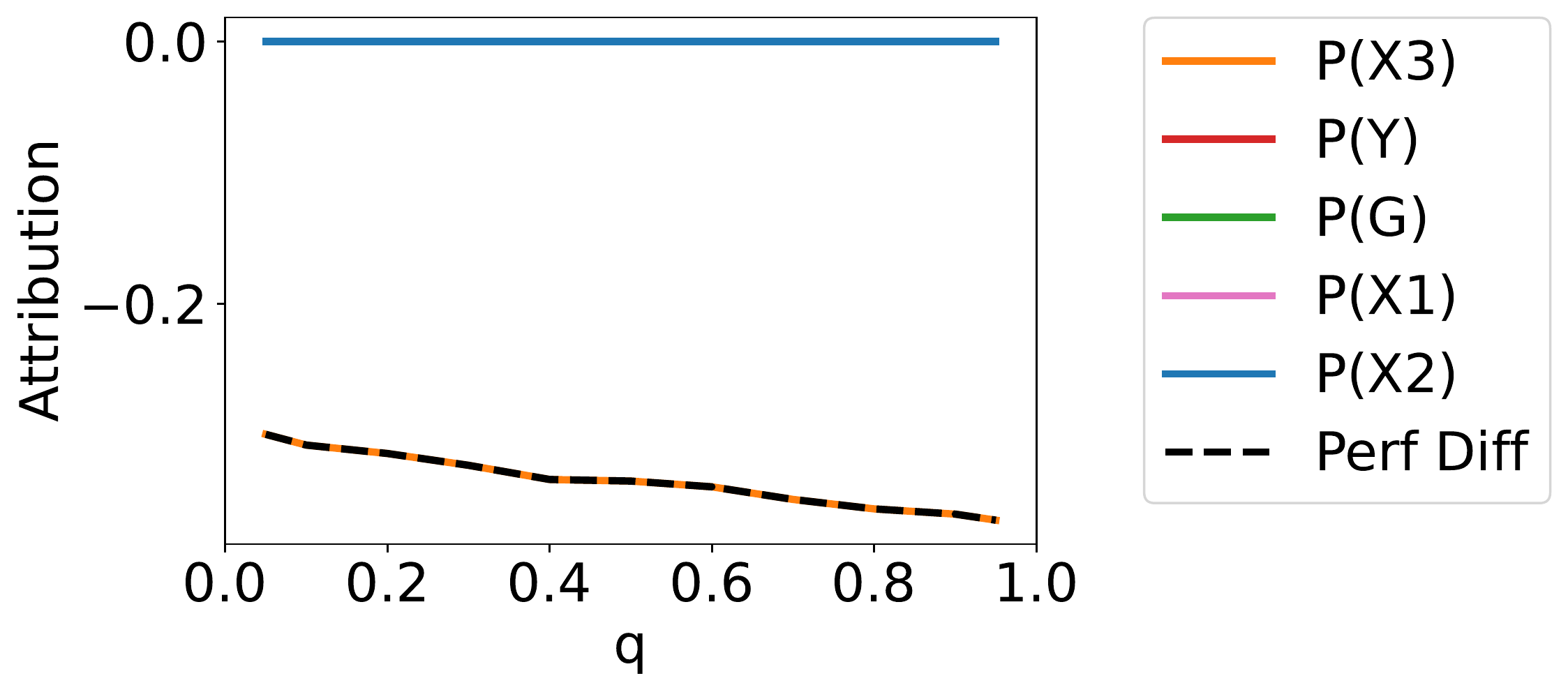}
  \caption{Combined Shift 2}
\end{subfigure}
\caption{Accuracy differences attributed by the SHAP baseline to five potential distributional shifts on the synthetic dataset for the \texttt{LR} model. }
\label{fig:lr_synthetic_shap}
\end{figure*}

\begin{figure*}[htbp!]
\begin{subfigure}{.43\linewidth}
  \centering
  \includegraphics[width=.94\linewidth]{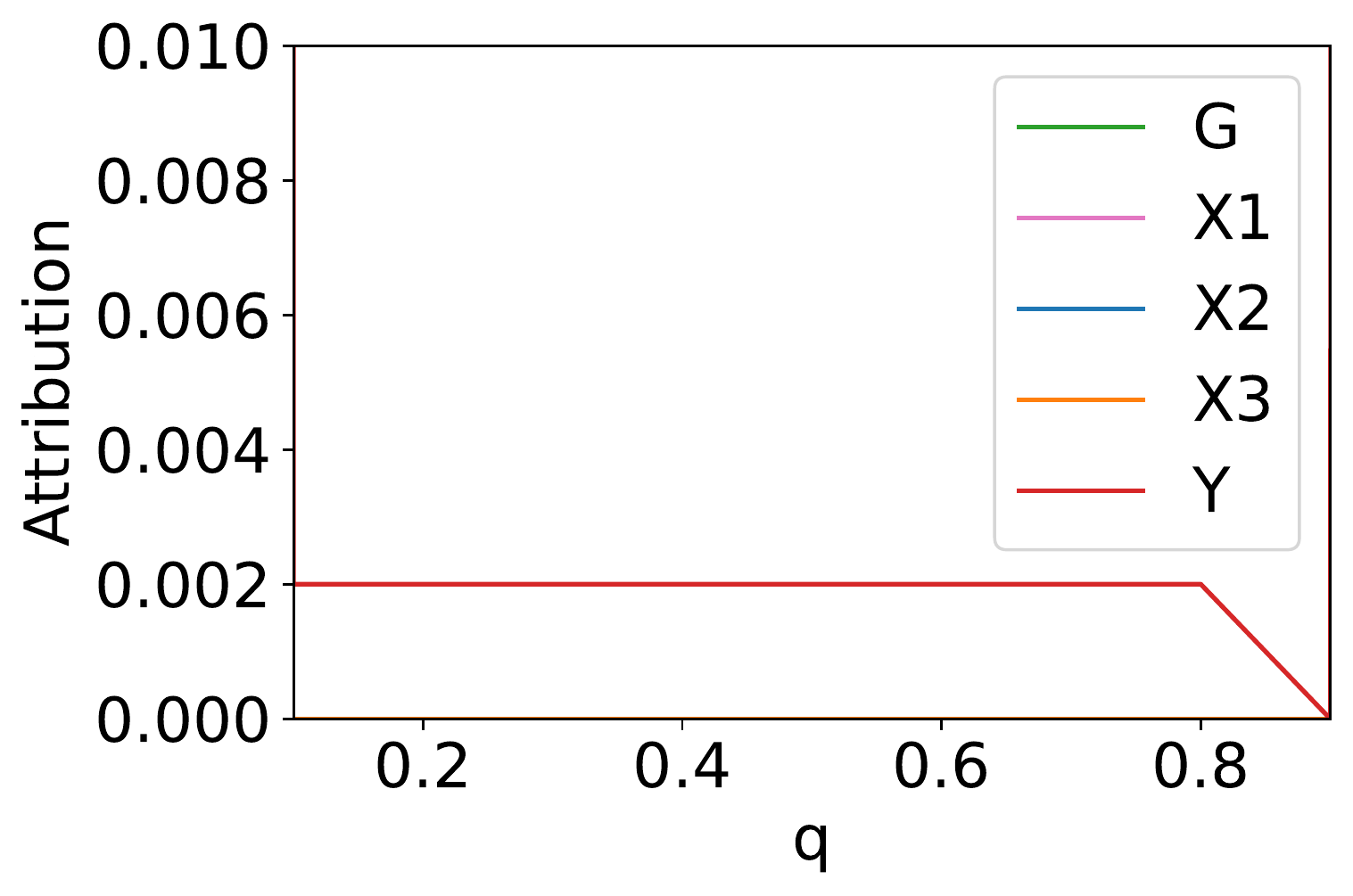}
  \caption{Label Shift}
\end{subfigure}%
\begin{subfigure}{.43\linewidth}
  \centering
  \includegraphics[width=.94\linewidth]{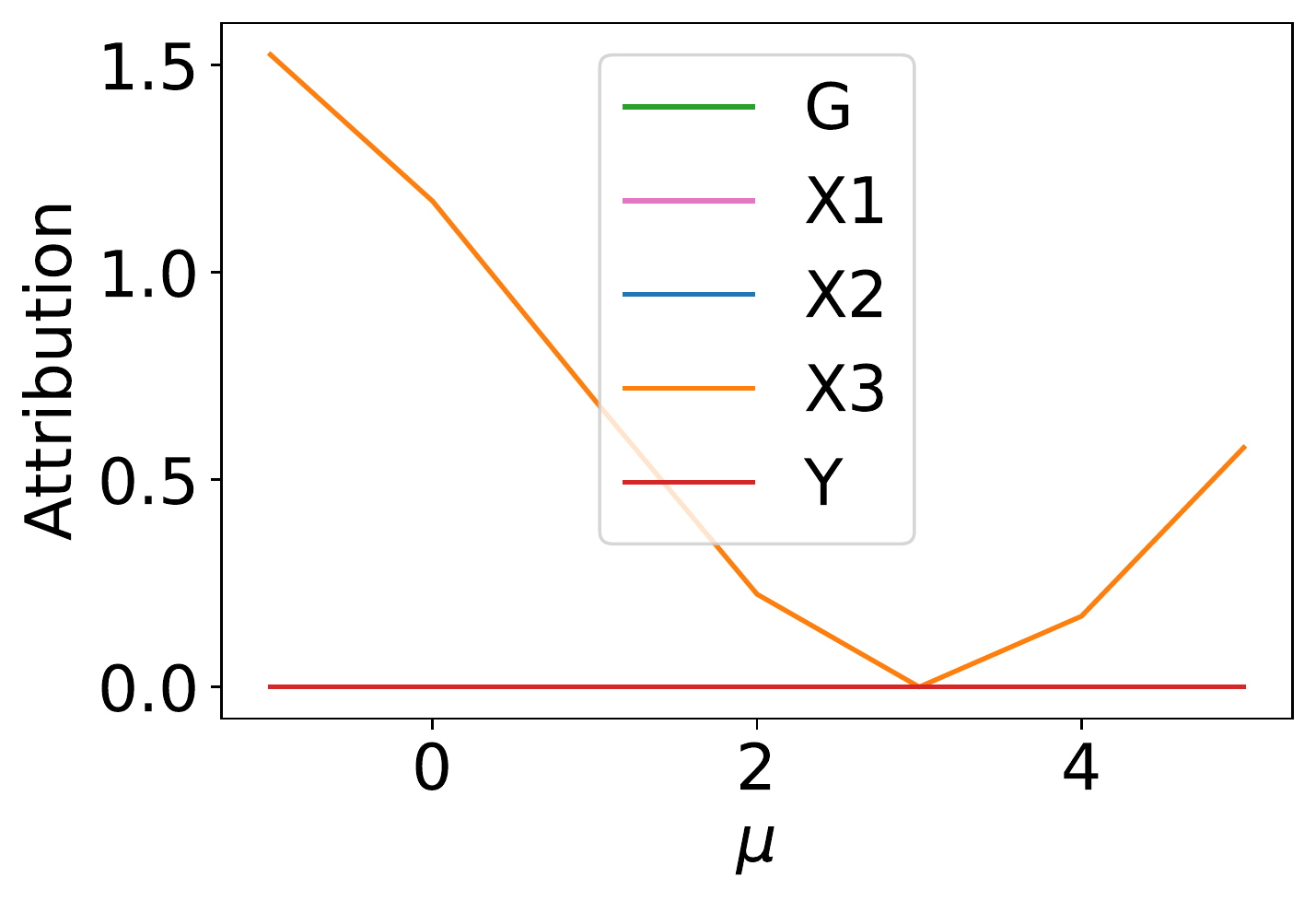}
  \caption{Conditional Covariate Shift}
\end{subfigure}
\begin{subfigure}{.48\linewidth}
  \centering
  \includegraphics[width=.99\linewidth]{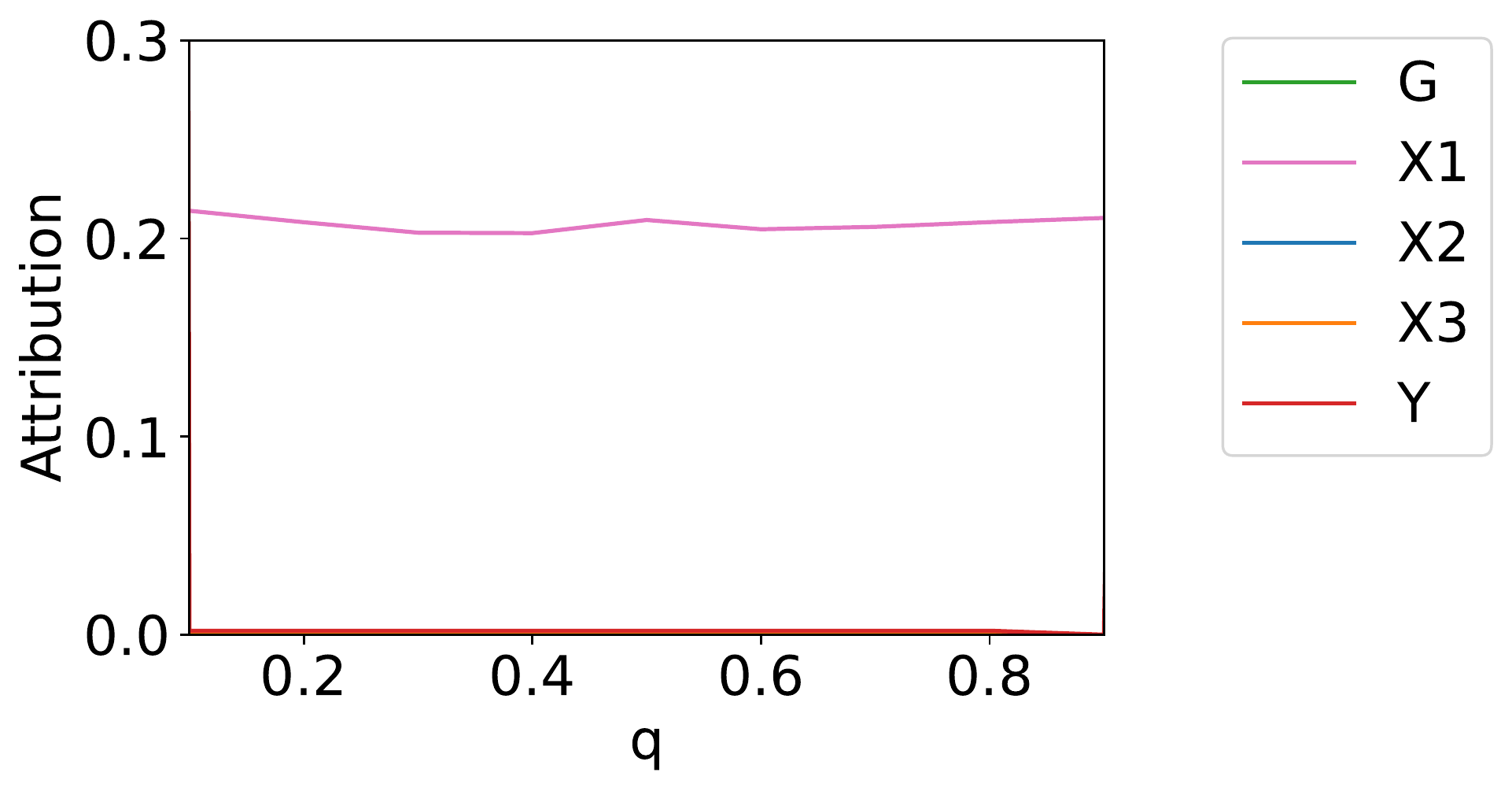}
  \caption{Combined Shift 1}
\end{subfigure}
\begin{subfigure}{.48\linewidth}
  \centering
  \includegraphics[width=.99\linewidth]{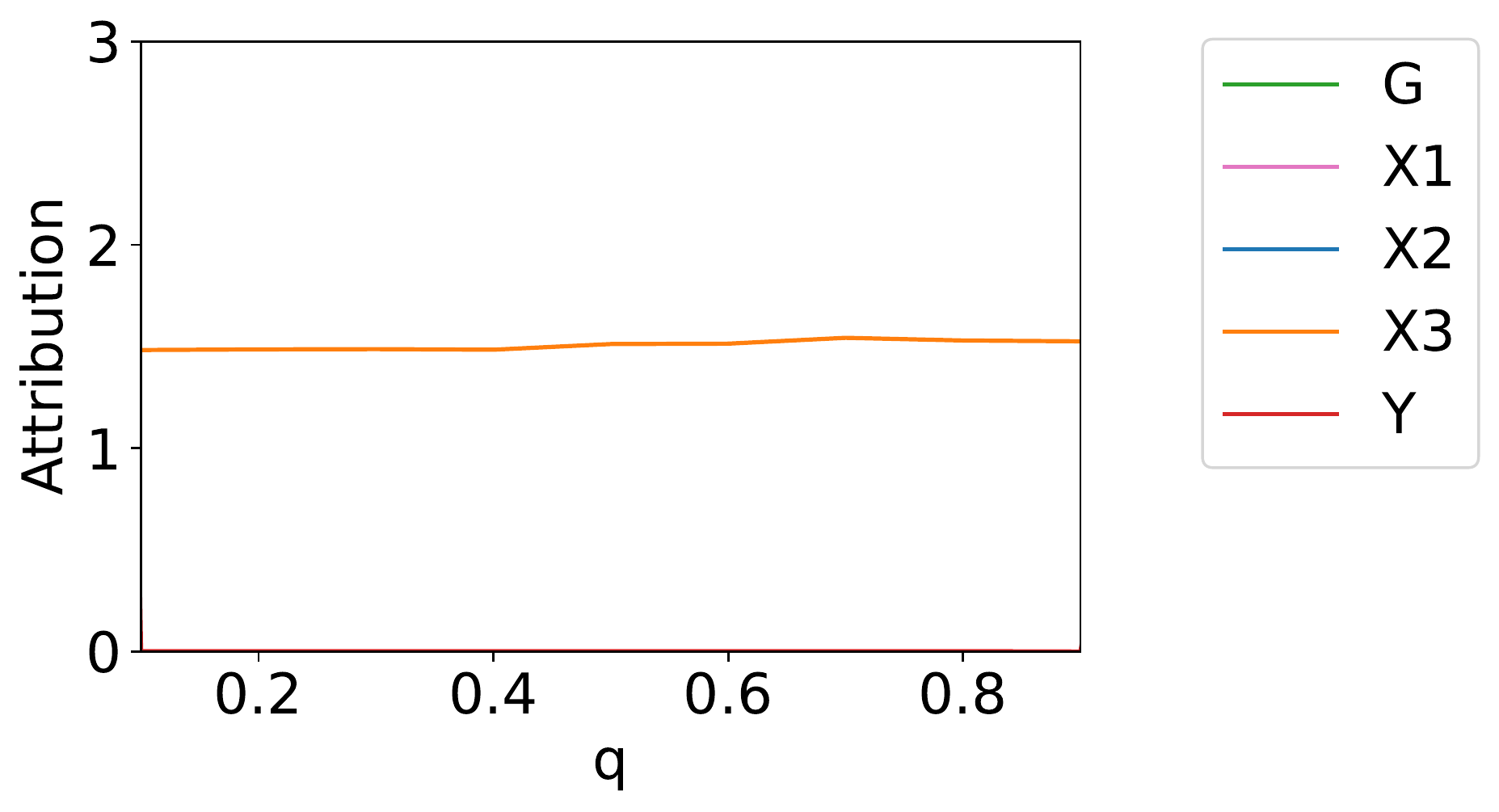}
  \caption{Combined Shift 2}
\end{subfigure}
\caption{Attributions by the joint method in \citet{budhathoki2021distribution} to five potential distributional shifts on the synthetic dataset. We note that the magnitude of the attribution is not informative in interpreting model performance changes, particularly when multiple shifts are present.}
\label{fig:synthetic_janzing}
\end{figure*}

\paragraph{Baseline methods have multiple flaws.} We find that the baselines methods all have several flaws which result in inadequate attributions in this setting. For example, the marginal candidate set (Figure \ref{fig:lr_synthetic_marginals}) does not provide meaningful attributions as it does not examine conditional relationships, especially as it attributes large shifts to $P(X_3)$. Similarly, the fully connected graph (Figure \ref{fig:lr_synthetic_conds}) demonstrates a large degree of noise, particularly in the combined shift, though the dominant distribution appears to largely be correct. Next, the SHAP baseline (Figure \ref{fig:lr_synthetic_shap}) completely fails in this setting, as it is not able to attribute any shift to the mechanism for $Y$. Finally, we find that the attributions provided by the joint method in \cite{budhathoki2021distribution} (Figure \ref{fig:synthetic_janzing}) are not meaningful, as the magnitude of the KL divergence varies wildly between distributions when multiple shifts are present.

\begin{table}[htbp!]
\caption{Performance of each model on the source environment for the synthetic dataset.}
\centering
\begin{tabular}{@{}lll@{}}
\toprule
    & Accuracy & Brier Score \\ \midrule
\texttt{LR}  & 0.871    & 0.102       \\
\texttt{XGB} & 0.870    & 0.099       \\ \bottomrule
\end{tabular}
\end{table}

\begin{table}[htbp!]
\caption{Feature importances of each model on the synthetic dataset. For \texttt{LR}, the model coefficient is shown, and for \texttt{XGB}, the total information gain from each feature.}
\centering
\begin{tabular}{@{}lll@{}}
\toprule
   & \texttt{LR} (Coefficient) & \texttt{XGB} (Gain) \\ \midrule
$X_1$ & 0.400            & 31.1       \\
$X_2$ & 0.381            & 29.2       \\
$X_3$ & 1.994            & 358.2      \\ \bottomrule
\end{tabular}
\label{tab:app_synthetic_feat_imp}
\end{table}

\FloatBarrier

\subsection{ColoredMNIST}
\label{sec:cmnist}

\paragraph {Setup.} We evaluate our method on the popular ColoredMNIST dataset \cite{arjovsky2019invariant}. We generalize the data generating process for this dataset to include several tunable dataset parameters, using the following procedure:
\begin{enumerate}[leftmargin=*,noitemsep,topsep=0pt]
    \item Generate a binary label $y_{obs}$ from the MNIST label $y_{num}$ by assigning $y_{obs} = 0$ if $y_{num} \in \{0, 1, 2, 3, 4\}$, and  $y_{obs} = 1$ otherwise.
    \item Flip $y_{obs}$ with probability $\eta$ to obtain $y$.
    \item Generate the color $a$ by flipping $y$ with probability $\rho$.
    \item Construct $X$ as $[X_{fig} \cdot (1 - a), X_{fig} \cdot a]$.
    \item Subsample the majority class so that the fraction of samples with $y = 1$ in the dataset is equal to $\beta$.
\end{enumerate}

\begin{figure}[htbp!]
    \centering
  \includegraphics[width=.2\linewidth]{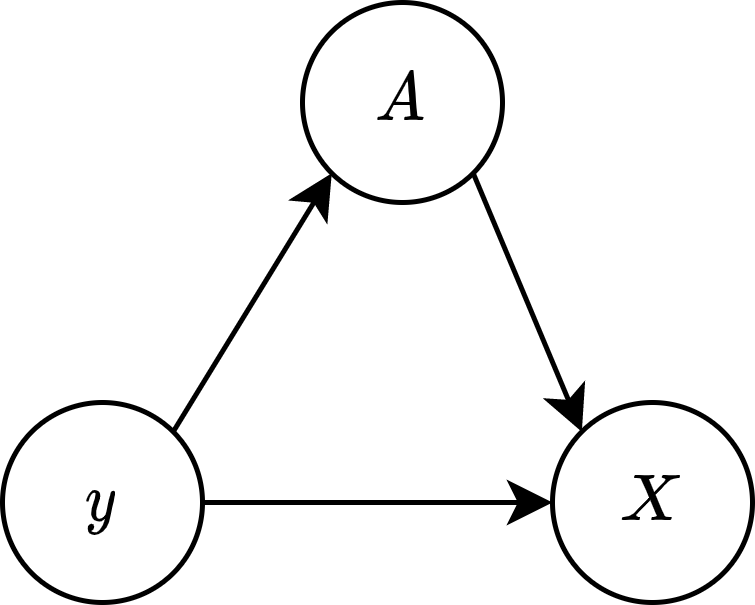}
  \caption{Causal graph for the ColoredMNIST dataset.}
  \label{fig:cmnist_graph}
\end{figure}

This corresponds to the causal graph in Figure \ref{fig:cmnist_graph}. Note that $X$ is a vector-valued node representing the image. We split the MNIST data equally into source and target environments. On the source environment, we set $\beta$ = 0.5, $\eta = 0.25$, and $\rho = 0.15$. We vary these two parameters independently on the target environment, keeping the other fixed at their source value. Note that shifting $\beta$, $\eta$, and $\rho$ correspond to shifting $P(Y)$, $P(X | A, Y)$ and $P(A | Y)$ respectively.

Following \cite{arjovsky2019invariant}, we use 3-layer MLPs to predict $y$ from $X$ on the source dataset. We train this MLP with standard ERM, as well as with GroupDRO \cite{sagawa2019distributionally}. The network trained with ERM should rely on the spurious correlation (i.e. the color), while the GroupDRO network should be invariant to the color. We experiment with using both the correct causal graph, and the all marginal causal graph.

\begin{figure*}[htbp!]
\begin{subfigure}{.28\linewidth}
  \centering
  \includegraphics[width=.94\linewidth]{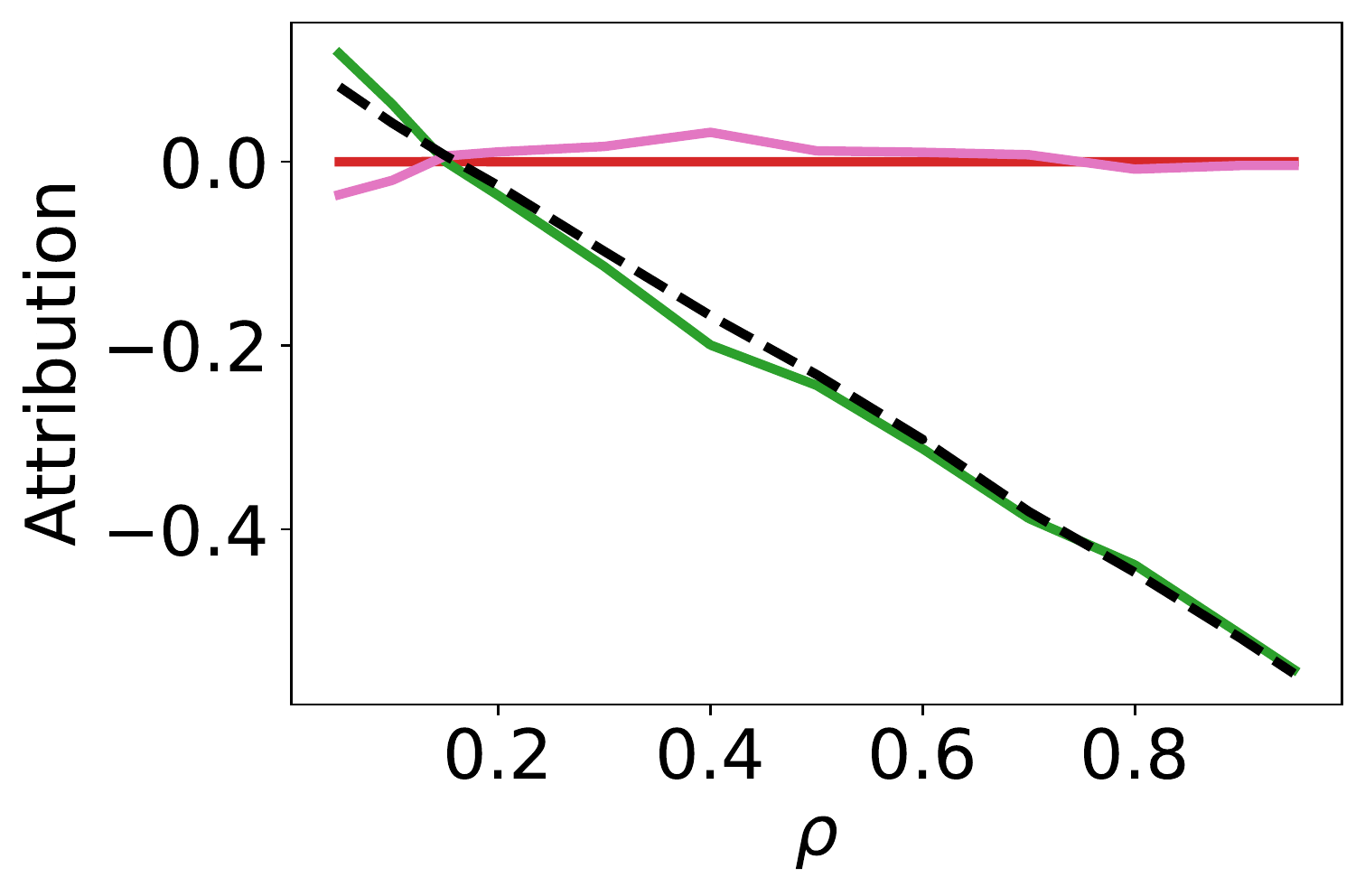}
  \caption{ERM}
\end{subfigure}%
\begin{subfigure}{.28\linewidth}
  \centering
  \includegraphics[width=.94\linewidth]{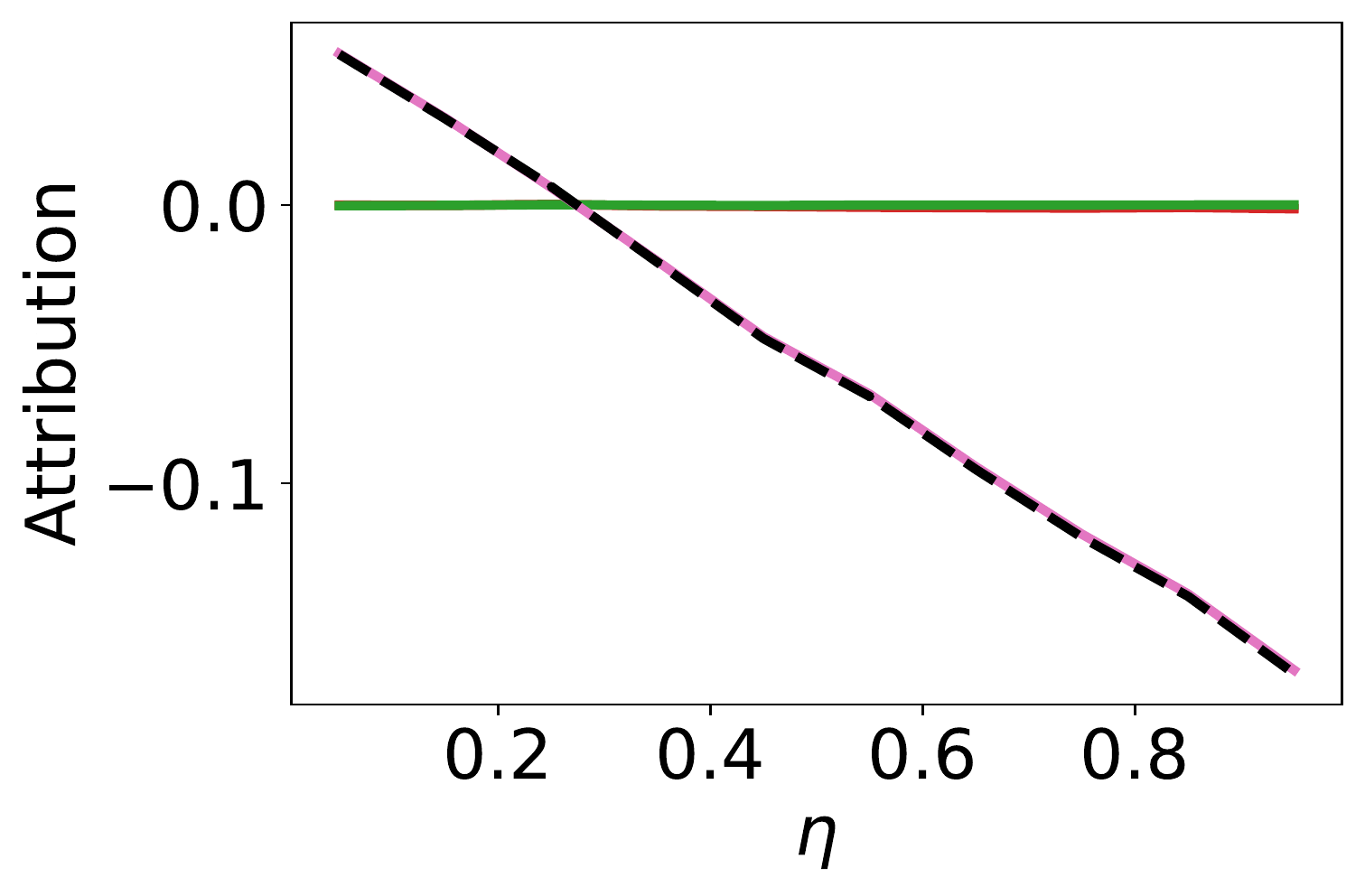}
  \caption{ERM}
\end{subfigure}
\begin{subfigure}{.39\linewidth}
  \centering
  \includegraphics[width=.99\linewidth]{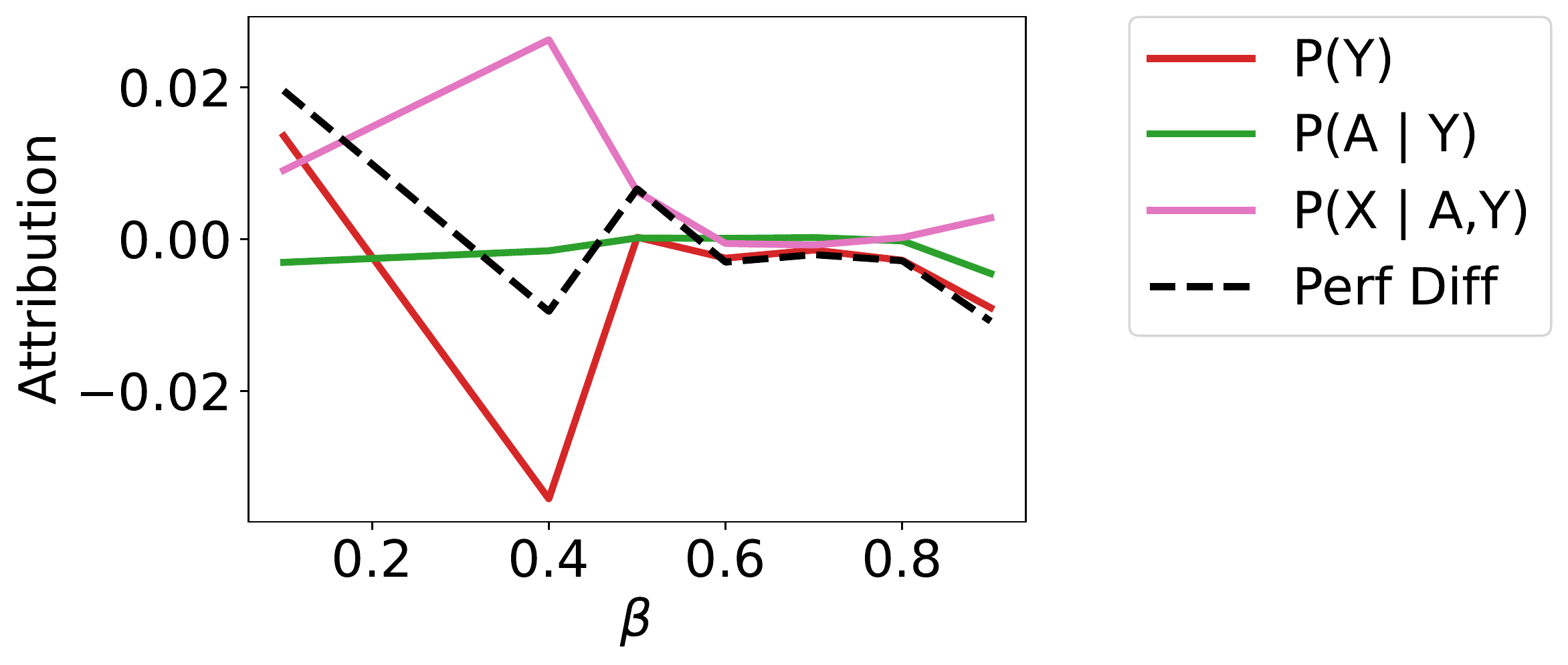}
  \caption{ERM}
\end{subfigure}%

\begin{subfigure}{.28\linewidth}
  \centering
  \includegraphics[width=.94\linewidth]{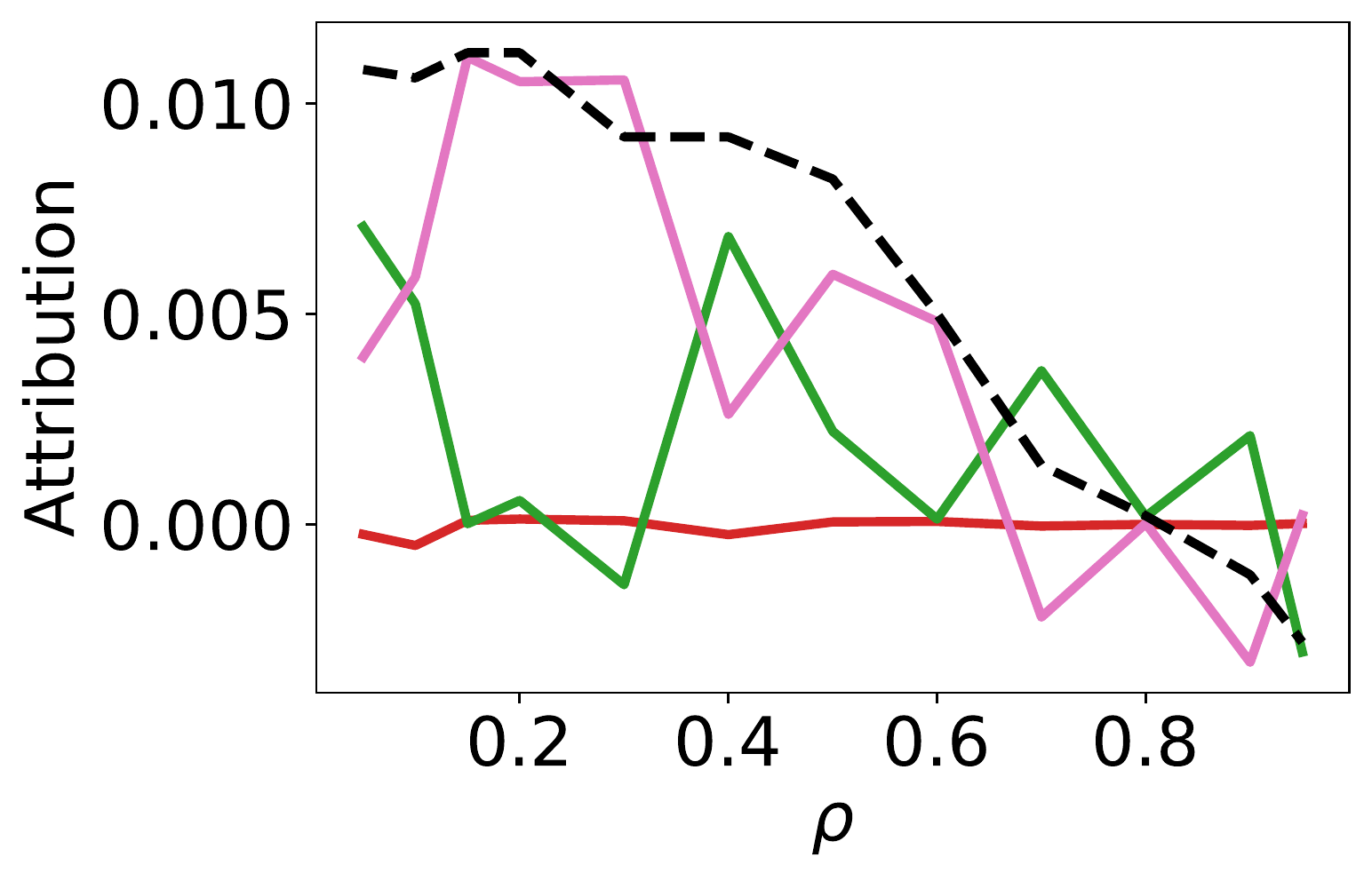}
  \caption{GroupDRO}
\end{subfigure}
\begin{subfigure}{.28\linewidth}
  \centering
  \includegraphics[width=.94\linewidth]{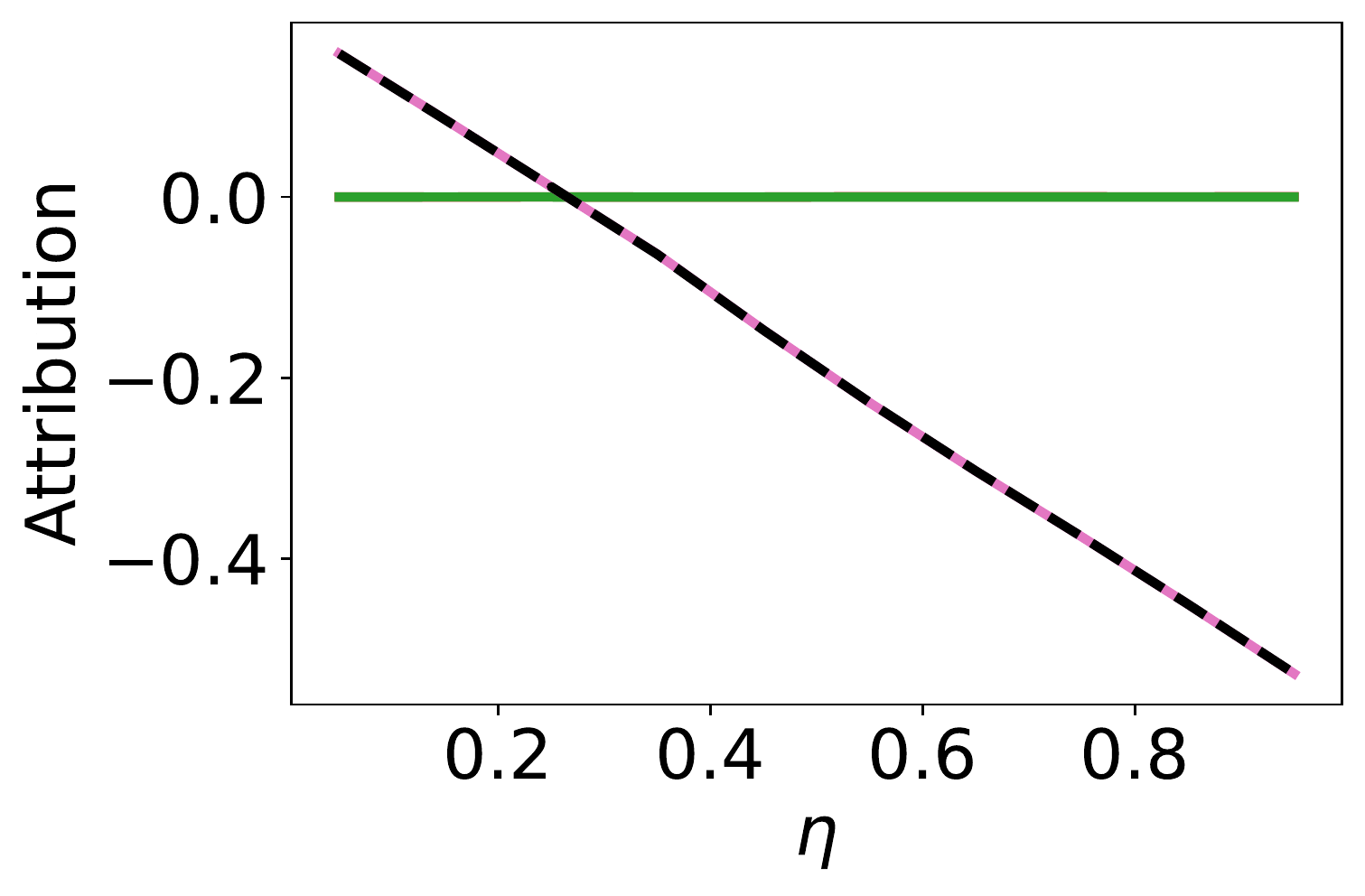}
  \caption{GroupDRO}
\end{subfigure}
\begin{subfigure}{.39\linewidth}
  \centering
  \includegraphics[width=.99\linewidth]{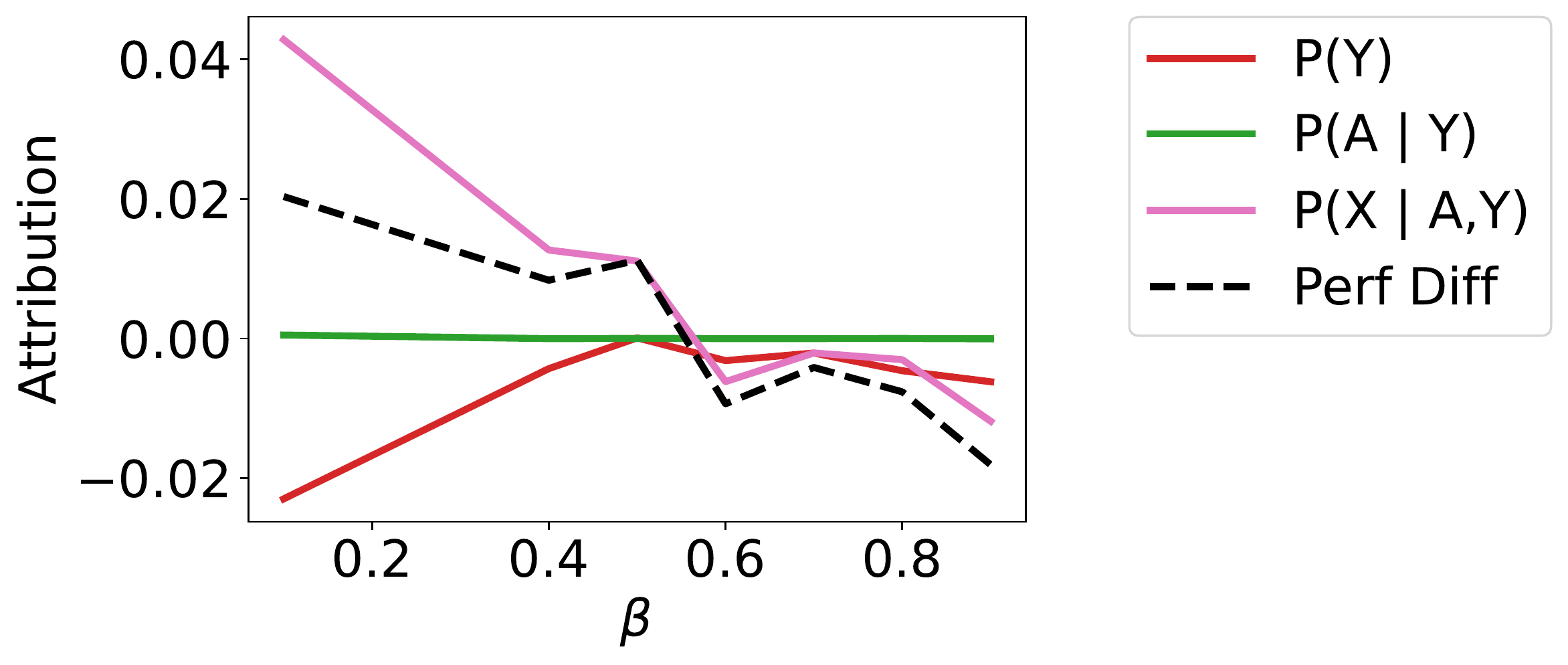}
  \caption{GroupDRO}
\end{subfigure}
\caption{Accuracy differences attributed by our model with the correct causal graph to three potential distribution shifts in ColoredMNIST. }
\label{fig:cmnist_normal}
\end{figure*}

\begin{figure*}[htbp!]
\begin{subfigure}{.28\linewidth}
  \centering
  \includegraphics[width=.94\linewidth]{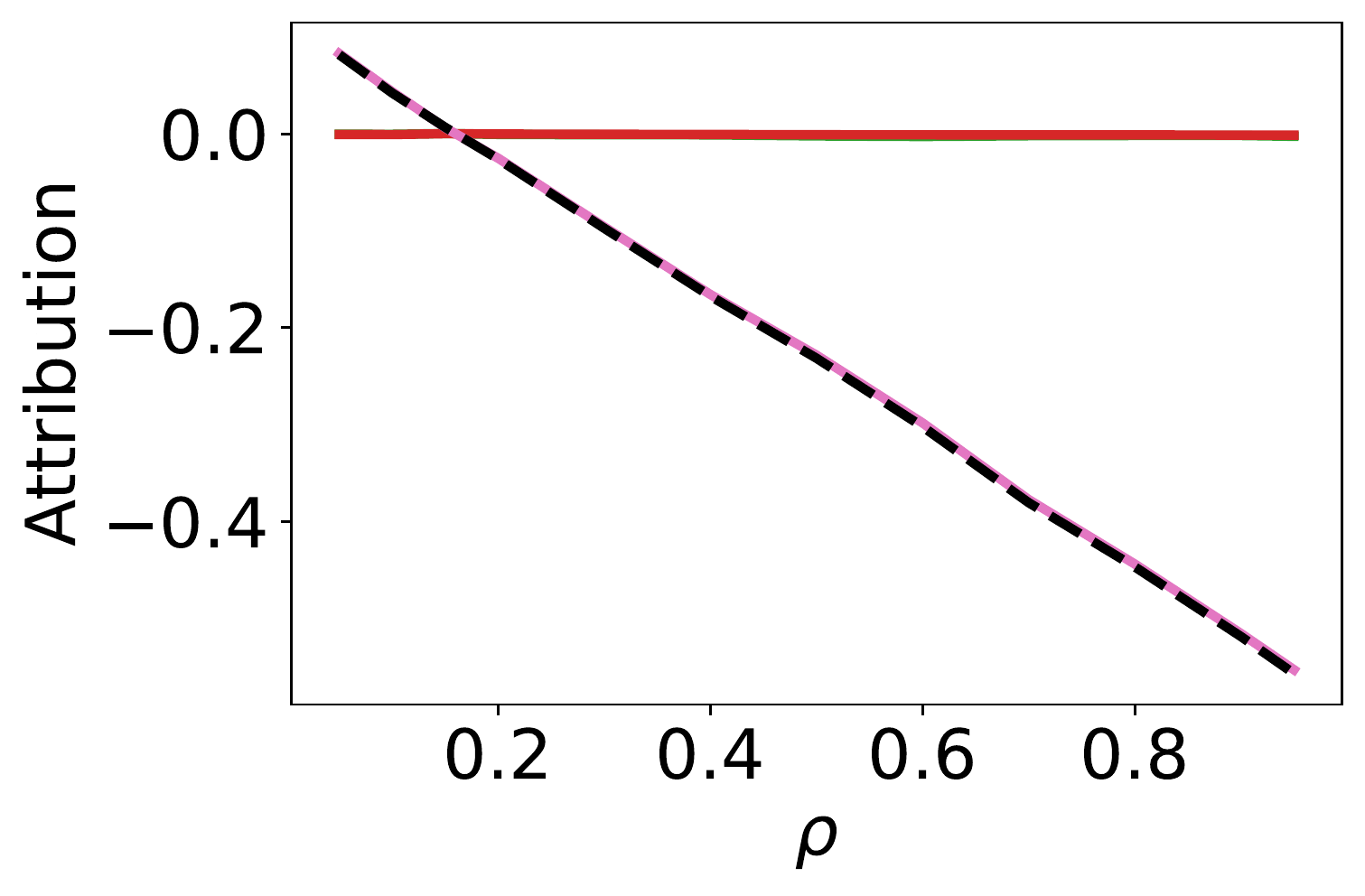}
  \caption{ERM}
\end{subfigure}%
\begin{subfigure}{.28\linewidth}
  \centering
  \includegraphics[width=.94\linewidth]{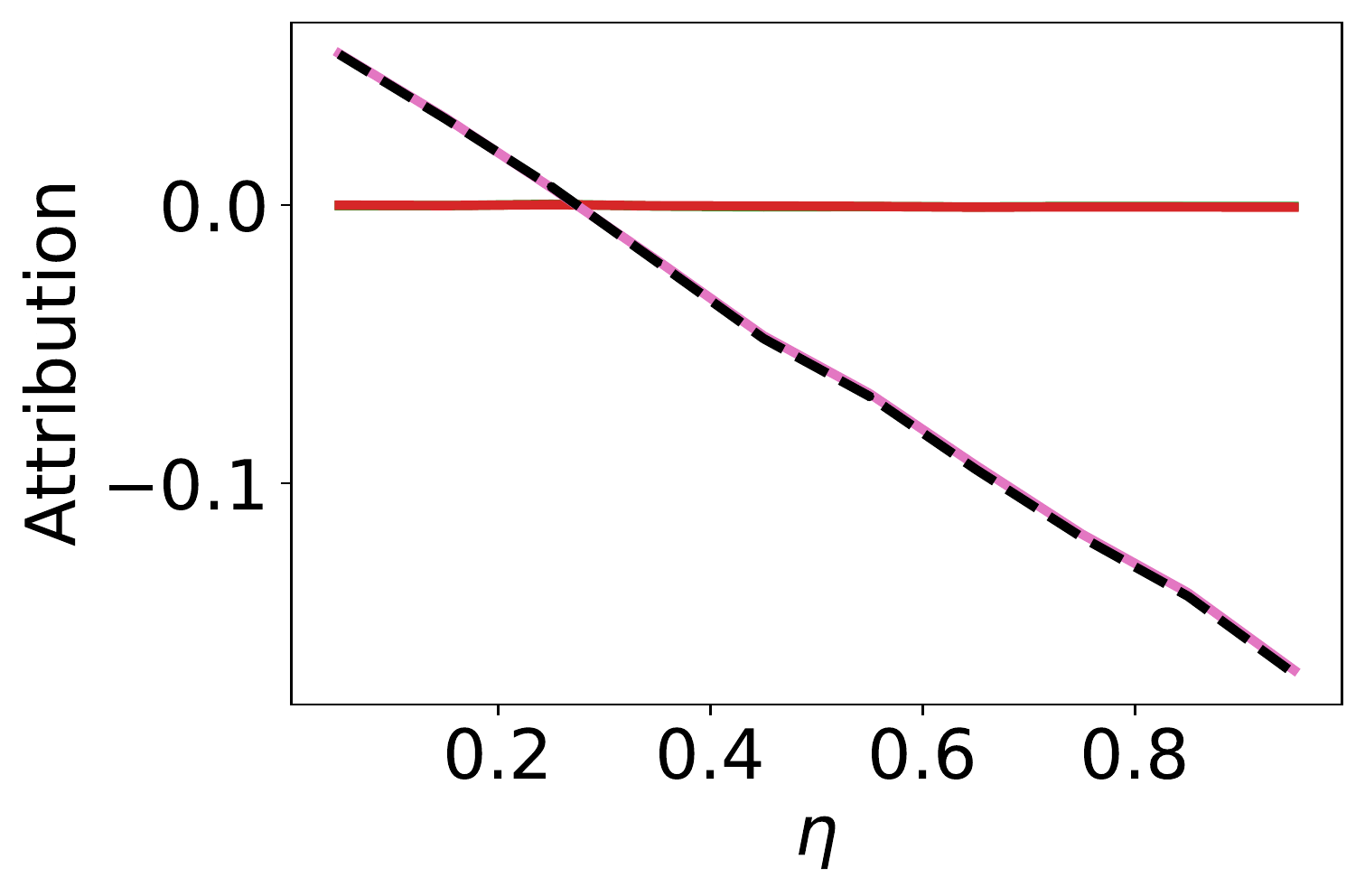}
  \caption{ERM}
\end{subfigure}
\begin{subfigure}{.39\linewidth}
  \centering
  \includegraphics[width=.99\linewidth]{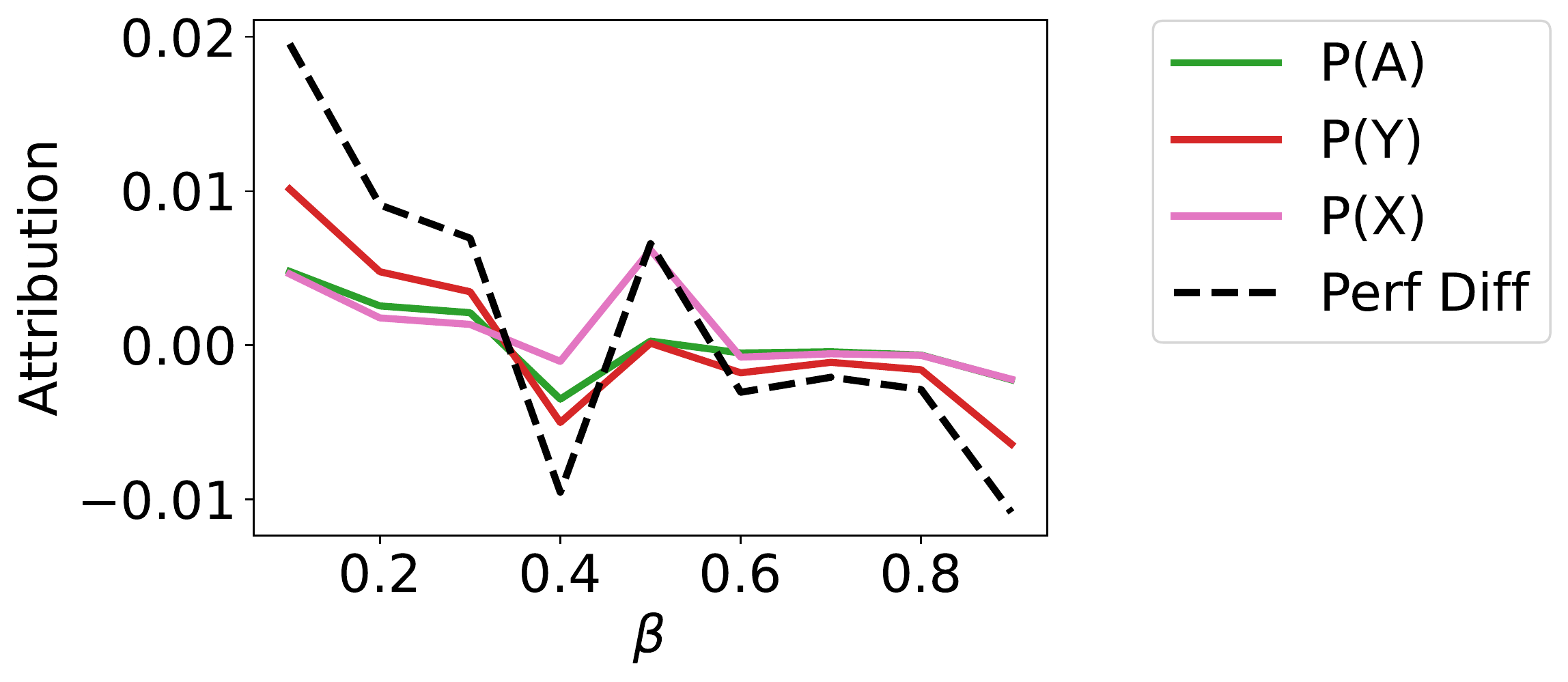}
  \caption{ERM}
\end{subfigure}%

\begin{subfigure}{.28\linewidth}
  \centering
  \includegraphics[width=.94\linewidth]{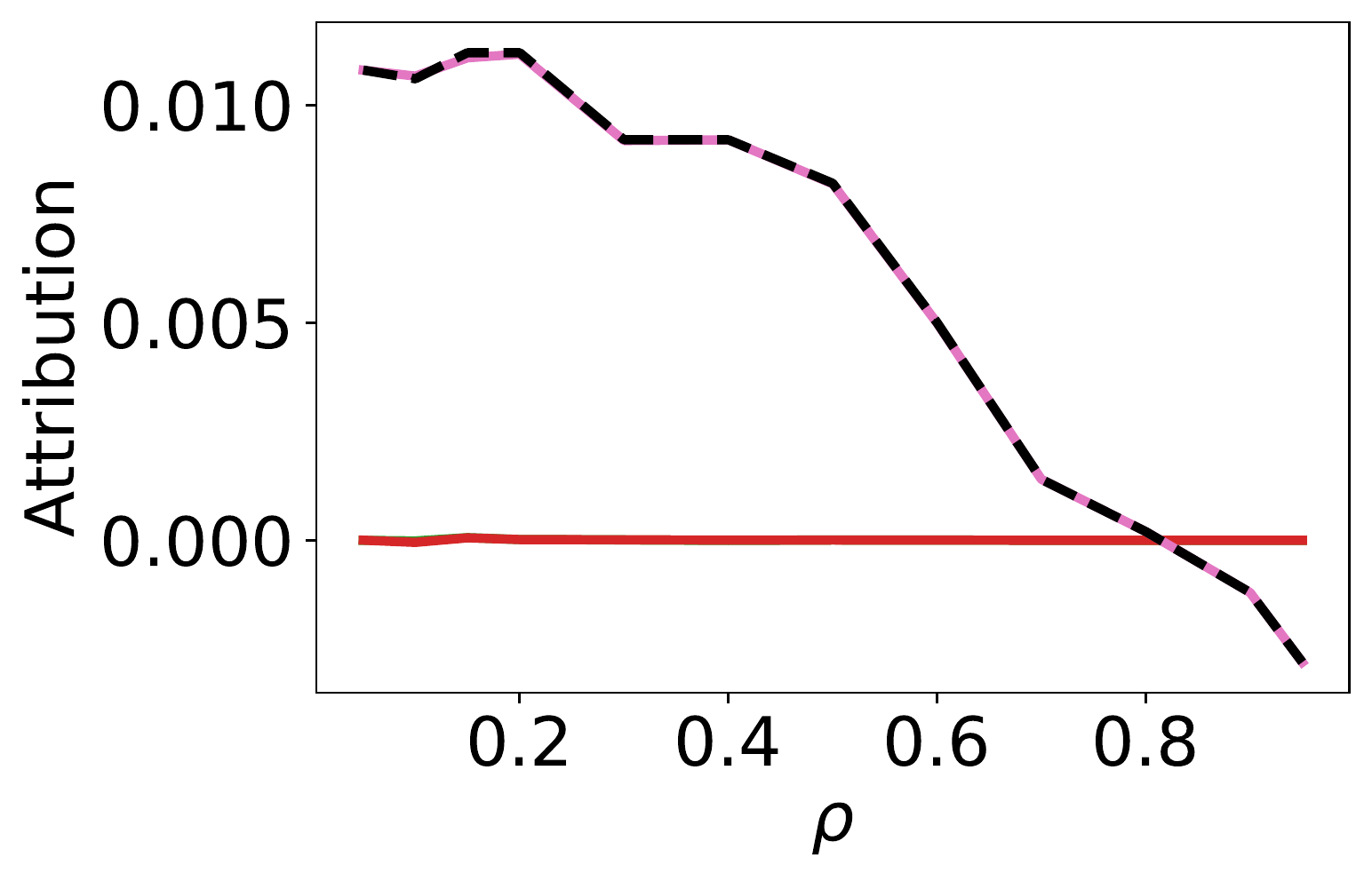}
  \caption{GroupDRO}
\end{subfigure}
\begin{subfigure}{.28\linewidth}
  \centering
  \includegraphics[width=.94\linewidth]{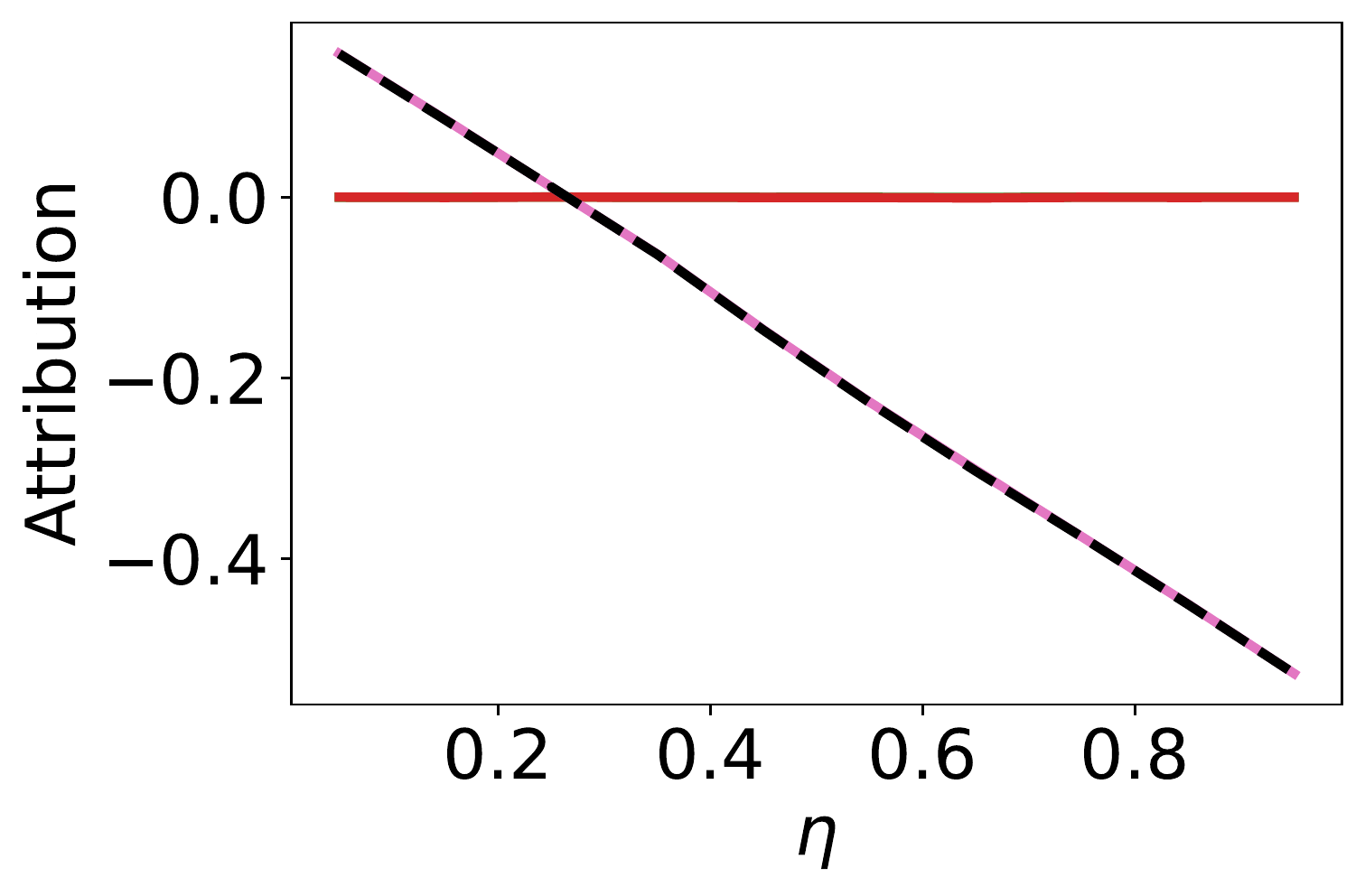}
  \caption{GroupDRO}
\end{subfigure}
\begin{subfigure}{.39\linewidth}
  \centering
  \includegraphics[width=.99\linewidth]{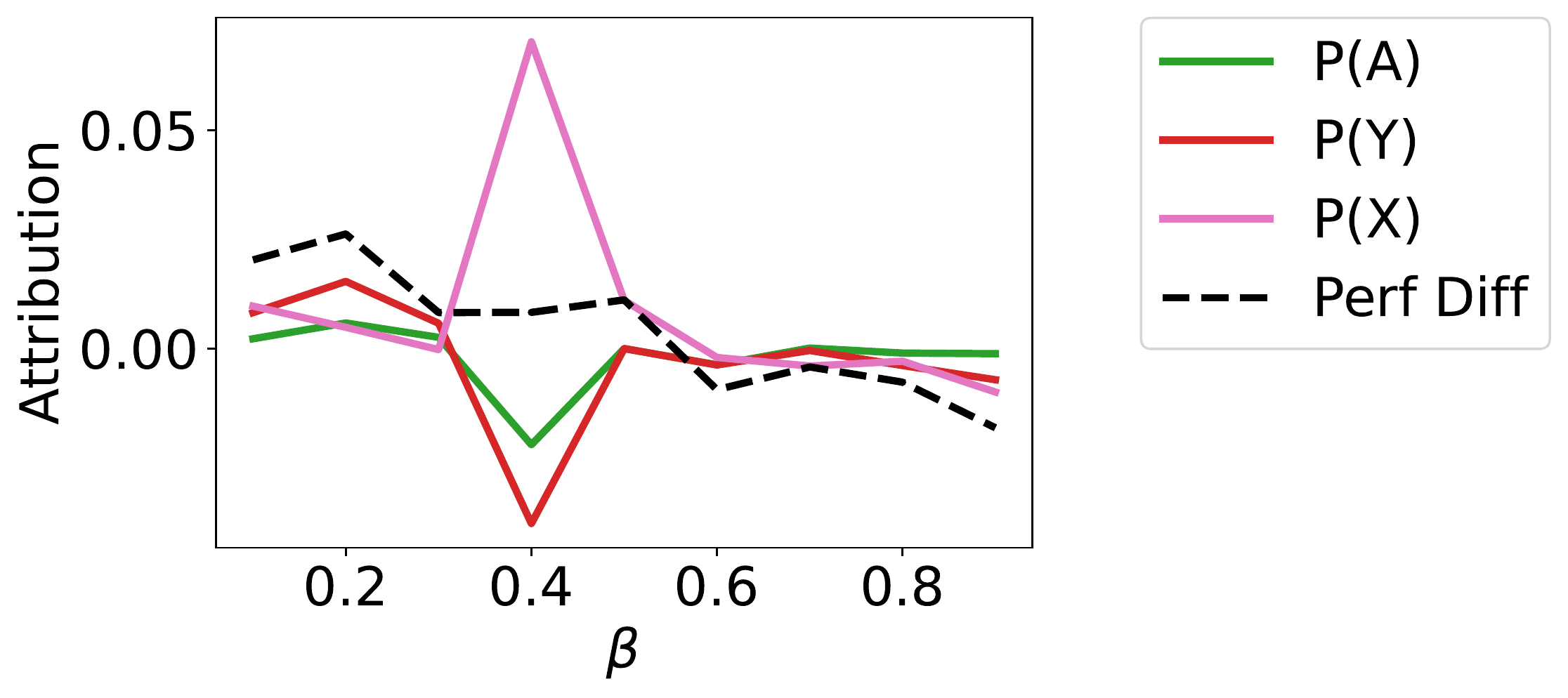}
  \caption{GroupDRO}
\end{subfigure}
\caption{Accuracy differences attributed by our model with the all marginals causal graph to three potential distribution shifts in ColoredMNIST. We observe that using a causal graph that does not match the underlying shifting mechanisms may lead to attributions that are not meaningful.}
\label{fig:cmnist_marginals}
\end{figure*}

\paragraph {Results.} In Figure \ref{fig:cmnist_normal}, we show the attributions provided by our method for the correct causal graph. We observe that the ERM model is highly susceptible to shifts in $\rho$ and correctly attributes all of the shift to $P(A|Y)$ in that setting. In contrast, the GroupDRO model receives almost no attribution to  $P(A|Y)$ as it does not use the attribute spuriously. However, since it makes use of the invariant signal, shifting $\eta$ results in large performance drops. Both models do not receive a significant attribution for $P(Y)$, as the error rate tends to be similar between the two classes. Looking at the results for the marginal causal graph in Figure \ref{fig:cmnist_marginals}, we find that attributions created using this candidate set are not meaningful as it deviates too far from the actual causal mechanism. For example, no shifts are ever attributed to $P(A)$, as this distribution is not changed by any parameters.

\FloatBarrier
\subsection{Gender Classification in CelebA}
\label{sec:celebA}

\begin{figure}[htbp!]
    \centering
  \includegraphics[width=.7\linewidth]{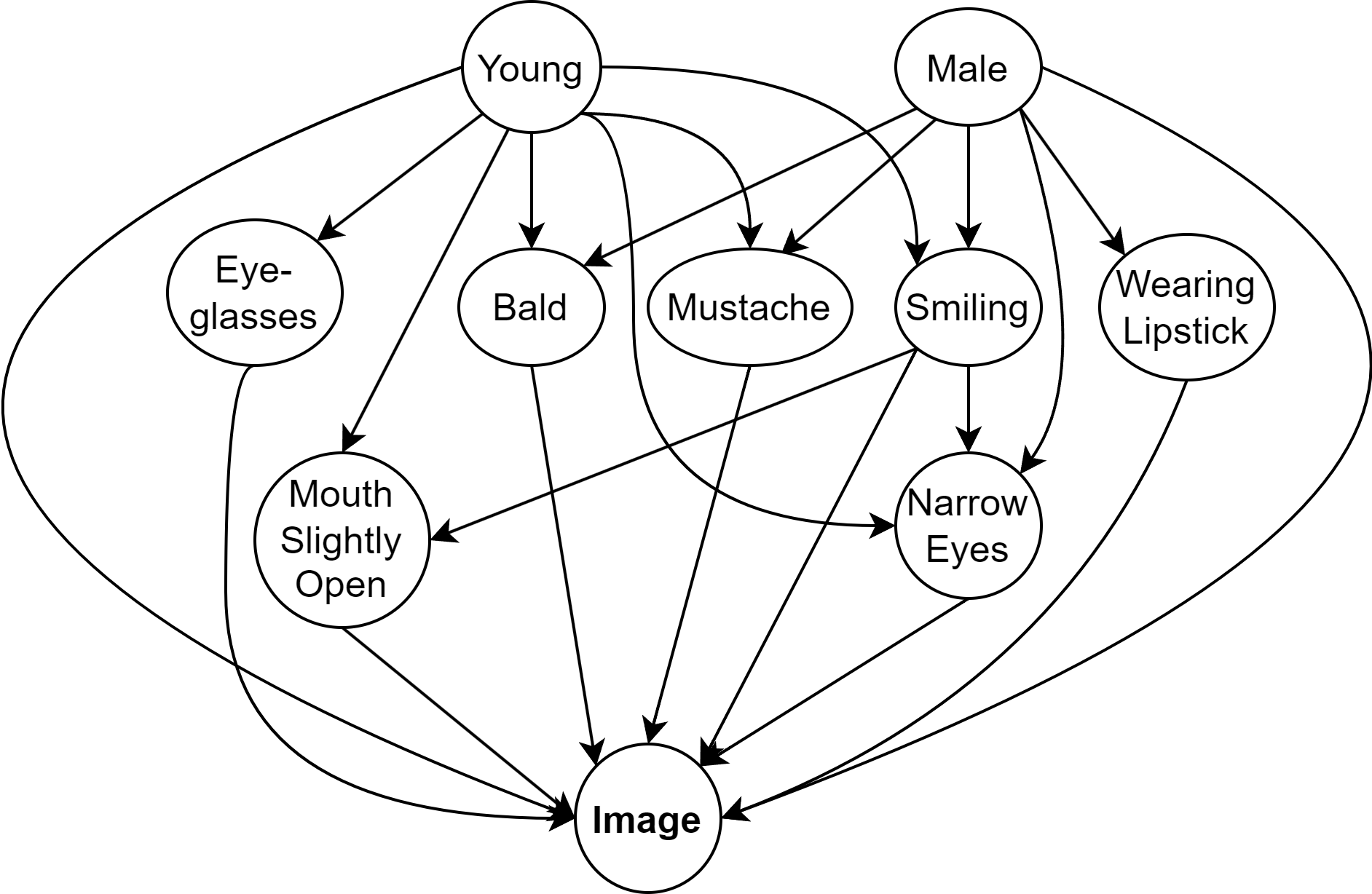}
  \caption{Causal graph for the celebA dataset.}
  \label{fig:celebA_graph}
\end{figure}

We use the CelebA dataset \citep{liu2015deep}, where the goal is to predict gender from facial images. We adopt a setup similar to the one presented in \citet{thams2022evaluating}. We assume this data is generated from the causal graph shown in Figure \ref{fig:celebA_graph}. We train a CausalGAN \citep{kocaoglu2017causalgan}, a generative model that allows us to synthesize images faithful to the graph. CausalGAN allows to train attribute nodes (young, bald, etc) which are binary-valued, and then synthesize images conditioned on specific attributes. This allows us to simulate known distribution shifts (in attributes and hence images) across environments.  We assume that the causal mechanisms in the source environment have log-odds equal to the ones shown in Table \ref{tab:celebA_log_odds}. We omit $\cD_{\text{Image} | \text{Pa(Image)}}$ from $\cand$, as 1) this distribution is parameterized by the CausalGAN and does not change, and 2) it is high-dimensional and difficult to work with. We investigate attribution to distribution shift of an  ImageNet-pretrained ResNet-18 \citep{he2016deep} finetuned to predict gender from the image using frozen representations. Note that the model is only given access to the image itself, but not any of the binary attributes in the causal graph. We conduct the following two experiments for evaluation.

\paragraph{Experiment 1.} The purpose of this experiment is to demonstrate that our method provides the correct attributions for a wide range of random shifts. To create the target environment, we first select the number of mechanisms to perturb, $n_p \in \{1, 2, ..., 6\}$. We select $n_p$ mechanisms from the causal graph, which we define as the ground truth shift. For each mechanism, we perturb one of the log odds by a quantity uniformly selected from $[-2.0, -1.0] \cup [1.0, 2.0]$. We then use the CausalGAN to simulate a dataset of $10,000$ images based on the modified mechanisms, and use our method to attribute the accuracy change between source and target. We select the $n_p$ distributions from our method with the largest attribution magnitude, and compare this set with the set of ground truth shifts to calculate an accuracy score. We repeat this experiment $20$ times for each value of $n_p \in \{1, 2, ..., 6\}$, and only select experiments with a non-trivial change in model performance (change in accuracy $\geq 1\%$).

\paragraph{Experiment 2.} The purpose of this experiment is to investigate the magnitude of our model attributions in the presence of multiple shifts. We perturb the log odds for $P(\text{Wearing Lipstick} | \text{Male})$ and $P(\text{Mouth Slightly Open} | \text{Smiling})$ jointly by $[-3.0, 3.0]$. We compare the magnitude of the attributions for the two associated mechanisms, relative to the total shift in accuracy. %

\begin{table}[htbp!]
\caption{Data generating process for the causal graph shown in Figure \ref{fig:celebA_graph}}
\centering
\begin{tabular}{@{}ll@{}}
\toprule
\textbf{Variable}   & \textbf{Log Odds}                               \\ \midrule
Young               & Base: 0.0                                       \\
Male                & Base: 0.0                                       \\
Eyeglasses          & Base: 0.0, Young: -0.4                          \\
Bald                & Base: -3.0, Male: 3.5, Young: -1.0              \\
Mustache            & Base: -2.5, Male: 2.5, Young: 0.5               \\
Smiling             & Base: 0.25, Male: -0.5, Young: 0.5              \\
Wearing Lipstick    & Base: 3.0, Male: -5.0                           \\
Mouth Slightly Open & Base: -1.0, Young: 0.5, Smiling: 1.0            \\
Narrow Eyes         & Base: -0.5, Male: 0.3, Young: 0.2, Smiling: 1.0 \\ \bottomrule
\end{tabular}
\label{tab:celebA_log_odds}
\end{table}

\begin{table}[htbp!]
\caption{Average accuracy of our method in attributing shifts to the ground truth shift in CelebA for each number of perturbed mechanisms ($n_p$).}
\centering
\begin{tabular}{@{}ll@{}}
\toprule 
 $n_p$ & Avg Accuracy \\
 \midrule
1 &  1.00 $\pm$ 0.00 \\
2 & 0.72 $\pm$ 0.36 \\
3 & 0.90 $\pm$ 0.16 \\
4 & 0.85 $\pm$ 0.13 \\
5 & 0.93 $\pm$ 0.10 \\
6 & 0.91 $\pm$ 0.09 \\
\bottomrule
\end{tabular}
\label{tab:celebA_exp1_acc}
\end{table}

\begin{figure}
\begin{subfigure}{0.33\textwidth}
  \centering 
  \includegraphics[width=0.95\linewidth]{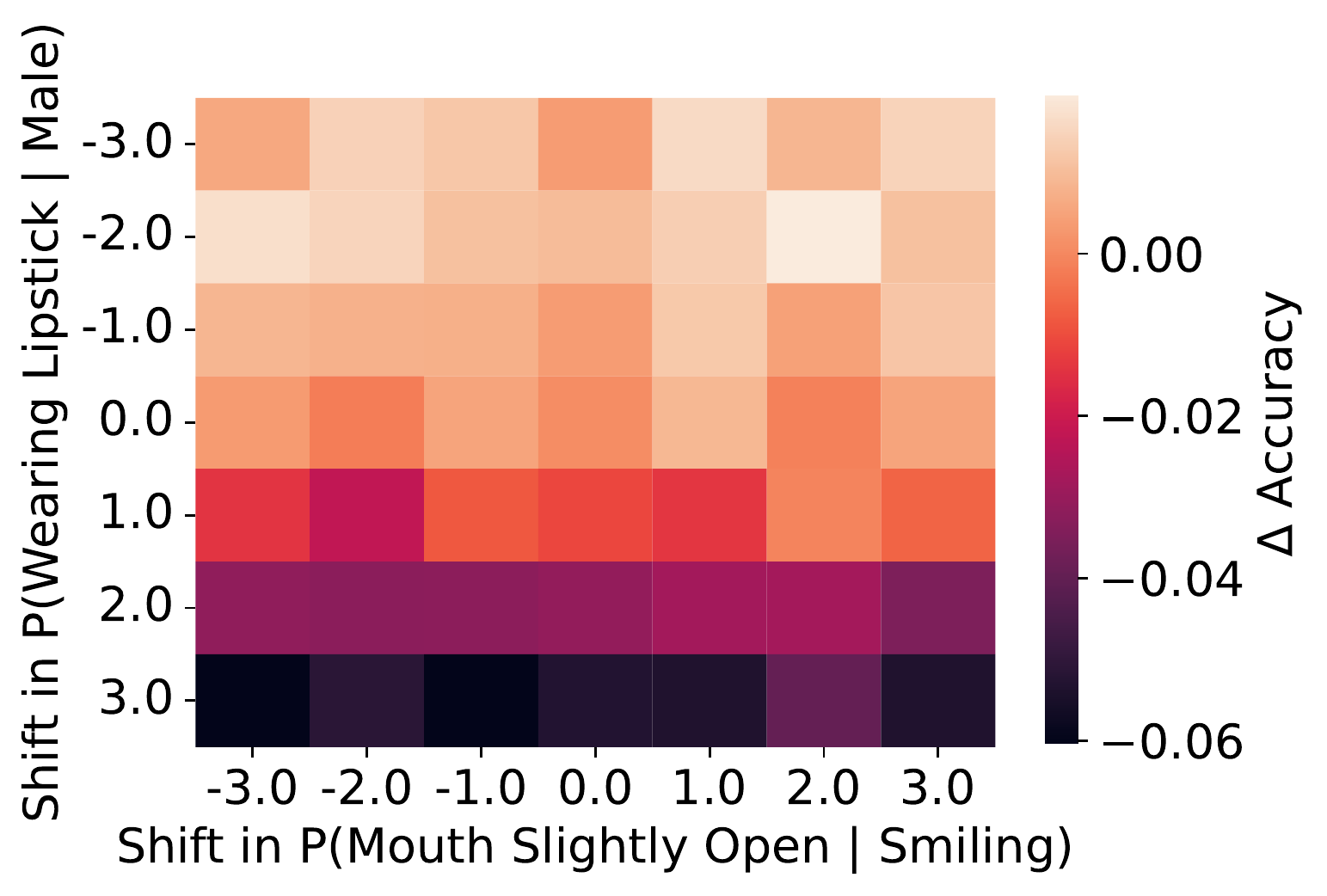}
    \caption{}
\end{subfigure}
\begin{subfigure}{0.33\textwidth}
  \centering 
  \includegraphics[width=0.95\linewidth]{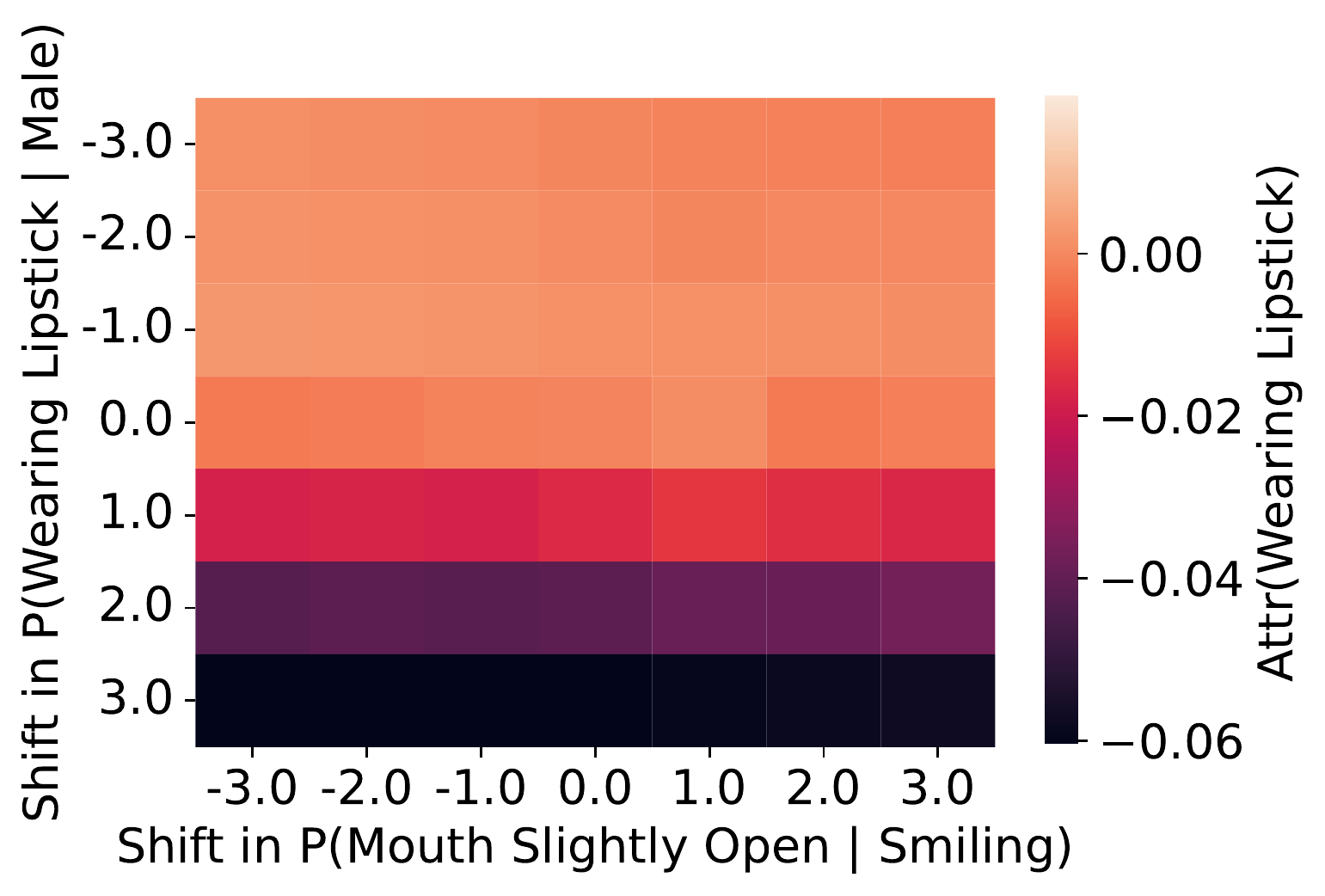}
    \caption{}
\end{subfigure}
\begin{subfigure}{0.33\textwidth}
  \centering 
  \includegraphics[width=0.95\linewidth]{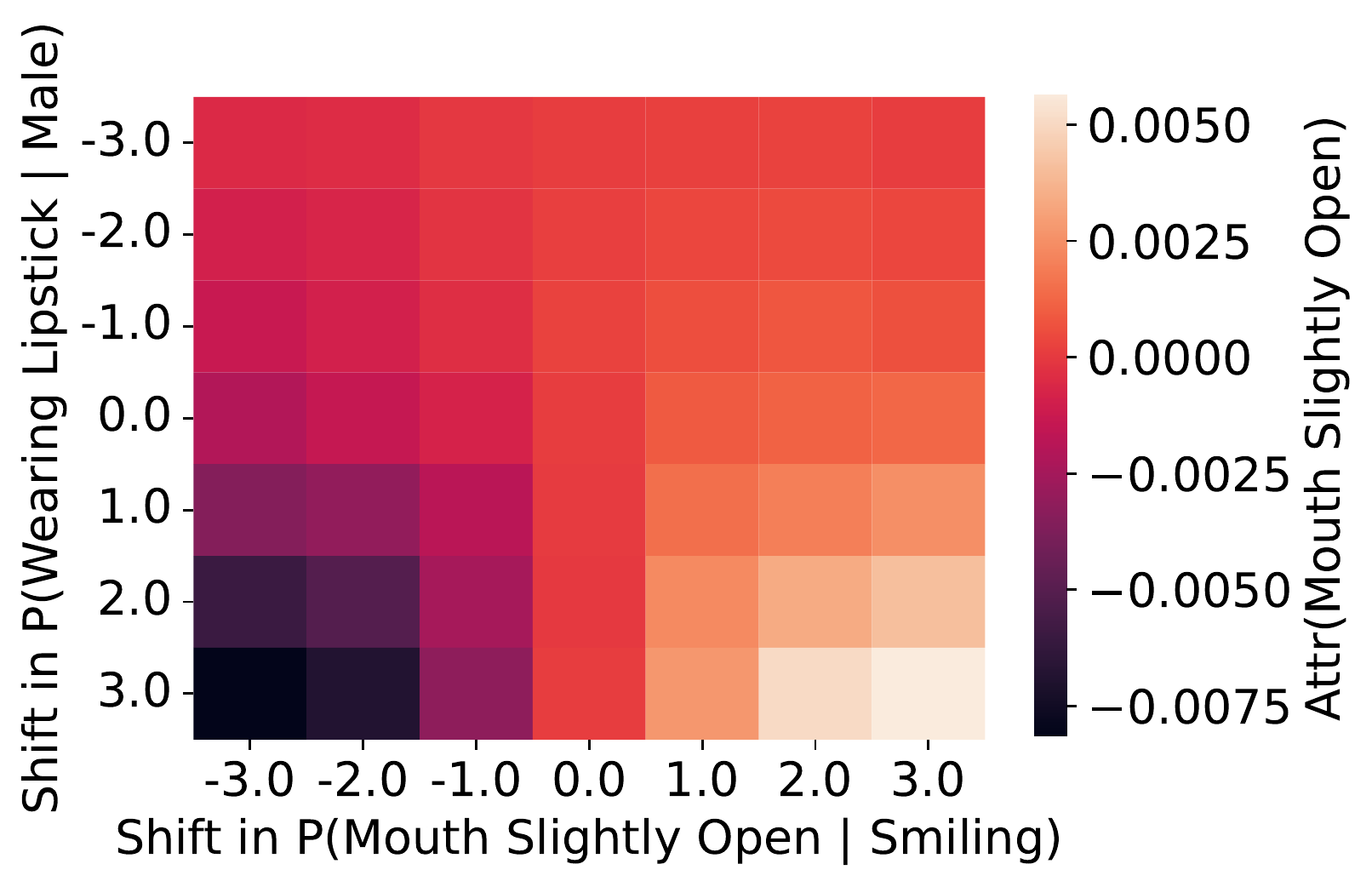}
    \caption{}
\end{subfigure}
    \caption{We vary the perturbation in log odds in the target environment for the ``wearing lipstick'' and ``mouth slightly open'' attributes. We show (a) the total shift in accuracy, (b) our attribution to $P(\text{Wearing Lipstick} | \text{Male})$, (c) our attribution to $P(\text{Mouth Slightly Open} | \text{Young, Smiling})$. }
    \label{fig:celebA_heatmaps}
\end{figure}

\begin{table}[htbp!]
\caption{Predictive performance of XGB models trained to predict attributes from the source environment in CelebA, and the correlation of each attribute the gender label, as measured by the Matthews Correlation Coefficient (MCC).}
\centering
\begin{tabular}{@{}llll@{}}
\toprule 
& \multicolumn{2}{l}{Predictive Performance} & Correlation \\
\cmidrule(lr){2-3} \cmidrule(lr){4-4} 
& AUROC & AUPRC & MCC \\ \midrule
Wearing Lipstick & 0.968 & 0.976 & -0.837 \\
Mouth Slightly Open & 0.927 & 0.924 & -0.036\\
\bottomrule
\end{tabular}
\label{tab:celebA_pred_perf}
\end{table}

\paragraph{Results.} In Table \ref{tab:celebA_exp1_acc}, we show the average accuracy of our method for each value of $n_p$. We find that our method achieves roughly 90\% accuracy at this task. However, we note that this is not the ideal scenario to validate our method, as not all shifts in the ground truth set will result in a decrease in the model performance. As our method will not attribute a significant value to shifts which do not impact model performance, this explains the accuracy discrepancy observed.

In Figure \ref{fig:celebA_heatmaps}, we show the output of our method in Experiment 2. First, we find that shifting these two attributes causes a large decrease in the accuracy (up to 6\%), and that $P(\text{Wearing Lipstick} | \text{Male})$ seem to be the stronger factor responsible for the decrease. Looking at our attributions, we find that we indeed attribute the large majority of the shift to $P(\text{Wearing Lipstick} | \text{Male})$. Here, the relative attribution to $P(\text{Wearing Lipstick} | \text{Male})$ is relatively unaffected by the shift in the other variable, as its effect on the total shift is so minuscule. However, looking at the attribution to $P(\text{Mouth Slightly Open} | \text{Young, Smiling})$, in addition to the small magnitude, we do observe an interesting effect, where the attributed accuracy drop is greater when the two shifts are combined.

To justify the magnitude of our attributions, we use an ad-hoc heuristic that attempts to approximate the model reliance on each attribute in making its prediction. First, we train XGBoost models on the ResNet-18 embeddings from the source environment to predict the two attributes. From Table \ref{tab:celebA_pred_perf}, we find that ``Wearing Lipstick'' is easier to infer from the representations than ``Mouth Slightly Open''. Next, we measure the correlation of each attribute to the label (gender), finding that the magnitude of the correlation is also much higher for ``Wearing Lipstick''. As ``Wearing Lipstick'' is both easier to detect from the image, and is also a stronger predictor of gender, it seems reasonable to conclude that the model trained on the source would utilize it more in its predictions, and thus our method should attribute more of the performance drop to the ``Wearing Lipstick'' distribution when it shifts. %

\FloatBarrier
\subsection{eICU Data}
\label{app:add_details_eicu}

\begin{figure}[htbp!]
   \centering
  \includegraphics[width=.3\linewidth]{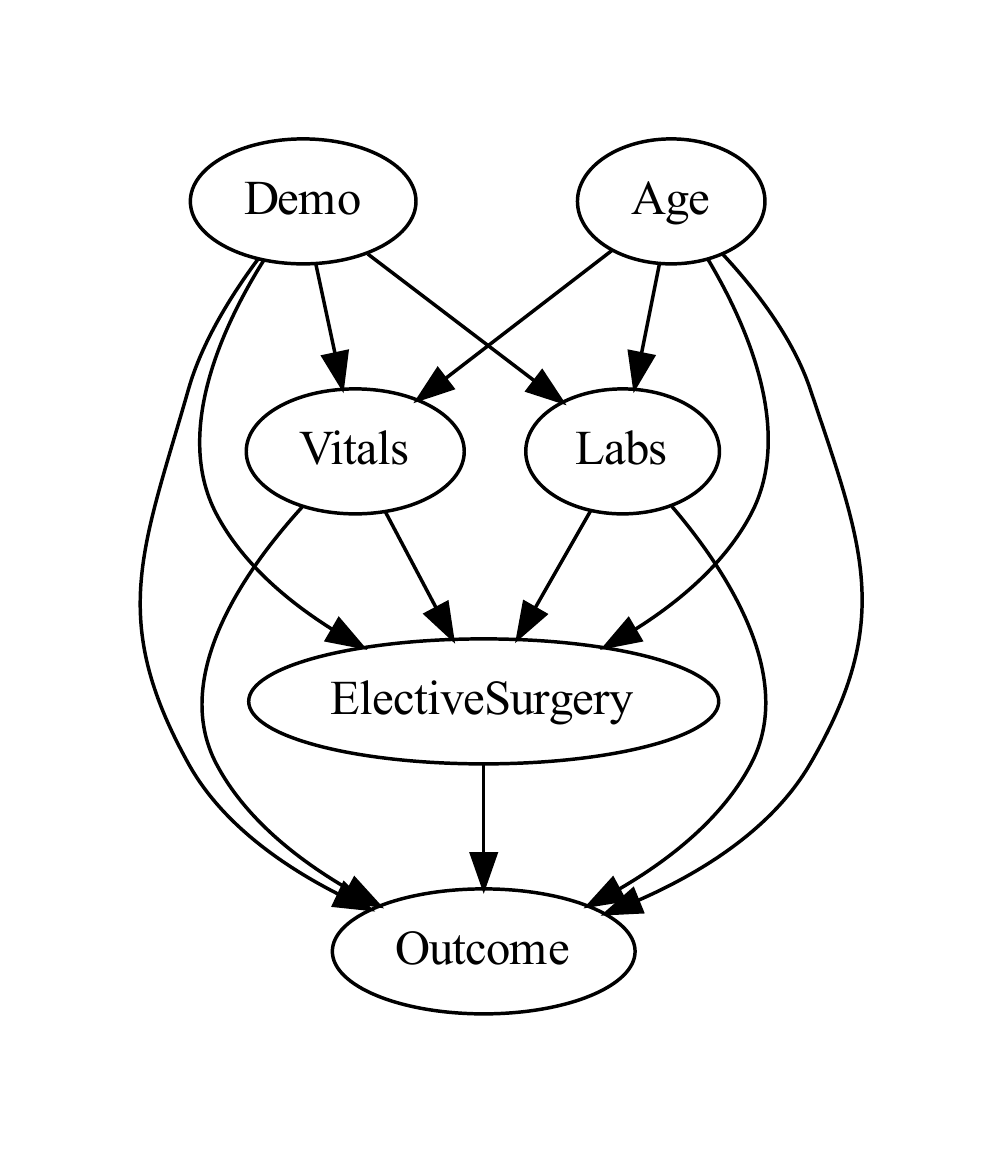}
  \caption{Causal Graph for eICU data}
\label{fig:eicu_graph}
\vspace{-2mm}
\end{figure}

Table \ref{tab:eICU_variables} lists the features that comprise the nodes in the causal graph. Please refer to \citep[Supporting Information Table C]{singh2022generalizability} for descriptions. Code for preprocessing the eICU database for the mortality prediction task is made available at the Github repository by \citet{johnson2018eicu}.

\begin{table}[htbp!]
\caption{Features comprising the nodes of the causal graph in Figure \ref{fig:eicu_graph}.}
\centering
\begin{tabular}{@{}ll@{}}
\toprule
\textbf{Variable}   & \textbf{Features}                               \\ \midrule
Demo               & is\_female, race\_black, race\_hispanic, race\_asian, race\_other                                       \\
Vitals                & heartrate, sysbp, temp, bg\_pao2fio2ratio, urineoutput                                       \\
Labs          & bun, sodium, potassium, bicarbonate, bilirubin, wbc, gcs                          \\
Age                & age              \\
ElectiveSurgery            & electivesurgery               \\
Outcome             & death              \\ \bottomrule
\end{tabular}
\label{tab:eICU_variables}
\end{table}

The Midwest domain has 10,056 samples, and the South domain has 7,836. Both domains have 20 features and a binary outcome. We randomly split each into 50\% for training the XGBoost model and 50\% for evaluation (and estimation of Shapley values). To create the resampled Midwest dataset, we subsample 67\% of the training set but selectively sample records with age less than 63 (which is the median age in Midwest) with probability 5 times that of the probability of sampling the rest of the records.
\section{Convergence Analysis}
\label{app:convergence}

We present a preliminary analysis to study the impact of errors resulting from estimating importance weights on the properties of the Shapley values. The theoretical analysis is informal and presented here with the goal of motivating further study. Importantly, we experimentally evaluate the error in a synthetic setup.

\subsection{Sketch for a Theoretical Analysis}

\begin{remark}
    Under bounded estimation error and for a bounded loss, Property 2.1 holds asymptotically.
\end{remark}

\begin{proof}
    Suppose $ 1 - \epsilon_d^n \leq  w_d \leq 1 + \epsilon_d^n$, that is we get approximate importance weights from finite samples. Then for a bounded loss,
    \begin{equation}
        \begin{aligned}
           - \frac{\epsilon_{d}^n l}{|\cand|} \sum_{\subcand\subseteq \cand \setminus \{d\}} \binom{|\cand| - 1}{|\subcand|}^{-1} \mathbb{E}_{\psrc}[\prod_{\tilde{d} \in \subcand \cup d} w_{\tilde{d}}]  \leq \text{Attr}^n(d)  \leq \frac{\epsilon_d^n l}{|\cand|} \sum_{\subcand\subseteq \cand \setminus \{d\}} \binom{|\cand| - 1}{|\subcand|}^{-1} \mathbb{E}_{\psrc}[\prod_{\tilde{d} \in \subcand \cup d} w_{\tilde{d}}]
        \end{aligned}
    \end{equation}

    Suppose that in the worst case all other distributions shift except $d$, but the shifts are bounded, i.e.  $\frac{1}{\eta} < w_{\tilde{d}} < \eta$ where $\eta>1$ for all $\tilde{d} \in \cand \setminus \{d\}$. Then,
    \begin{equation}
        \begin{aligned}
        &- \frac{\epsilon_{d}^n l}{|\cand|} \sum_{\subcand\subseteq \cand \setminus \{d\}} \binom{|\cand| - 1}{|\subcand|}^{-1} \left(\frac{1}{\eta}\right)^{|\subcand| - 1}  \leq \text{Attr}^n(d)  \leq \frac{\epsilon_d^n l}{|\cand|} \sum_{\subcand\subseteq \cand \setminus \{d\}} \binom{|\cand| - 1}{|\subcand|}^{-1} {\eta}^{|\subcand| - 1} \\
        & - \frac{\epsilon_{d}^n l}{|\cand|} \left(1 + \frac{1}{\eta}\right)^{|\subcand| - 1}  \leq \text{Attr}^n(d)  \leq \frac{\epsilon_d^n l}{|\cand|}  {(1+\eta)}^{|\subcand| - 1}
        \end{aligned}
        \end{equation}

        The error in attribution is dominated by the shifts in other distribution and the error in estimating the weight for distribution $d$. Thus as $n \to \infty$, so long as $\epsilon_d^n \to 0$, the attribution $\text{Attr}^n(d) \to 0$. 
\end{proof}
The above suggests that our Property 2.1 may not hold exactly in finite samples due to estimation error.

\subsection{Empirical Analysis}
To empirically examine the estimation error as a function of the number of samples, we adopt the synthetic setup described in Appendix \ref{app:synthetic} with 
$\theta_2 = 0.5$ and $\mu_2 = 0.5$. We choose this setup because the exact importance weights can be computed analytically, and thus allows us to quantify the error of an importance weight estimator. We experiment with the KLIEP method \cite{sugiyama2008direct}, as well as using logistic regression (LR) and XGBoost (XGB) as probabilistic estimators, the last of which we use in the paper. Given $n$ samples, we randomly choose $n/2$ samples to train the importance weight estimator, and the remaining $n/2$ samples to evaluate the attribution. In each run, we compute the mean squared error between empirical and exact importance weights for 
$\cD_X$ and $\cD_{Y|X}$, as well as the mean squared error between the empirical and exact attributions $Attr(\cD_X)$ and $Attr(\cD_{Y|X})$. Note that the analytical attributions are 
$Attr(\cD_X) = -0.06375$ and 
$Attr(\cD_{Y|X}) = 0.16875$. We display the result in Table \ref{tab:convergence}.

\begin{table}[!ht]
    \centering
    \caption{Estimation error in Shapley attributions for finite samples, using the synthetic setup described in Appendix \ref{app:synthetic}.}
    \begin{tabular}{lrrrrr}
    \toprule
        \textbf{Model} & $n$ & $MSE(w_X)$ & $MSE(w_{X, Y})$ & $MSE(Attr(\cD_X))$ & $MSE(Attr(\cD_Y))$ \\ 
        \midrule
        \multirow{ 3}{*}{KLIEP} & 100 & 0.557 $\pm$ 0.339 & 1.093 $\pm$ 0.422 & 0.016 $\pm$ 0.003 & 0.029 $\pm$ 0.015 \\ 
         & 200 & 0.344 $\pm$ 0.194 & 1.612 $\pm$ 0.809 & 0.004 $\pm$ 0.000 & 0.037 $\pm$ 0.002 \\ 
         & 1000 & 0.212 $\pm$ 0.145 & 0.588 $\pm$ 0.203 & 0.003 $\pm$ 0.001 & 0.041 $\pm$ 0.012 \\ \midrule
        \multirow{8}{*}{LR} & 20 & 0.237 $\pm$ 0.133 & 0.428 $\pm$ 0.221 & 0.004 $\pm$ 0.000 & 0.028 $\pm$ 0.000 \\ 
         & 50 & 0.748 $\pm$ 0.707 & 14.175 $\pm$ 22.402 & 0.004 $\pm$ 0.000 & 0.023 $\pm$ 0.005 \\ 
         & 100 & 0.555 $\pm$ 0.064 & 0.549 $\pm$ 0.342 & 0.004 $\pm$ 0.000 & 0.033 $\pm$ 0.009 \\ 
         & 200 & 0.543 $\pm$ 0.083 & 1.861 $\pm$ 1.864 & 0.004 $\pm$ 0.000 & 0.036 $\pm$ 0.012 \\ 
         & 1000 & 0.600 $\pm$ 0.042 & 4.549 $\pm$ 3.196 & 0.004 $\pm$ 0.000 & 0.010 $\pm$ 0.007 \\ 
         & 5000 & 0.583 $\pm$ 0.061 & 3.208 $\pm$ 0.269 & 0.004 $\pm$ 0.000 & 0.008 $\pm$ 0.003 \\ 
         & 10000 & 0.566 $\pm$ 0.038 & 5.887 $\pm$ 3.753 & 0.004 $\pm$ 0.000 & 0.007 $\pm$ 0.004 \\ 
         & 50000 & 0.325 $\pm$ 0.008 & 3.631 $\pm$ 0.430 & 0.004 $\pm$ 0.000 & 0.009 $\pm$ 0.002 \\  \midrule
        \multirow{8}{*}{XGB} & 20 & 1.191 $\pm$ 1.034 & 0.974 $\pm$ 0.740 & 0.003 $\pm$ 0.003 & 0.068 $\pm$ 0.038 \\ 
         & 50 & 1.003 $\pm$ 0.848 & 40.087 $\pm$ 66.979 & 0.010 $\pm$ 0.008 & 0.022 $\pm$ 0.004 \\ 
         & 100 & 13.151 $\pm$ 19.926 & 30.083 $\pm$ 25.079 & 0.017 $\pm$ 0.016 & 0.013 $\pm$ 0.020 \\ 
         & 200 & 10.053 $\pm$ 8.450 & 7.491 $\pm$ 9.321 & 0.001 $\pm$ 0.001 & 0.031 $\pm$ 0.015 \\ 
         & 1000 & 0.418 $\pm$ 0.378 & 39.845 $\pm$ 66.373 & 0.005 $\pm$ 0.003 & 0.031 $\pm$ 0.023 \\ 
         & 5000 & 0.419 $\pm$ 0.502 & 3.485 $\pm$ 2.964 & 0.003 $\pm$ 0.003 & 0.030 $\pm$ 0.025 \\ 
         & 10000 & 0.065 $\pm$ 0.062 & 0.903 $\pm$ 0.429 & 0.001 $\pm$ 0.001 & 0.014 $\pm$ 0.018 \\ 
         & 50000 & 0.083 $\pm$ 0.050 & 0.396 $\pm$ 0.095 & 0.000 $\pm$ 0.000 & 0.002 $\pm$ 0.001 \\ 
        \bottomrule
    \end{tabular}
    \label{tab:convergence}
\end{table}

We first find that KLIEP did not converge for smaller values of 
$n$, and takes prohibitively long to run for larger values of 
$n$. It was for these reasons that we did not select it for use in the main paper. Next, we note that a linear model is not complex enough to differentiate between the source and target, and thus results in a biased estimator (non-zero MSE for large n), though this still results in relatively small MSE for the attributions. Finally, we observe that the errors of XGB converge to zero both in the estimated importance weights as well as in the resulting attributions.

\end{document}